%% file: anonymous-submission-latex-2025.tex
\documentclass[letterpaper]{article} 
\usepackage[draft]{aaai25}  
\usepackage{times}  
\usepackage{helvet}  
\usepackage{courier}  
\usepackage[hyphens]{url}  
\usepackage{graphicx} 
\usepackage{natbib}  
\usepackage{caption} 
\usepackage{amsmath}
\usepackage{amssymb}
\usepackage[ruled,vlined,noresetcount]{algorithm2e}
\usepackage[utf8]{inputenc} 
\usepackage[T1]{fontenc}    
\usepackage{url}            
\usepackage{booktabs}       
\usepackage{amsfonts}       
\usepackage{nicefrac}       
\usepackage{microtype}      
\usepackage[table]{xcolor}  
\usepackage{amsmath}        
\usepackage{paralist}       
\usepackage{xcolor}
\usepackage{hyperref}

\SetKwComment{Comment}{// }{}
\RestyleAlgo{ruled}
\usepackage{newfloat}
\usepackage{listings}
\usepackage{caption}
\usepackage{graphicx}

\definecolor{darker}{rgb}{0.357,0.608,0.835}
\definecolor{dark}{rgb}{0.824,0.87,0.937}
\definecolor{light}{rgb}{0.917,0.937,0.968}
\definecolor{subtotal}{rgb}{0.76, 0.8, 0.88}
\definecolor{total}{rgb}{0.6, 0.75, 0.88}

\newcommand{\argmax}{\mathop{\mathrm{argmax}}\limits}

\DeclareCaptionStyle{ruled}{labelfont=normalfont,labelsep=colon,strut=off}

\title{Active Reinforcement Learning Strategies for Offline Policy Improvement}
\author{
    Ambedkar Dukkipati,
    Ranga Shaarad Ayyagari\equalcontrib,
    Bodhisattwa Dasgupta\equalcontrib,\\
    Parag Dutta\equalcontrib,
    Prabhas Reddy Onteru
}
\affiliations{
    Department of Computer Science and Automation\\
    Indian Institute of Science\\

    Bengaluru, India - 560012\\
    \{ambedkar, rangaa, bodhisattwad, paragdutta, prabhasreddy\}@iisc.ac.in
}

\begin{document}
\maketitle

\begin{abstract}
    Learning agents that excel at sequential decision-making tasks must continuously resolve the problem of exploration and exploitation for optimal learning. 
    However, such interactions with the environment online might be prohibitively expensive and may involve some constraints, such as a limited budget for agent-environment interactions and restricted exploration in certain regions of the state space. Examples include selecting candidates for medical trials and training agents in complex navigation environments.
    This problem necessitates the study of active reinforcement learning strategies that collect minimal additional experience trajectories by reusing existing offline data previously collected by some unknown behavior policy.  In this work, we propose an active reinforcement learning method capable of collecting trajectories that can augment existing offline data. 
    With extensive experimentation, we demonstrate that our proposed method reduces additional online interaction with the environment by up to $75\%$ over competitive baselines across various continuous control environments such as \textbf{\texttt{Gym-MuJoCo}} locomotion environments as well as 
    \textbf{\texttt{Maze2d}}, \textbf{\texttt{AntMaze}}, \textbf{\texttt{CARLA}} and \textbf{\texttt{IsaacSimGo1}}.
    To the best of our knowledge, this is the first work that addresses the active learning problem in the context of sequential decision-making and reinforcement learning. 
\end{abstract}

%
\begin{links}
    \link{Code}{https://github.com/sml-iisc/ActiveRL}
\end{links}

\section{Introduction}
\label{Section: Introduction}
Reinforcement learning~\cite{kaelbling1996reinforcement, sutton2018reinforcement} tackles the problem of sequential decision-making in unknown environments. This involves agents exploring the environment to learn from the interactions online. However, relying solely on online interactions may not be feasible in many practical applications like navigation and clinical trials. To overcome this limitation, recently, offline reinforcement learning~\cite{levine2020offline, fujimoto2019offpolicy} has emerged, where agents can learn from offline interactions with unknown policies. Although this presents significant challenges and some advancements have been made, a pertinent question arises: In the presence of offline data, how can one enhance an agent's performance when it is permitted to explore only a limited number of interactions with the environment? 

For example, in the context of clinical trials, determining the optimal treatment regimen may require access to existing data from related studies while seeking to gather more relevant and informative data by administering treatments to new patients. Each prospective patient presents unique treatment needs and medical histories, often at varying stages of diagnosis. With a constrained budget, it is essential to devise algorithms capable of selecting the optimal subset of participants for treatment, thereby enhancing the utility of the existing data. 
While this is similar to active learning in a supervised learning setting, it is still an open problem to learn active strategies in reinforcement learning. The main aim of the paper is to address this problem.  


\begin{figure}[t]
    \centering
    \includegraphics[width=\linewidth]{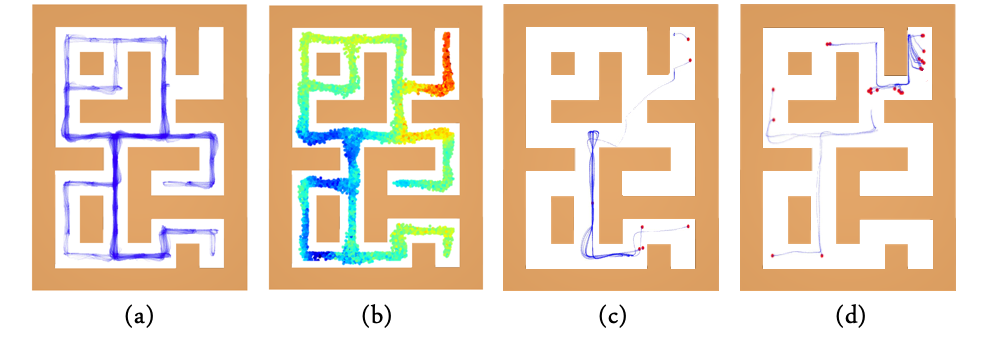}
    \caption{[Best viewed in color] Consider an offline dataset as shown in (a). Our method computes uncertainties in various regions of the environment according to the dataset. As shown in (b), the uncertainties are high in regions where data is not present in the dataset. We collect new trajectories starting from the uncertain regions since that provides more information to the learning algorithm. As can be seen in (c), a simple online trajectory collection policy collects redundant trajectories, while our method focuses on previously unobserved regions, as evident from (d).}
    \label{Figure: BlockDiagram}
\end{figure}

Due to budget limitations, there exists a further challenge in deciding whether to pursue a lengthy trajectory or to gather several shorter trajectories across different areas. For instance, in the context of enhancing the navigation capabilities of autonomous vehicles, it is unnecessary to collect data in areas where the existing dataset already contains representative samples. In such instances, any trajectory that enters these regions can be truncated, allowing for the initiation of a new trajectory in a location that offers more informative data.

In the realm of supervised learning, the issue of determining effective methods for data collection is referred to as Active Learning~\cite{cohn1996activelearning, settlesactivelearninginpractice, bachman2017learningalgforactivelearning}. In this scenario, the agent operates with a limited quantity of labeled data alongside a vast pool of unlabeled data, which incurs significant costs for annotation. The primary goal is to select a small subset of unlabeled data for labeling, thereby enhancing the performance of a model trained on this enriched labeled dataset.

This task becomes increasingly complex in the context of sequential decision-making problems, as the data is represented by samples and trajectories that remain unknown until the agent engages with the environment through an exploration policy. Consequently, rather than merely selecting which data points to label, the agent must decide where, how, and to what extent to explore the environment.



In this work, we develop learning algorithms for enhancing active exploration of the environment utilizing an existing offline dataset. We focus on scenarios where the agent is allocated a limited exploration budget, necessitating the efficient collection of data that can augment the offline data.

\subsection*{Contributions}

We consider the problem of active exploration in the context of offline reinforcement learning to minimally augment the offline dataset with informative trajectories to learn an optimal policy.
\begin{enumerate}
	\item We propose a representation-aware epistemic uncertainty-based method for determining regions of the state space where the agent should collect additional trajectories.
	\item We propose an uncertainty-based exploration policy for online trajectory collection that re-uses the representation models.
	\item Through extensive experimentation, we empirically demonstrate that our approaches can be widely applicable across a range of continuous control environments. Our active trajectory collection method reduces the need for online interactions by up to $75\%$ when compared to existing fine-tuning approaches.
	 \item We also perform ablation experiments to demonstrate the importance of each component of our algorithm.
\end{enumerate}


\section{Problem Setting}
A sequential decision-making problem is formalized by a Markov Decision Process (MDP) defined by the tuple $\mathcal{M} = ( \mathcal{S}, \mathcal{A}, \mathcal{T}, r, \rho, \gamma)$, where $\mathcal{S}$ is the set of possible states of an environment, $\mathcal{A}$ is the set of actions that can be taken by the agent, $\mathcal{T} : \mathcal{S} \times \mathcal{A} \rightarrow \Delta \mathcal{S}$ is the transition function, $r: \mathcal{S} \times \mathcal{A} \rightarrow \mathbb{R}$ is the reward function, $\rho$ is the initial state distribution and $\gamma \in [0, 1)$ is the discount factor.
Let $V^{\pi}(s)$ be the expected sum of discounted rewards obtained by an agent following policy $\pi : \mathcal{S} \rightarrow \Delta \mathcal{A}$ from starting state $s$. The goal of reinforcement learning is to find a policy that maximizes $V^\pi$. In offline reinforcement learning, the agent has access only to a fixed set of transitions $\mathcal{D}$ and cannot interact with the environment to collect any additional data. This offline dataset is of the form
$$\mathcal{D} = \left\{ (s_i, a_i, s'_i, r_i, d_i)\right\}_{1 \leq i \leq N},$$
where $s'_i$ is a next state sample due to taking action $a$ at state $s_i$, $r_i$ is the resultant reward, and $d_i$ denotes whether it resulted in episode termination.

First, we explain the problem in a supervised learning setting. 
Here, given an unlabeled set of data points $\mathcal{D}_{\text{Unl}}$ and a small set of labeled points $\mathcal{D}_{\text{lab}}$, the agent has to choose which points in $\mathcal{D}_{\text{Unl}}$ to label, in the context of what is already available as labeled data.
More precisely, consider a supervised learning algorithm $\textsf{Alg}$ that takes a labeled dataset $\mathcal{D}_{\text{lab}}$ and returns a trained model $\pi_{\textsf{Alg}}(\mathcal{D}_{\text{lab}})$. Further, any model $\pi$ has to finally perform well on some unknown distribution, quantified by some performance measure $V^\pi$.
At each stage, the goal of the active learning agent is to choose a point $\tau \in \mathcal{D}_{\text{Unl}}$ to label so as to optimize $V$ of the model learned on this new dataset, i.e.,
\[
\tau^* = \argmax_{\tau \in \mathcal{D}_{\text{Unl}}} V^{\pi_{\textsf{Alg}} \left( \mathcal{D}_{\text{lab}} \cup \left\{ \tau \right\} \right)}.
\]


The aim here is to pose this problem in an MDP setting. 
Due to the sequential nature of the problem, the unlabeled data points in supervised learning correspond to unknown trajectories. Further, a trajectory (which translates to a data point in supervised learning) is not available to be chosen here but can only be observed as a stochastic function of some exploration strategy employed by the active learning agent in the context of constraints imposed by the environment. We formalize these aspects below. 

Consider the following MDP $\mathcal{M}_{\text{Act}} = (\mathcal{S}, \mathcal{A}, \mathcal{T}, r, \hat{\rho}, \gamma)$, with a different initial state distribution $\hat{\rho}$ for the active trajectory collection phase. At the start of each episode, a set of candidates initial states
$\mathcal{C} = \left\{ s_i \sim \hat{\rho} : 1 \leq i \leq N \right\}$
is sampled from $\hat{\rho}$. The agent can choose to start exploring the environment from any subset of these candidates and can also stop exploring at any time, if necessary, to preserve its budget.

Thus, the active learning agent $\mu = (\mathcal{I}, \pi, \beta)$ has three components: (i) an initial state selection function $\mathcal{I}: \mathcal{S} \rightarrow \mathbb{R}$ that decides the utility of collecting a trajectory from a given state, (ii) an exploration policy $\pi: \mathcal{S} \rightarrow \Delta \mathcal{A}$ that maps states to probability distributions over actions, and (iii) a termination function $\beta: \mathcal{S} \rightarrow \left\{ 0, 1 \right\}$ that decides whether to terminate the current trajectory.

Such an agent $\mu$ induces a distribution $\mathfrak{T}^\mu$ over trajectories due to the stochasticity of the initial state distribution and the exploration policy. The objective of the agent is to collect new samples within a budget in the context of the current dataset $\mathcal{D}$, so as to maximize the performance in the original MDP $\mathcal{M}$ of the policy $\pi_{\mathsf{Alg}}$ trained on this augmented dataset using the offline algorithm $\mathsf{Alg}$, i.e.,
\begin{align*}
\mu = \argmax_{\mu = (\mathcal{I}, \pi, \beta)} \mathsf{E}_{\tau \sim \mathfrak{T}^\mu} & \left[ V^{\pi_{\mathsf{Alg}} \left( \mathcal{D} \cup \left\{ \tau \right\} \right)} \right].
\end{align*}

For example, for the medical trials problem discussed in Section~\ref{Section: Introduction}, candidate initial states are generated by $\hat{\rho}$ corresponding to patients with varied medical history till that point. The agent should choose a subset of these based on the state and existing data.
Similarly, in the navigation setting, when a known state is reached while exploring the environment, the termination function $\beta$ stops the current trajectory so as to prevent the collection of redundant samples.
The procedure for active exploration is listed in Algorithm~\ref{alg:algorithm_active}. 

\section{Algorithm}
\label{Section: ActiveORL}

\begin{algorithm}[t]
\KwIn{$\mathcal{D}$: Offline dataset \\
\hspace{29pt}$\mathsf{Alg}$: Offline RL algorithm \\
\hspace{29pt}$B$: Interaction budget
}

\KwOut{$T = \left\{ \tau_t \right\}$: Trajectories actively collected from the MDP $\mathcal{M}_{\text{Act}}$}
	\textbf{Initialization:} $T \leftarrow \phi$ \;
	\While{$B > 0$}{
        \textsc{Learn} $\mu = (i, \pi, \beta)$ based on $\mathcal{D} \cup T$ \;
        $\mathcal{C} \sim \hat{\rho} \left( \mathcal{M}_{\text{Act}} \right) $\;
        $s_{\text{init}} \leftarrow \argmax_{1 \leq i \leq |\mathcal{C}|} \mathcal{I}(s_i)$\;
        \tcp{Collect trajectory from $s_{\text{init}}$}
        $s \leftarrow s_{\text{init}}$
        $\tau = \phi$\;
        \While{$B > 0 \text{\normalfont \text{ and} } \beta(s) = 0$}{
            $a \sim \pi(s)$\;
            $s' \sim \mathcal{T}(s, a)$\;
            $r = r(s, a)$\;
            $\tau \leftarrow \tau \cup \left\{ (s, a, s', r) \right\}$\;
            $B \leftarrow B - 1$\;
            $s \leftarrow s'$\;
        }
        $T \leftarrow T \cup \tau$\;
    }
    $\pi_{\mathsf{Alg}} \leftarrow \mathsf{Alg} \left( \mathcal{D} \cup T \right)$
\caption{Active Offline Reinforcement Learning}
\label{alg:algorithm_active}
\end{algorithm}

A practical implementation of actively collecting trajectories in addition to offline data and learning an optimal agent consists of two components: (i) The base offline reinforcement learning algorithm $\mathsf{Alg}$ that learns a policy given a dataset of transitions, and (ii) The active collection strategy $\mu = (\mathcal{I}, \pi, \beta)$.
In this work, we consider existing offline algorithms for the first component. For the second active component, the goal is to collect transitions that are diverse and underrepresented in the given offline dataset. To solve this, we propose an epistemic uncertainty-based method, where we learn an ensemble of representation models for encoding states and state-action pairs and use them to estimate the uncertainty of the agent in state and state-action space.

\subsection{Representation-based Uncertainty Estimation}
\label{Subsection: RepresentationModel}

We consider representation models of the form $\mathcal{E} = \left( \mathcal{E}^\mathbf{s}, \mathcal{E}^\mathbf{a} \right)$ to encode state and state-action representations, where $\mathcal{E}^\mathbf{s}$ and $\mathcal{E}^\mathbf{a}$ encode state and action information respectively. $\mathcal{E}$ has the following modes of operation: (a)
given a \textbf{state} $\mathbf{s}$ as input, the latent embedding is obtained by passing it through the state encoder, i.e., $\mathcal{E}(\mathbf{s}) \equiv \mathcal{E}^\mathbf{s}(\mathbf{s})$, and
	(b) given a \textbf{state-action} pair $(\mathbf{s},\mathbf{a})$ as input, the latent embedding of the state and action is added after passing the state and action through their respective encoders, i.e., $\mathcal{E}(\mathbf{s},\mathbf{a})\equiv\mathcal{E}^\mathbf{s}(\mathbf{s})+\mathcal{E}^\mathbf{a}(\mathbf{a})$.

We enforce the following two modeling objectives to align the latent representations learned by $\mathcal{E}$: (i) similar states (or observations) must be clustered in the latent space, and (ii) the embedding of a state-action pair must align with the latent representation of the corresponding next state.


Consider a transition tuple $(\mathbf{s}, \mathbf{a}, \mathbf{s}', \mathbf{r}, \mathbf{d}) \sim \mathcal{D}$.
To satisfy the clustering objective, we use $\mathbf{s}$ as the anchor sample, $\mathbf{s}'$ as the positive sample, and any other observation $\mathbf{s}'$ sampled from $\mathcal{D}$ is considered as the negative sample. Embedding vectors $\mathbf{v}$, $\mathbf{v}^+$, and $\mathbf{v}^-$ are obtained by passing $\mathbf{s}$, $\mathbf{s}'$, and $\mathbf{s}''$ respectively through the state encoder $\mathcal{E}^{\mathbf{s}}$. Moreover, to satisfy the transition dynamics modeling objective, we enforce the encoding $\mathbf{\hat{v}}^+ = \mathcal{E} (\mathbf{s}, \mathbf{a})$ to be close to $\mathbf{v}^+$.

We train an ensemble $\{\mathcal{M}_k\}_{k=1}^K$ of such models to maximize the following augmented noise-contrastive loss:
\begin{align*}
	\mathrm{L} = \log(\sigma(\mathbf{v} \cdot \mathbf{v}^+)) + \log(1 - \sigma(\mathbf{v} \cdot \mathbf{v}^-)) - \lambda || \mathbf{\hat{v}}^+ - \mathbf{v}^+ ||^2,
\end{align*}
where $\sigma(x) = 1/(1+\exp{(-x)})$ is the sigmoid function and $\lambda$ is a hyper-parameter.


We consider epistemic uncertainty for both initial state selection and exploration. A natural strategy to estimate the same is to use the ensemble $\{\mathcal{E}_k\}_{k=1}^K$ of state representation models and calculate the amount of dissimilarity among the latent representations of said models, i.e., disagreement, for a given state or state-action pair.

Let $\mathbf{v}_k$ denote the latent representation of a state or state-action pair by model $\mathcal{E}_k$. So, for a given state or state-action pair, we have $k$ vectors $\{\mathbf{v}_k\}_{k=1}^K$. We construct a similarity matrix $\mathbb{S}$ as follows:
\begin{align}
	\mathbb{S}_{i,j} = || \mathbf{v}_i - \mathbf{v}_j ||^2 \quad \mathrm{for} \; i,j =  1, 2, ..., K .
 \label{eqn:uncertainty}
\end{align}
We use the value of the largest element in $\mathbb{S}$  as our proxy for epistemic uncertainty of the model w.r.t the environment.

\subsection{Practical Implementation of Algorithm}
\label{Subsection: UnceratintyExploration}

In the first phase of our algorithm, we learn the representation models and an RL agent using just the offline dataset. The agent is learned using a suitable offline reinforcement learning algorithm, depending on the environment.

The second active exploration phase consists of two components: (i) Initial state selection and (ii) Trajectory collection from thereon.
Given candidate initial states $\mathcal{C} = \{ s_i \}$ from $\hat{\rho}$, those initial states are chosen that result in maximum uncertainty for the representation model ensemble $\{\mathcal{E}_k\}_{k=1}^K$, calculated as
\[
\text{Uncertainty}(s_i) = \max_{k, k'} \left| \left|\mathcal{E}^{\mathbf{s}}_k (s_i) - \mathcal{E}^{\mathbf{s}}_{k'} (s_i) \right| \right|^2.
\]

For trajectory collection starting at the chosen $s_k$'s, consider $\pi$ to be the current policy. At the beginning of the active phase, this is just the policy learned on the offline dataset with the appropriate offline RL algorithm. At each step, $M$ actions are sampled from $\pi( . | s)$ for current state $s$, resulting in $M$ state-action pairs. The policy $\pi$ could be deterministic, in which case a scaled isotropic Gaussian noise is added to $\pi(s)$ in order to sample multiple actions.

The uncertainty for each pair is calculated using Equation~\ref{eqn:uncertainty}, and the action resulting in the most uncertain state-action pair is chosen as the action to execute. This exploration strategy can be continued until the uncertainty of the current state falls below a certain threshold or the episode terminates.

The degree of exploration is controlled by an $\epsilon$-greedy variant of the exploration policy that explores using the aforementioned environment-aware uncertainty-based procedure with probability $\epsilon$, and simply follows the policy $\pi$ otherwise. The detailed algorithm is listed in the Appendix.

\subsection{Restricted initial states}

In certain environments, it might be infeasible to modify the initial state distribution. The environment could provide us with candidate initial states, albeit restricting us from starting directly at those states. To solve this problem, we propose a modified version of our algorithm in which the agent follows a two-stage policy during active collection. The first stage starts from the default initial state of the environment and goes to the optimal candidate initial state, from which the actual exploration can be done in the second stage. We train two separate policies for these two stages.

We train a goal-based policy on the offline dataset for the first stage.  We then create a weighted undirected graph $\mathcal{G}$ from the offline dataset, with each node corresponding to a state and the weight of edge $\{ s_i, s_j \}$ being $e^{- || s_i - s_j ||}$, and divide the nodes into clusters. The second stage policy is the uncertainty-based exploration policy described previously.

During active collection, the goal-based policy is first used to reach the cluster of states nearest to the identified optimal candidate state, i.e., the candidate state corresponding to the most uncertainty. From here, the agent begins uncertainty-based exploration as described in the previous subsection.

\section{Related Work}

\begin{figure*}[t]
    \centering
    \begin{minipage}{0.3\textwidth}
        \includegraphics[width=\textwidth]{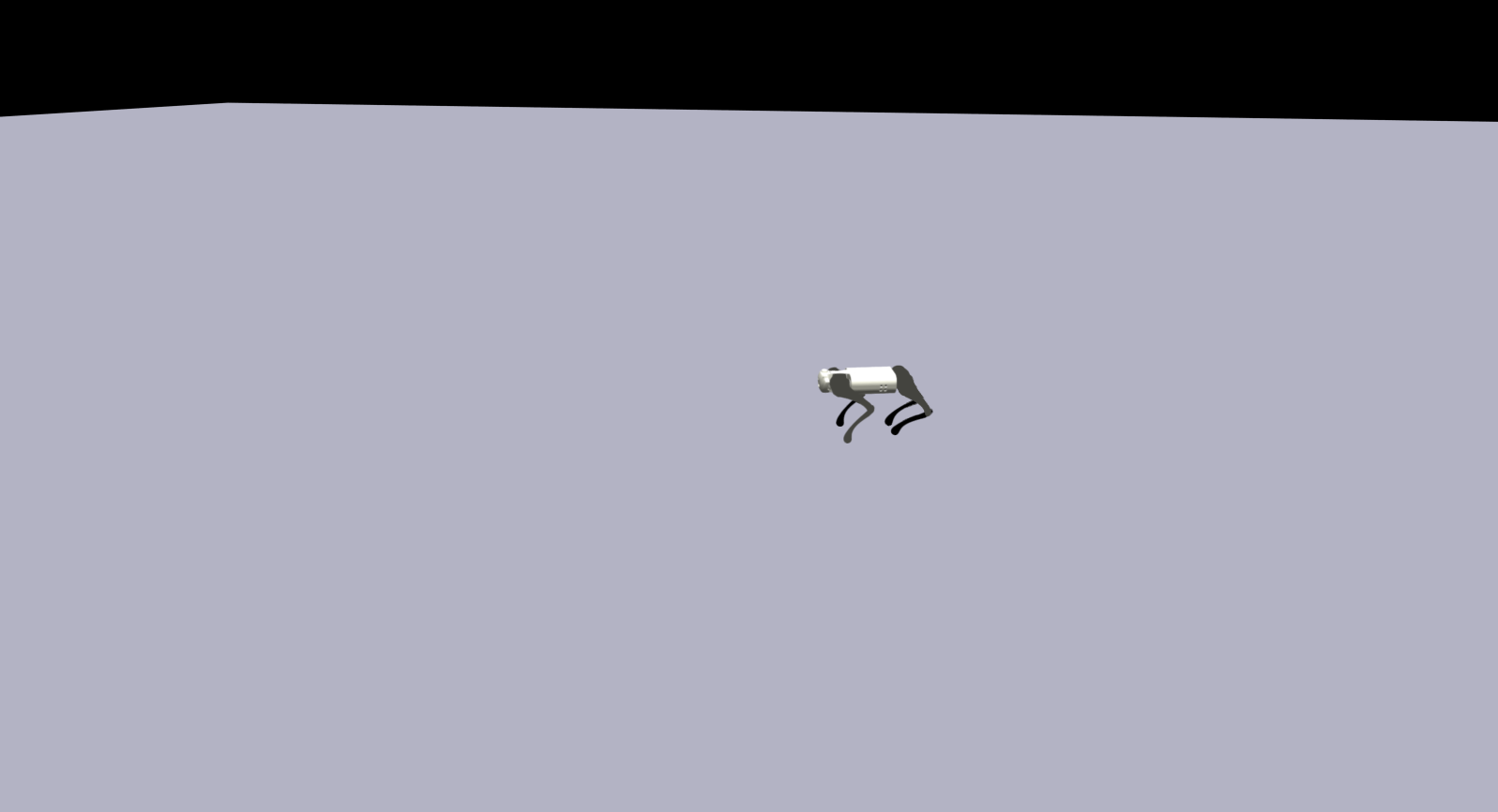}
    \end{minipage}
    \begin{minipage}{0.3\textwidth}
        \includegraphics[width=\textwidth]{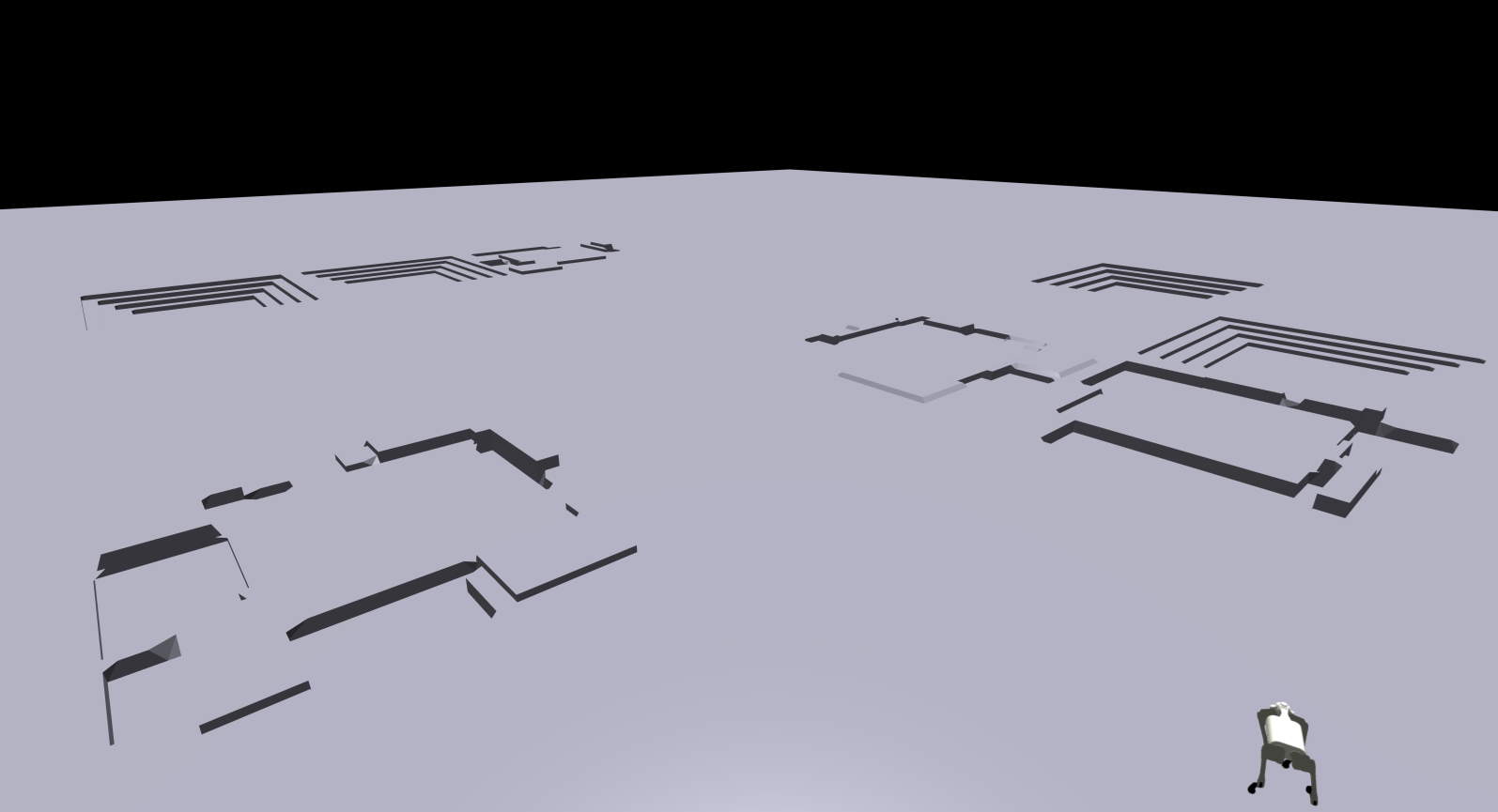}
    \end{minipage}
    \begin{minipage}{0.3\textwidth}
        \includegraphics[width=\textwidth]{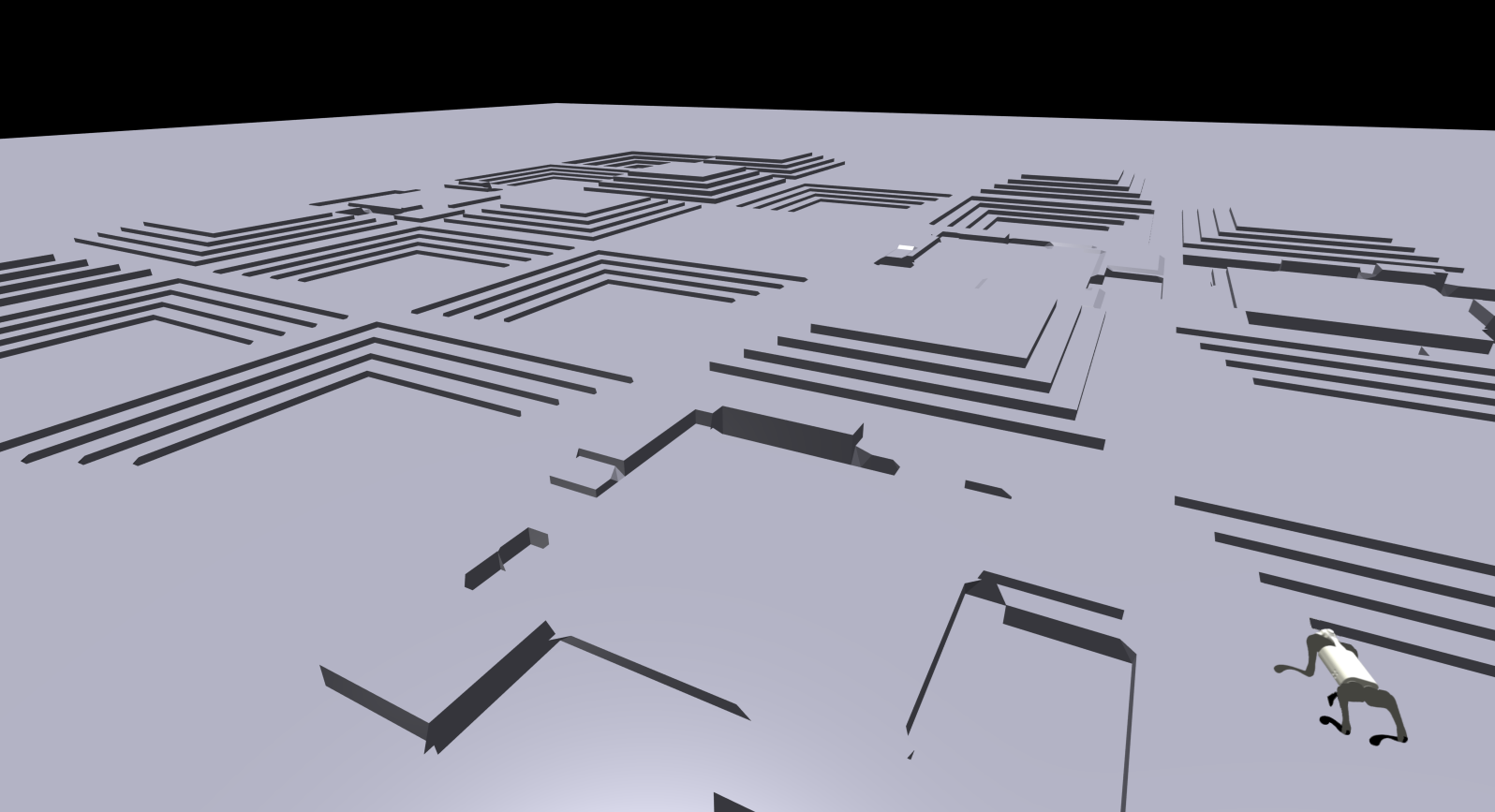}
    \end{minipage}
    \caption{The figures display the terrains for the Unitree Go1 robot experiments in the Nvidia Isaac Simulator. We named the three terrains (from left to right in order) \texttt{go1-easy}, \texttt{go1-medium} and \texttt{go1-hard}. The behavior policy was trained on the \texttt{go1-easy} terrain and achieves reasonably high rewards for the locomotion task on the flat surface, as shown. However, we assume that the environment has been modified, and the agent needs to update its policy as quickly as possible in the modified environment. If the agent efficiently uses its exploration budget, then it will be able to generalize the experiences gathered during Active Collection and be able to get high rewards in the \texttt{go1-hard} terrain in spite of being given access to \texttt{go1-medium} terrain during Active trajectory collection. The accompanying video in the supplementary materials demonstrates the advantage of using our active trajectory collection method.}
    \label{Figure: Go1Ablation}
\end{figure*}

\begin{table*}[t]
    \centering
    \begin{tabular}{l|rrrrrr}
        & BC & Offline & \begin{tabular}[c]{@{}c@{}}Offline\\+ FT\end{tabular} & \begin{tabular}[c]{@{}c@{}}Offline +\\RND\end{tabular} & \begin{tabular}[c]{@{}c@{}}Offline +\\AC (Ours)\end{tabular} & \begin{tabular}[c]{@{}c@{}}\%age of less\\interactions\end{tabular} \\
        \hline
        maze2d-medium-easy-v1 & -4.5 & -4.0  & 77.5 & 59.1 & \textbf{134.3} & 62.5\\
        maze2d-large-easy-v1 & 1.7 & -2.0 & 21.7 & 10.2 & \textbf{197.3} & 75\\
        maze2d-large-hard-v1 & -2.3 & -2.0 & 6.0 & 1.0 & \textbf{201.7} & 62.5\\
        \hline
        Maze-pruned total & -5.1 & -8.0 & 105.2 & 70.3 & \textbf{533.3} & \\
        \hline
        antmaze-umaze-v0 & 62.0 & 86.7 & 86.1 & 81.5 & \textbf{88.1} & 37.5\\
        antmaze-umaze-diverse-v0 & 54.0 & 56.0 & 43.9 & 39.2 & \textbf{71.6} & 50\\
        antmaze-medium-play-v0 & 0.0 & 59.0 & 68.9 & 56.8 & \textbf{73.1} & 37.5 \\
        antmaze-medium-diverse-v0 & 1.3 & 62.3 & 68.5 & 62.1 & \textbf{73.8} & 25 \\
        antmaze-large-play-v0 & 0.0 & 10.3 & 19.9 & 14.0 & \textbf{22.8} & 25\\
        antmaze-large-diverse-v0 & 0.0 & 9.0 & 19.8 & 9.9 & \textbf{22.9} & 37.5\\
        \hline
        AntMaze-subsampled total & 117.3 & 283.3 & 307.1 & 268.2 & \textbf{352.3} & \\
        \hline
        halfcheetah-random-v2 & 2.3 & 13.5 & 36.9 & 41.8 & \textbf{42.5} & 60\\
        hopper-random-v2 & 4.2 & 8.2 & 26.3 & 23.6 & \textbf{28.1} & 55.5\\
        walker2d-random-v2 & 2.0 & 7.9 & 9.1 & 10.8 & \textbf{11.4} & 33.3 \\
        halfcheetah-medium-v2 & 42.8 & 48.3 & 59.1 & 58.1 & \textbf{62.7} & 50\\
        hopper-medium-v2 & 54.0 & 68.1 & 93.4 & 88.4 & \textbf{96.7} & 37.5 \\
        walker2d-medium-v2 & 73.1 & 83.6 & 84.9 & 85.2 & \textbf{88.5} & -\\
        \hline
        Locomotion-subsampled total & 178.4 & 229.6 & 309.7 & 307.9 & \textbf{329.9} & \\
        \hline
        CARLA & 0.0 & 0.0 & 72.1 & 88.8 & \textbf{98.4} & 67 \\
        \hline
        unitree-go1-hard & 23.1 & 23.1 & 34.6 & 46.7 & \textbf{59.0} & 50 \\
        \hline
        \midrule
        Combined total & 313.7 & 528.0 & 828.7 & 781.9 & \textbf{1372.9} & \\
        \bottomrule
    \end{tabular}
    \caption{Results for our active method compared to respective baselines. Mean normalized scores (according to D4RL) are reported across various runs. As can be observed, we consistently improve the performance of the offline trained policy across multiple environments when compared to existing SOTA methods. We also observe a significant reduction in the number of samples required to reach the same performance as the baselines. (We denote inconclusive reduction by `-').}
    \label{Table: MainResults}
\end{table*}

The use of uncertainty-based methods for actively labeling data points has been studied in the context of supervised learning~\cite{balcan2006agnostic, gal2017deep}.

Similarly, in online reinforcement learning, successful methods for exploring MDPs typically rely on estimates of uncertainty about the Q-values (of state-action pairs) in order to encourage the agent to explore the environment \cite{NIPS2016_8d8818c8}. Some exploration strategies also rely on uncertainty-based intrinsic rewards or bonuses.
Popular approaches include indirect methods for uncertainty estimation such as approximate count \cite{10.5555/3157096.3157262}, random network distillation \cite{1810-12894}, and curiosity-driven exploration \cite{8014804}. \citet{mai2022sample} learn variance ensembles for capturing the uncertainty.

In offline reinforcement learning, both model-free and model-based methods incorporate uncertainty in different ways. MOPO~\cite{yu2020mopo} and MOREL~\cite{kidambi2020morel} are model-based methods in which the epistemic uncertainty of models learned on the offline dataset is explicitly used to induce pessimism in the trained policy. On the other hand, COMBO~\cite{yu2021combo} incorporates conservatism without explicitly estimating the uncertainty of the model.

In a model-free setting, UWAC~\cite{wu2021uncertainty} approximates the uncertainty through dropout variational inference. EDAC~\cite{an2021uncertainty} uses the variance of the gradients of an ensemble of $Q$ networks.

The work of \citet{yin2023sample} comes closest to our work in terms of application. However, their approach is applicable to the online setting and is primarily constrained to discrete control settings such as Atari 2600. Our approach differs in the following ways: (i) Unlike our approach, they use an ensemble of Q-Networks, and the variance across Q-values defines the uncertainty metric, (ii) they allow resetting to a previously observed state, and (iii) they sample actions from a uniform distribution and use local planning for exploration.

Active Offline Policy Selection~\cite{konyushova2021active} studies a related problem where the goal is to collect additional trajectories for evaluating a given set of policies and determining the best among them. In contrast, our method deals with collecting trajectories for the final goal of learning an optimal policy from the augmented dataset and not just evaluating given policies.

In Go-Explore algorithms~\cite{ecoffet2021goExplore}, the agent explores and comes back to already observed states to explore again, which does not work when the environment is largely unexplored. Our method, by contrast, assumes a given set of states and chooses the optimal states from which to start based on the available offline dataset.

\section{Experiments}
\label{Section: Experiments}

\subsection{Environments and Datasets}
\label{Subsection: Datasets}

We consider the following continuous control environments for evaluating our representation-aware uncertainty-based active exploration algorithm:
\begin{enumerate}
    \item \textbf{\texttt{Maze2d}}: The state is the 2D location on a plane of the agent in the form of a 2D ball. The objective is to navigate towards a goal by adjusting its velocity and direction.
    \item \textbf{\texttt{AntMaze}}: An extension to the \textbf{\texttt{Maze2d}} environment that includes a virtual ant agent instead that can be manipulated by controlling its joints.
    \item \textbf{\texttt{HalfCheetah}}, \textbf{\texttt{Hopper}}, and \textbf{\texttt{Walker2d}}: Locomotion environments in which the objective is to control a 2D stick figure with multiple joints to stably move forward.
    \item \textbf{\texttt{CARLA}}: A self-driving vehicle simulator wherein the agent has to control the acceleration and steering of a vehicle so as to stay in its lane and avoid collisions.
    \item \textbf{\texttt{IsaacSimGo1}}: A GPU-based simulator to control a legged $4\times3$ DOF quadrupedal robot using proprioceptive measurements along with ego-centric height information of the terrain.
\end{enumerate}
D4RL \cite{d4rl} is a collection of offline datasets for training and testing offline RL algorithms. To validate the performance of our active algorithm in the context of limited data, we prune these datasets and create new smaller versions.

We prune the medium and large \textbf{\texttt{Maze2d}} datasets by removing trajectories near the goal state. Figure~\ref{Figure: BlockDiagram} shows an example of a pruned dataset. Refer to the Appendix for visualization of all the pruned datasets in detail.

Additionally, we randomly subsample $30\%$ of the trajectories in the \textbf{\texttt{AntMaze}} datasets and the random and medium versions of the locomotion datasets.

For \textbf{\texttt{CARLA}}, we use a predefined expert policy to collect the offline dataset. We consider a roundabout with 4 exits. 8 starting "waypoints" are located equidistant from each other throughout the roundabout. The offline dataset is collected with 1 entry and 2 exits. However, the goal exit is not present in the offline dataset. The state space is augmented with the coordinates of the vehicle.

For the \textbf{\texttt{IsaacSim}} experiments, we use the \texttt{legged\_gym} API \cite{rudin2022learningwalkminutesusing} to simulate Unitree \textbf{\texttt{Go1}} robots. Initially, we used the default walking policy\footnote{The walking policy is described as Mode 2 in https://unitree-docs.readthedocs.io/en/latest/get\_started/Go1\_Edu.html}  for the physical robot to collect an offline dataset, which consists of trajectories on a flat surface. For the final evaluation, the robot is expected to walk on a complex terrain consisting of flights of stairs and discrete obstacles, shown in Figure~\ref{Figure: Go1Ablation}, requiring it to choose suitable starting locations during the active collection phase.

\subsection{Experimental Settings}
\label{Subsection: ExperimentalSettings}

\begin{figure}[!h]
    \centering
    \includegraphics[width=\linewidth]{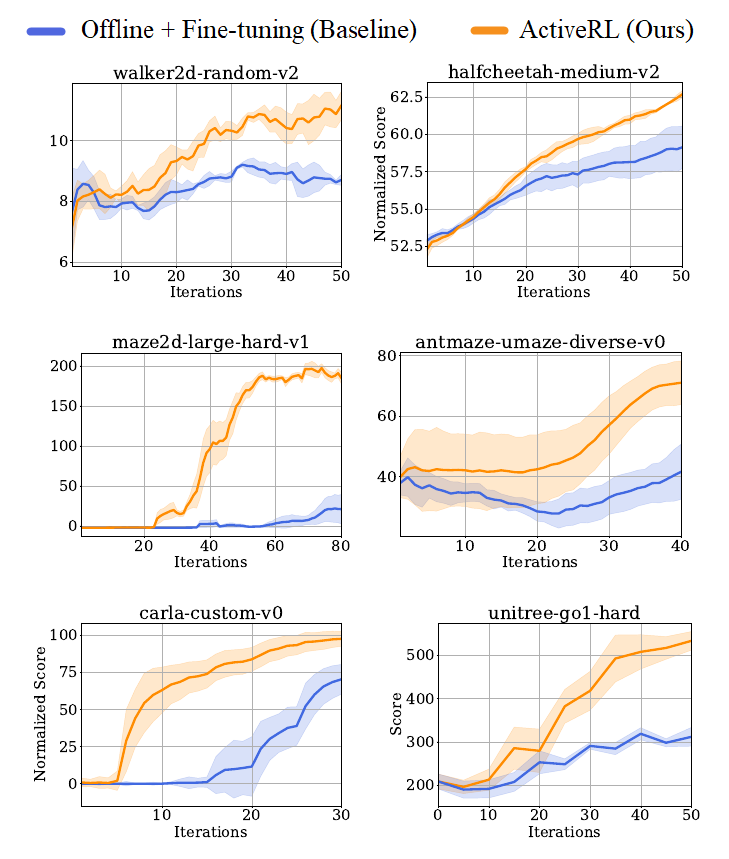}
    \caption{[Best viewed in color] Results of our algorithm compared with the corresponding fine-tuning baseline. In the shaded plots, the results are averaged over multiple random seeds, with the shaded region denoting the standard deviation.}
    \label{Figure: MainResults}
\end{figure}
For the offline phase of our algorithm, we use (i) TD3+BC~\cite{NEURIPS2021_a8166da0}, (ii) IQL~\cite{IQL}, (iii) CQL, and (iv) Behavior Cloning, as the base offline RL algorithms. We use TD3+BC in environments such as \texttt{maze2d-pruned} and locomotion, Behavior cloning for legged quadrupedal locomotion, and CQL and IQL in \textbf{\texttt{CARLA}} and \textbf{\texttt{AntMaze}} environments respectively, since TD3+BC does not work in environments where some notion of stitching is required.

In the next (active phase), the policy $\pi$ obtained offline is improved by using the data collected by our proposed active trajectory collection approach based on the uncertainty estimates induced by the representation model ensemble.


To validate the effectiveness of our proposed active collection, we compare our method with baselines that collect new trajectories without active initial state selection and active exploration. More precisely, the offline phase remains the same, wherein an offline RL algorithm is trained on the given dataset. In the second (fine-tuning) phase (denoted by FT in Table~\ref{Table: MainResults}), new trajectories are collected starting from the original initial state distribution $\rho$ of the MDP $\mathcal{M}$, using the learned offline policy $\pi$ as the exploration policy.

Specifically, for the \textbf{\texttt{Maze2d}} and locomotion environments, TD3+BC is used as the offline algorithm in the first phase. In the fine-tuning phase, the same training is continued on the newly collected data, with the $\alpha$ value being exponentially decayed to deal with the distribution shift~\cite{beeson2022improving}.

For the \textbf{\texttt{AntMaze}} and \textbf{\texttt{CARLA}} environments, IQL and CQL, respectively, are used directly for both the offline and online fine-tuning phases, as in~\citet{IQL, CQL}.

For legged locomotion, we use BPPO~\cite{zhuang2023behaviorproximalpolicyoptimization} as the offline policy learning algorithm in the active phase, which is perturbed for exploration based on the uncertainty models as described in Section~\ref{Section: ActiveORL}.

As an additional trajectory collection baseline for the active phase, we consider Random Network Distillation~\cite{1810-12894}, in which the offline learned policy is distilled into an ensemble of smaller networks and used for exploration.

The results are given in Table \ref{Table: MainResults} and Figure \ref{Figure: MainResults}. Along with the above baselines, we also report the performance of Behavior Cloning (BC) and the base offline algorithm without any additional data collection.

In the final column, we report the percentage of fewer additional interactions with the environment required by our algorithm to reach the best performance of the corresponding Offline + Fine-tuning baseline.

\begin{table}[t]
    \centering
    \begin{tabular}{c|cc}
        \toprule
        Algorithm & maze2d-large-easy & maze2d-large-hard \\
        \hline
        BC & 1.7 & -2.3 \\
        Offline & -2.0 & -2.0 \\
        \hline
        I+R & 45.5 & 25.0 \\
        I+P & 0.7 & 0.2 \\
        I+U & 51.1 & -1.5 \\
        \hline
        A+R & 88.1 & 74.6 \\
        A+P & 92.9 & 139.9 \\
        A+U & \textbf{133.8} & \textbf{176.3} \\
        \bottomrule
    \end{tabular}
    \caption{Ablation results to understand various components of our approach. For the initial state selection, `A' denotes active initial state selection, and `I' denotes usage of the unaltered initial states from the MDP. `R', `P', and `U' denote random policy, offline policy, and uncertainty-based exploration policy, respectively. Active initial state selection followed by an uncertainty-based exploration policy performs best. Further, A and U individually improve performance too.}
    \label{Table: Ablations}
\end{table}



\section{Results and Discussion}

\begin{figure*}
    \centering
    \includegraphics[width=0.8\textwidth]{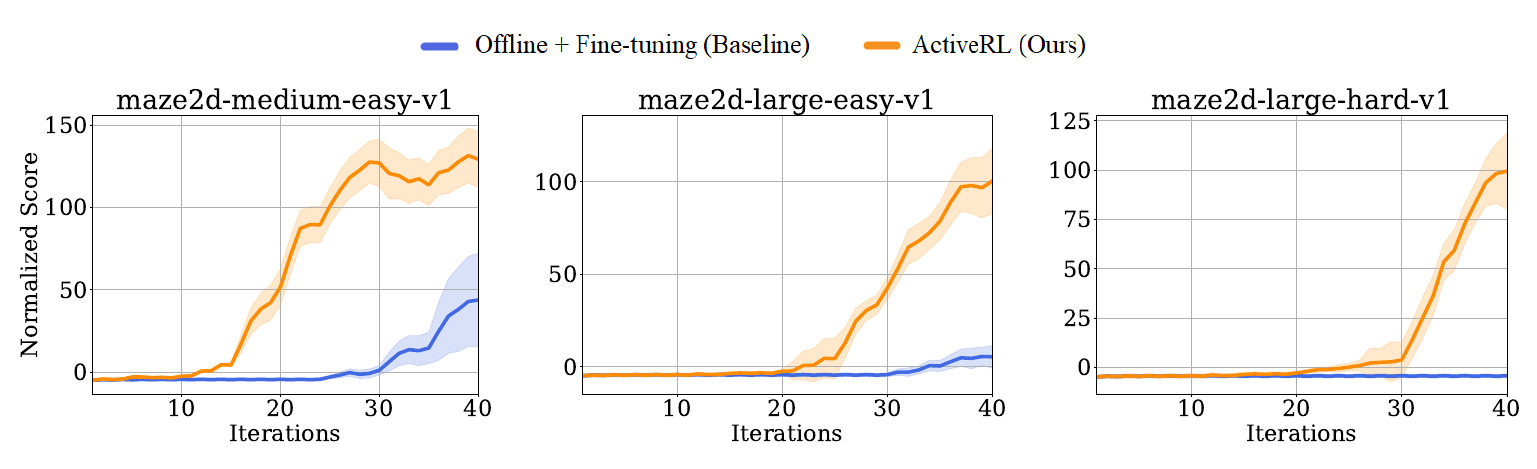}
    \caption{Plots corresponding to experiments where the agent is restricted to start from the original initial state distribution rather than the modified initial state distribution. We use a goal-based policy to reach a state close to the uncertain state and then switch over to our exploration policy.}
    \label{Figure: GoalAblation}
\end{figure*}

From Table \ref{Table: MainResults} and Figure~\ref{Figure: MainResults}, one can observe that across the various environments and corresponding datasets, our method demonstrates a significant advantage over the corresponding baselines, both in terms of the rewards obtained as well as the number of samples required to achieve a certain performance. In particular, one can see that
our method performs well in scenarios where the behavior policy is sub-optimal and has not learned to explore certain areas of the environment where better rewards are present. Hence, our method augments the offline dataset that does not have good coverage of the state space in a given MDP. For instance, in Table~\ref{Table: MainResults}, one can observe that our method achieves the most performance gain in the pruned \texttt{maze2d} datasets where certain regions are missing from the offline dataset.

Additionally, our active method was applied on top of  TD3+BC, IQL, and CQL, depending on the environment, showing that it is compatible with multiple offline algorithms.

\begin{table}[t]
    \centering
    \begin{tabular}{l|rrr}
        \toprule
        & TD3 & ActiveRL (ours) \\
        \hline
        maze2d-umaze-v1 & 142.79 & \textbf{164.8} \\
        maze2d-medium-v1 & 148.1 & \textbf{178.8} \\
        maze2d-large-v1 & 98.7 & \textbf{169} \\
        unitree-go1-hard & 43.9 & \textbf{52.8} \\
        \hline
        Total & 433.5 & \textbf{565.4} \\
        \bottomrule
    \end{tabular}
    \caption{Results for ablation with zero initial dataset (analogous to online). It is evident that (using TD3+BC as the base offline algorithm) we perform better than the corresponding online algorithm (TD3).}
    \label{Table: OnlineAblation}
\end{table}

\begin{table}[t]
    \centering
    \begin{tabular}{c|cccc}
        \toprule
         & var & mean & min & max \\
        \hline
        maze2d-medium-easy & 97.1 & 108.8 & 105.6 & \textbf{110.1} \\
        maze2d-large-easy & 111.3 & 122.5 & 123.9 & \textbf{133.8} \\
        maze2d-large-hard & 144.6 & 170.8 & 159.2 & \textbf{176.3} \\
        \bottomrule
    \end{tabular}
    \caption{Uncertainty metric ablation on \textbf{\texttt{Maze2d}}.}
    \label{Table: Uncertainty Ablation}
\end{table}

\paragraph{Ablations}
To verify the utility of active initial state selection, we perform an ablation by starting exploration only from the given initial state samples from $\rho$ of the original MDP $\mathcal{M}$. The results of this ablation are given in Table~\ref{Table: Ablations}.

To study the importance of a suitable exploration policy, we conduct an ablation by replacing our uncertainty-based exploration strategy (U) with random exploration (R) and exploration using the learned offline policy (P). The results are shown in Table \ref{Table: Ablations}.

From Table \ref{Table: Ablations}, we can clearly see that selecting initial states with our method provides an advantage irrespective of the exploration policy used. Conversely, our exploration policy by itself provides an advantage over random and naive strategies. This is true irrespective of how the initial states are chosen.

We also performed ablations to compare our uncertainty estimation technique with other variants. In our method, we take the maximum of the squared difference between estimates by different models in the ensemble. Table~\ref{Table: Uncertainty Ablation} shows the performance of our algorithm when this metric is replaced by the variance of the model estimates and the mean, minimum, and maximum of the squared differences, respectively.

Further, we studied the effectiveness of our algorithm for exploring the environment from scratch without any offline dataset. We skip the initial offline policy training step and start from a random policy, collect trajectories, and train on these experiences to simulate online learning. The results are shown in Table~\ref{Table: OnlineAblation}.

One can see that even in the absence of an initial offline dataset, our exploration strategy gains a significant advantage both in terms of samples used and final performance compared to the corresponding online algorithms.

\section{Conclusion}
By taking motivation from active learning, which is well-studied in supervised learning settings, we formulated this problem
for reinforcement learning in this paper. In the presence of an offline dataset, we proposed active learning strategies with which agents can acquire trajectories of agent-environment interactions to enhance their performance under a limited budget. 
 Our proposed approach consistently performs well across many environments and is compatible with multiple base offline RL algorithms. We compared the performance of our approach with strong baselines and performed ablation studies to understand the role of each component of our method.
\section*{Acknowledgements}
The authors would like to thank the SERB, Department of Science and Technology, Government of India, for the generous funding towards this work through the IMPRINT Project: IMP/2019/000383.

\bibliography{bibfile}

\input{appendix}

\end{document}

%% file: appendix.tex
\newpage

\appendix

\twocolumn[
\begin{center}
\textbf{\Large Active Reinforcement Learning Strategies for Offline Policy Improvement - Appendix}
\end{center}
\hfill \break
\hfill \break
]

\begin{table*}[h!]
    \centering
    \caption{Description of the notations used in the paper}
    \begin{tabular}{cl}
        \toprule
        \textbf{Notation} & \textbf{Description}  \\
        \midrule
        $\mathcal{S}$ & State space of the MDP \\
        $\mathcal{A}$ & Action space of the MDP \\
        $\mathcal{T}$ & Transition dynamics of the MDP \\
        $r$ & Reward function for a given state-action pair \\
        $\rho$ & Initial state distribution \\
        $\hat{\rho}$ & Distribution over candidate initial states \\
        $\gamma$ & Discount factor \\
        $\mathcal{D}$ & Offline dataset \\
        $\mathcal{G}$ & Graph created for the clustering the states \\
        $\pi$ & Policy \\
        $\pi^*$ & Optimal policy \\
        $\mathbb{E}$ & Expectation of a random variable \\
        $\mathbf{s}$ & State in state space $\mathcal{S}$ \\
        $\mathbf{a}$ & Action in action space $\mathcal{A}$ \\
        $\mathbf{r}$ & Reward obtained from a transition \\
        $\mathbf{d}$ & Represents whether the state obtained after transitioning to $\mathbf{s}^\prime$ is a terminal state.\\
        $\mathcal{C}$ & Set of candidate states obtained after sampling from $\hat{\rho}$ \\
        $\mathcal{E}$ & Representation model \\
        $\mathcal{E}^\mathbf{s}$ & State encoder of $\mathcal{E}$ \\
        $\mathcal{E}^\mathbf{a}$ & Action encoder of $\mathcal{E}$ \\
        $\mathbf{v}$ & Encoder output with $\mathbf{s}$ as the input \\
        $\mathbf{v}^+$ & Encoder output with $\mathbf{s}^\prime$ as the input \\
        $\mathbf{v}^-$ & Encoder output with $\mathbf{s}^{\prime\prime}$ as the input \\
        $\mathbf{\hat{v}}^+$ & Encoder output with the state-action pair $(\mathbf{s}, \mathbf{a})$ as input \\
        $\sigma$ & Sigmoid function \\
        $\mathbb{U}$ & Epistemic-uncertainty corresponding to a particular state or state-action pair \\
        $\mathbb{S}$ & $k\times k$ symmetric similarity matrix \\
        $\mathcal{N}(\mathbf{0}, \mathbf{I})$ & Gaussian distribution with zero mean and the identity matrix as the covariance \\
        $\lambda$ & Weight of the transition objective in the joint objective \\
        $\beta$ & The scale term to restrict the variance of $\mathcal{N}$\\
        $\phi$ & Null set \\
        $\sim$ & Operator that represents sampling from a distribution \\
        \bottomrule
    \end{tabular}
    \label{Table: Notations}
\end{table*}

\section{Offline RL Algorithms}

We use TD3+BC \cite{NEURIPS2021_a8166da0} and IQL \cite{IQL} as our base offline algorithm, on top of which we build our data collection method. Further, to allow fine-tuning of TD3+BC, we make use of policy relaxation as suggested by \cite{beeson2022improving}.

\textbf{TD3+BC} is a simple approach to offline reinforcement learning that combines the TD3 algorithm with behavior cloning. TD3 is a popular and efficient online algorithm for reinforcement learning, and behavior cloning is an imitation learning algorithm that can be used to learn to imitate a policy from a dataset of state-action pairs. TD3+BC works by first learning a Q-function from the dataset of state-action pairs. The Q-function is then used to train a deterministic actor with behavior cloning as a regularizer. Essentially, the policy is obtained by solving the following optimization problem:
\begin{equation*}
    \pi^* = \arg \max_\pi \mathbb{E}_{(\mathbf{s},\mathbf{a})\sim \mathcal{D}}\left[Q(\mathbf{s}, \pi(\mathbf{s})) - \alpha (\pi(\mathbf{s}) - \mathbf{a})^2\right]
\end{equation*}

\textbf{CQL} (Conservative Q-Learning) is an offline extension of the SAC algorithm that minimizes the following additional term for learning the critic along with the standard Bellman error:
\begin{equation*}
    \alpha \mathbb{E}_{s \in \mathcal{D}} \left[ \log \sum_{a} \exp \left( Q(s, a) \right) - \mathbb{E}_{a \sim \mathcal{D}} Q(s, a) \right].
\end{equation*}
This prevents overestimation of the Q-values at state-action values that are not in the distribution of the offline dataset, leading to a \emph{conservative} estimate of the Q-function.

For continuous action spaces, the summation over actions in the first term is replaced by an empirical average based on actions sampled from the uniform distributions and the current policy. The $\alpha$ term is updated via Lagrangian dual descent. The policy update is as in SAC. In our work, we used the d3rlpy implementation of this algorithm.

\textbf{IQL} works by first fitting an upper expectile value function to the dataset. The upper expectile value function estimates the value of the best available action at a given state. IQL then backs up the upper expectile value function into a Q-function. The Q-function is then used to extract a policy via advantage-weighted behavior cloning.

The policy estimation updates occur as follows:
\begin{equation*}
    L_V(\psi) = \mathbb{E}_{(\mathbf{s},\mathbf{a})\sim \mathcal{D}}\left[L_2^\tau(Q_{\hat{\theta}}(\mathbf{s},\mathbf{a}) - V_\psi(\mathbf{s}))\right]
\end{equation*}
\begin{equation*}
    L_Q(\theta) = \mathbb{E}_{(\mathbf{s},\mathbf{a},\mathbf{s}^\prime) \sim \mathcal{D}}\left[(r(\mathbf{s},\mathbf{a}) + \gamma V_\psi(\mathbf{s}^\prime) - Q_\theta(\mathbf{s},\mathbf{a}))^2\right]
\end{equation*}
where $L_2^\tau(u) = |\tau - \mathbb{I}(u < 0)|u^2$. Once the policy is estimated reasonably well, the policy is updated as:
\begin{equation*}
    L_\pi(\phi) = \mathbb{E}_{(\mathbf{s},\mathbf{a})\sim \mathcal{D}}\left[\exp (\beta(Q_{\hat{\theta}}(\mathbf{s},\mathbf{a}) - V_\psi(s)))\log_\phi(\mathbf{a}|\mathbf{s})\right]
\end{equation*}

\textbf{BPPO} (Behavior Proximal Policy Optimization) is an offline version of the PPO algorithm. Initially, a policy is learned to mimic the behavior policy using the given offline dataset. This policy is then updated just as in PPO, except that the transitions are sampled from the offline dataset, and the advantage function is estimated using a learned Q-function corresponding to the behavior policy instead of the observed rewards.

\section{Environments and Offline dataset}


We use the standard D4RL offline datasets for all of our experiments. All the modifications, namely subsampling and pruning, are done on top of these standard D4RL datasets.

\begin{figure*}[ht]
    \centering
    \includegraphics[width=\textwidth]{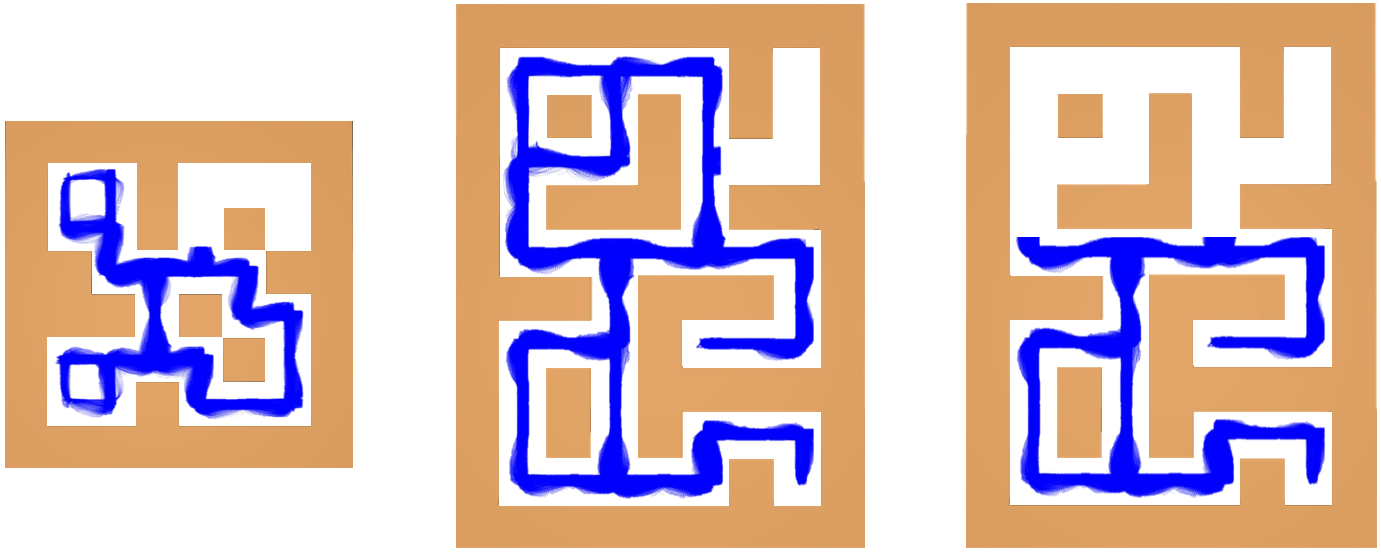}
    \caption{The pruned \texttt{maze2d} D4RL datasets. The first image (on the left) corresponds to the \texttt{maze2d-medium-v1} environment. We prune the dataset near the goal state to create \texttt{maze2d-medium-easy-v1}. The other two images correspond to versions of the \texttt{maze2d-large-v1} environment, an easy version \texttt{maze2d-large-easy-v1}, and a hard version \texttt{maze2d-large-hard-v1} respectively. The blue lines in the images correspond to the offline transitions remaining in the final dataset after pruning.}
    \label{Figure: PrunedDataset}
\end{figure*}

\begin{compactitem}
    \item [1.] \textbf{\texttt{maze2d}}: These environments involve the agent moving a force actuated ball in a maze-like environment along the X and Y axis to a fixed target location. The state space $\mathcal{S} \in \mathbb{R}^4$ and the action-space $\mathcal{A} \in \mathbb{R}^2$. These environments have varying levels of difficulties such as \texttt{maze2d-open-v1}, \texttt{maze2d-umaze-v1}, \texttt{maze2d-medium-v1} and \texttt{maze2d-large-v1}. The agent gets a positive reward when close to the target location and zero everywhere else. (There are also dense-reward versions of these environments, but they make the problem even simpler).
    
    We only use the \texttt{medium} and \texttt{large} variants as \texttt{umaze} and \texttt{open} are exceedingly simple.
    We increase the difficulty and simultaneously leave some room for exploration in the active trajectory collection phase by pruning the datasets for these (\texttt{medium} and \texttt{large}) datasets. We remove transitions close to the target location from the offline dataset as seen in Figure \ref{Figure: PrunedDataset} with varying levels of difficulty.

    \item [2.] \textbf{\texttt{antmaze}}: The setting is similar to that of \texttt{maze2d} where the agent is replaced by an ``Ant'' from OpenAI Gym MuJoCo benchmarks. The state-space, $\mathcal{S} \in \mathbb{R}^{29}$ and action-space, $\mathcal{A} \in \mathbb{R}^7$. Here, we subsample $30\%$ of the trajectories from the offline dataset and use the same as the initial offline dataset.

    \item [3.] \textbf{\texttt{locomotion}}: These are the standard OpenAI Gym MuJoCo tasks. We included results for \texttt{HalfCheetah-v2}, \texttt{Hopper-v2} and \texttt{Walker2d-v2}. For all of these environments, we used the \texttt{random} and \texttt{medium} datasets. Similar to \texttt{antmaze}, we subsample trajectories uniformly at random $30\%$ of the offline dataset.

    \item [4.] \textbf{\texttt{CARLA}}: We run an expert deterministic policy\footnote{As described in \url{https://vladlen.info/papers/carla.pdf}} to collect the offline trajectories with a cumulative $500$K transitions.

    \item [5.] \textbf{\texttt{IsaacSim-Go1}}: We use an offline dataset collected from an Unitree Go1 physical robot. The state space was $45$-dimensional, and we had $500$K total transitions in the offline trajectory dataset. We adapt it to the IsaacSim where the size of the state space is $48$, and the ego-centric height observation space is $187$, totaling to a total of $235$-dimensional observation space. Since the offline dataset was collected on flat regions, the $187$-dimensional height observations were almost constant and were easy to adapt. We estimated the $3$ velocities along the $x, y, z$ coordinates from the offline dataset itself.

    We consider three terrains, \texttt{go1-easy}, \texttt{go1-medium} and \texttt{go1-hard}. The behavior policy was trained on the \texttt{go1-easy} terrain and achieves reasonably high rewards for the locomotion task on the flat surface. We then assume that the environment has been modified and the agent needs to update its policy as quickly as possible in the modified environment. We call this the \texttt{go1-medium} terrain where on a grid of size $49 (7\times7)$, $5$ cells are upward pyramids with steps, $5$ cells are downward pyramids with steps, $5$ cells are uneven terrain (discrete obstacles), and the remaining $34$ cells are still flat (plain). The policy already trained on the flat regions can gain more information while adapting to the new terrain by exploring the unexplored obstacles ($30\%$ of the terrain). Finally, the \texttt{go1-hard} has only $5$ cells of flat terrain whereas the remaining comprise of $19$ upward pyramids, $19$ downward pyramids, and finally $6$ uneven terrain. The agent needs to generalize the experiences gathered during Active Collection and be able to get high rewards in the \texttt{go1-hard} terrain in spite of being given access to \texttt{go1-medium} terrain during Active trajectory collection. The accompanying video in the supplementary materials demonstrates the advantage of using our active trajectory collection method.
    
\end{compactitem}

We determine the set of valid \textit{initial states} of the environment as the full set of states of the offline dataset corresponding to each of the above-mentioned environments. The environment provides the agent a subset of these states as candidate states from where to start the trajectories.

\begin{algorithm*}[t]
	\caption{A pseudo-code for implementing the practical algorithm corresponding to our proposed approach}
	\label{Algo: ActiveORL}
	\KwData{$\mathcal{D}$: Offline dataset}
    \KwIn{Interaction budget \texttt{Budget}}
	Train $\pi$ and $\{\mathcal{E}_k\}_{k=1}^K$ using $\mathcal{D}$ \Comment*[r]{Train offline policy \& representation models}
	\While{$\text{\normalfont \texttt{Budget}} > 0$}{
		$\mathrm{Buffer} \leftarrow \phi$\Comment*[r]{Initialize an empty buffer}
		\While{$\mathrm{Buffer}$ is not full}
		{
			$\mathbf{d} \leftarrow \texttt{False}$\; Sample $\mathcal{C}$ from $\hat{\rho}$\Comment*[r]{Get a set of candidate initial states}
			$\mathbb{U}_c \leftarrow \texttt{Uncertainty}(\mathbf{s}_c)$ where $c\in\{1,...,|\mathcal{C}|\}$\Comment*[r]{Calculate uncertainties}
			$\mathbf{idx} \leftarrow \argmax_{1 \leq c \leq |\mathcal{C}|} U_c$\;
            $\mathbf{s} \leftarrow \mathcal{C}_\mathbf{idx}$\Comment*[r]{Most uncertain starting state}
            \Comment{Explore uncertain regions}
			\While{$\text{\normalfont \texttt{Uncertainty}}(s) \geq \text{\normalfont \texttt{threshold}} \text{\normalfont \textbf{ and }} \text{\normalfont $\mathbf{d}$} \text{ \normalfont is not } \text{\normalfont \texttt{True}}$}
			{
                    $\mathbf{a} \leftarrow \pi(s)$\Comment*[r]{Get action from offline policy}
                    \Comment{Explore with probability $\epsilon$}
                    \If{$\epsilon \leq \text{\normalfont\texttt{Uniform}}(0,1)$} 
                    {
                        Sample $\mathbf{M}$ actions as $\mathbf{a}_m\leftarrow\mathbf{a}+\beta\;\mathcal{N}(\mathbf{0}, \mathbf{I}) \; \text{ for } m\in\{1,...,\mathbf{M}\}$\;
                        $\mathbb{U}_m \leftarrow \texttt{Uncertainty}(\mathbf{s}, \mathbf{a}_m) \; \forall m\in\{1,...,\mathbf{M}\}$\Comment*[r]{Get uncertainties}
				        $\mathbf{idx} \leftarrow \argmax_{m  \in \{ 1, \dots, M \}} \mathbb{U}_{m}$\;
                        $\mathbf{a} \leftarrow \mathbf{a}_\mathbf{idx}$\;
                    }
				$\mathbf{s}', \mathbf{r}, \mathbf{d} \leftarrow \texttt{env.step}(\mathbf{s}, \mathbf{a})$\;
				$\mathrm{Buffer} \leftarrow \mathrm{Buffer} \cup \{(\mathbf{s}, \mathbf{a}, \mathbf{s}', \mathbf{r}, \mathbf{d})\}$\Comment*[r]{Add transition to buffer}
                    $\mathbf{s} \leftarrow \mathbf{s}'$\Comment*[r]{Update current state to next state}
			}
		}
		$\mathcal{D} \leftarrow \mathcal{D} \cup \mathrm{Buffer}$ \Comment*[r]{Append buffer to dataset}
		Update $\pi$ and $\{\mathcal{E}_k\}_{k=1}^K$ using augmented dataset $\mathcal{D}$ \Comment*[r]{Update policy \& models}
		$\text{\normalfont \texttt{Budget}} \leftarrow \text{\normalfont \texttt{Budget}} - |\mathrm{Buffer}|$ \Comment*[r]{Update remaining budget}
	}
	\textbf{Return}: $\pi$ \Comment*[r]{Return Improved Policy}
\end{algorithm*}

\section{Uncertainty Model}

\subsection{Representation model}
For our purpose, we use $K=5$ representation models without any weight sharing to compute the epistemic uncertainty for each of the datasets for the corresponding environment. We use $2$ affine layers for encoding the state and action representations. For \texttt{maze2d}, \texttt{locomotion} and \texttt{CARLA} environments we use a hidden dimension of $256$, while for \texttt{antmaze} and \texttt{IsaacSim-Go1} we increase it to $512$. We set $\lambda=1.0$ for all our experiments.

\subsection{$\epsilon$-greedy uncertainty-based exploration policy}
In the case of algorithms such as TD3+BC, which has a deterministic actor that produces the same action for the same state, we cannot ``sample'' actions for a particular state. To remedy this, we add $\beta\;\mathcal{N}(\mathbf{0}, \mathbf{I})$ to get a candidate set of actions from which we pick the action that maximizes our uncertainty metric. For stochastic actor algorithms like IQL, we do not require the noise to sample actions, as we find that the underlying stochasticity of the policy is enough.

The uncertainty based exploration policy is used in an $\epsilon$-greedy setup, where with probability $\epsilon$ we select our uncertain action, otherwise we select the policy action. We tried combinations of $\epsilon \in \{0.1, 0.15, 0.2, 0.25, 0.5\}$ and $\beta \in \{0.1, 0.2, 0.3, 0.5\}$. For \texttt{locomotion} environments, we find $\beta=0.1$ and $\epsilon=0.25$ to work best, while for simpler \texttt{maze2d} environments, we find larger value $\beta=0.3$ and $\epsilon = 0.5$ work better. In \texttt{antmaze} environments where we use IQL, we do not add the noise and keep $\beta = 0$ with $\epsilon = 0.3$.

\section{Details of Training}

\textbf{Libraries used}: We use the following programming languages and libraries.
\begin{compactitem}
    \item [1.] PyTorch 1.12.1 \cite{NEURIPS2019_9015}
    \item [2.] Gym 0.21.0 \cite{brockman2016openai}
    \item [3.] Mujoco 2.3.2 \cite{mujoco} and mujoco-py 2.1.2.14
    \item [4.] D4RL 1.1 \cite{d4rl}
    \item [5.] d3rlpy \cite{d3rlpy}
\end{compactitem}

\begin{figure*}[t]
    \centering
    \begin{minipage}{0.33\textwidth}
        \includegraphics[width=\textwidth]{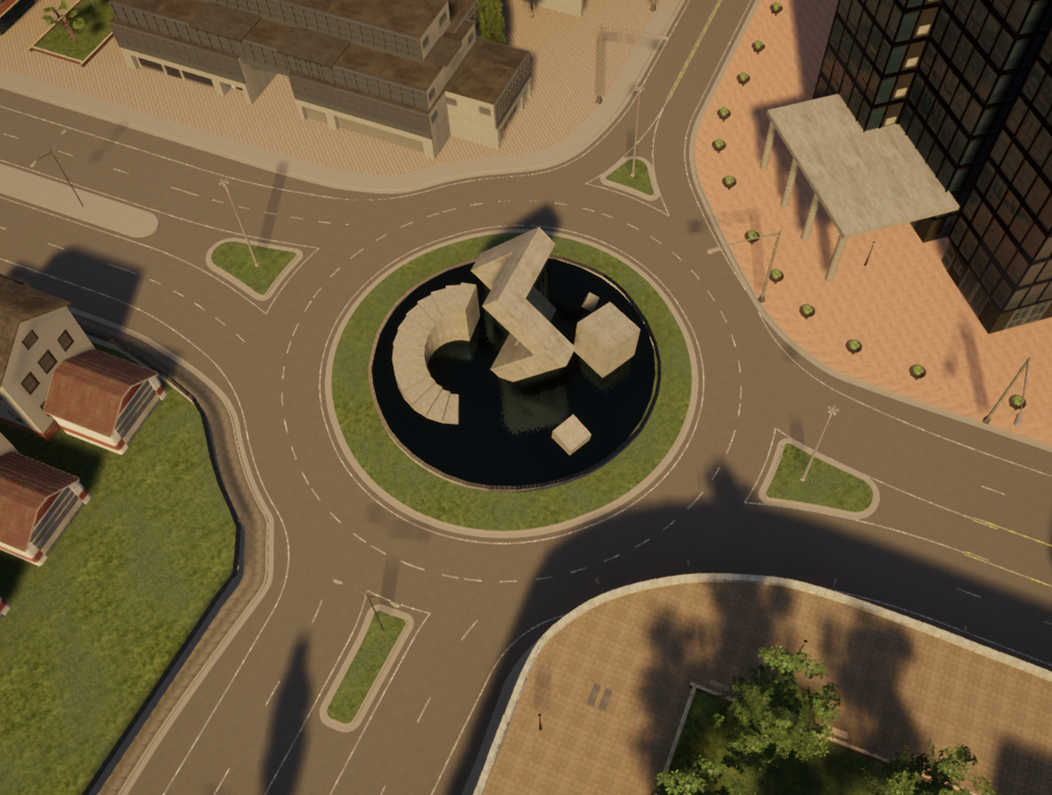}
    \end{minipage}
    \begin{minipage}{0.33\textwidth}
        \includegraphics[width=\textwidth]{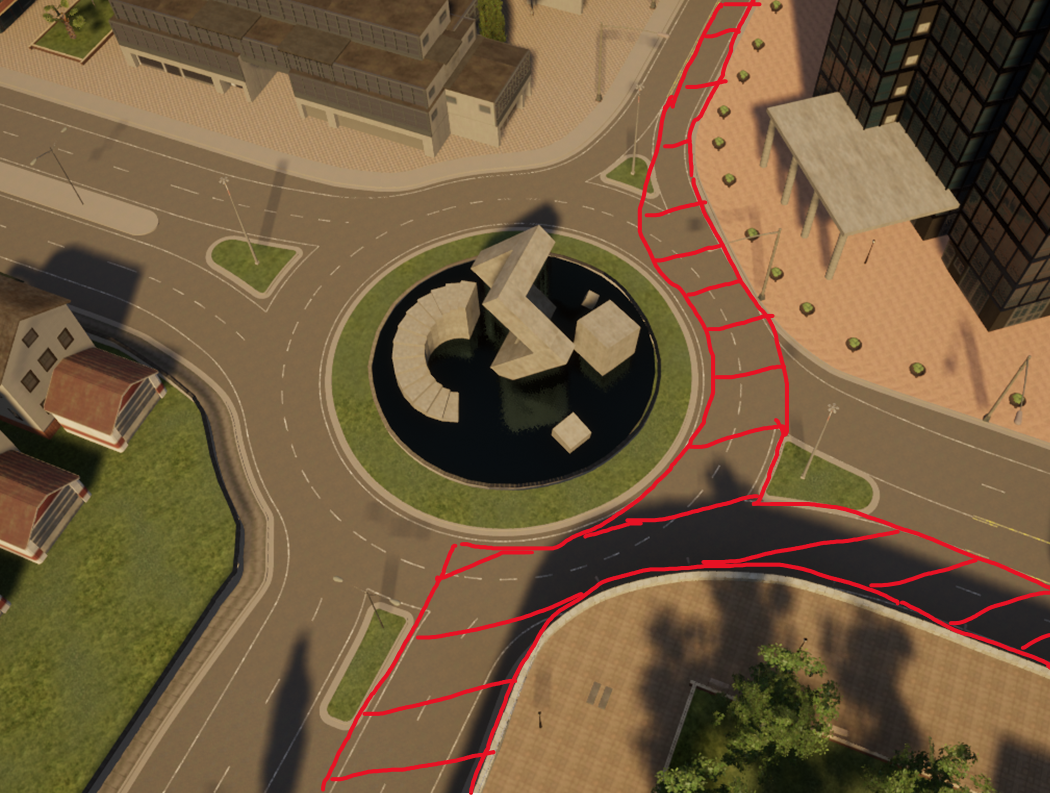}
    \end{minipage}
    \begin{minipage}{0.33\textwidth}
        \includegraphics[width=\textwidth]{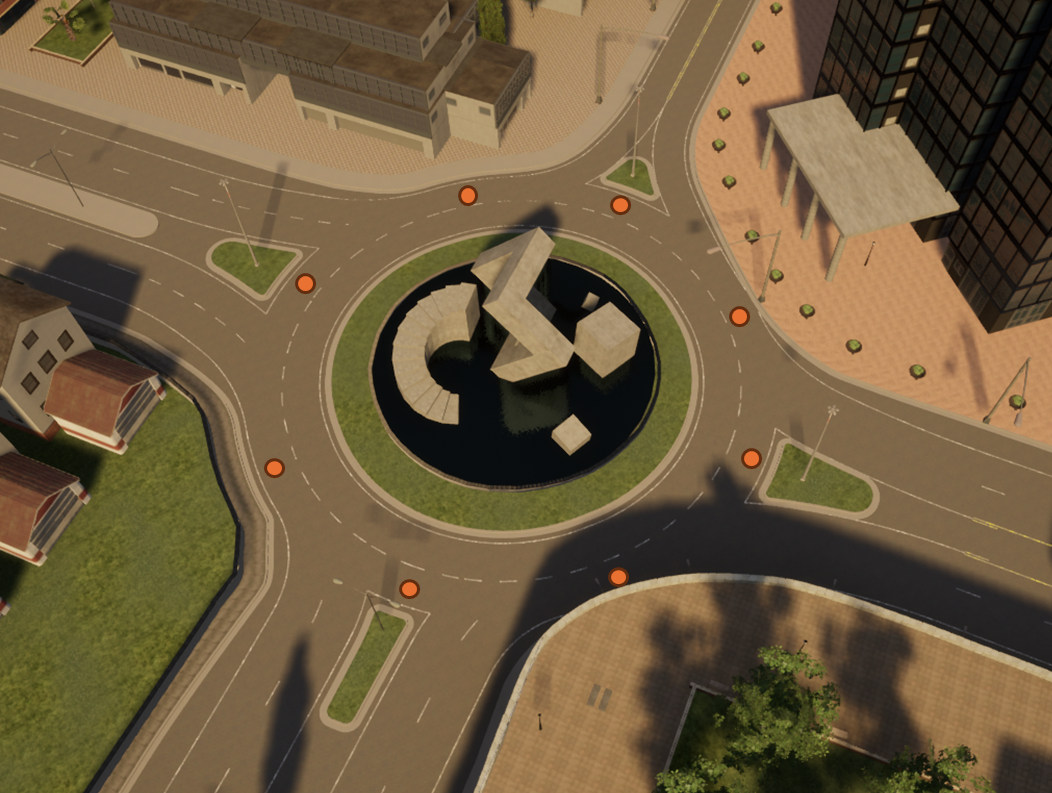}
    \end{minipage}
    \caption{[Best viewed in color] The figures depict the 4-way roundabout mentioned in the CARLA experiments. The image on the left depicts the junction in detail. The car enters from the south and has three choices for exit (east, north, and west). The rewards for the east exit is $+1$ each whereas the west exit has $+5$ reward and is consequently the goal exit. The image in the center depicts the region where the trajectory was collected using the behavior policy (i.e., the car entering from the south and going through either the east or the north exit). By following the behavior policies, it can be noted that the car will always take the north or the east exit. Only by exploring can the updated policy learn to use the goal exit. The image on the right depicts the $8$ candidate starting locations for the car. The starting locations already present inside the offline dataset will have low epistemic uncertainty, and the remaining will have high epistemic uncertainty and should, therefore, be explored.}
    \label{Figure: CarlaAblation}
\end{figure*}

\subsection{Main Experiments}
\textbf{BC}: We use d3rlpy's implementation of BC for running the experiments on all environments. The hidden size of the neural network was set to be $256$, and we updated the network for $30$ thousand gradient steps with a batch size of $1024$. We evaluated every $5000$ step and reported the best-achieved performance.

\textbf{Offline}: We use TD3+BC for \texttt{maze2d} and \texttt{locomotion} tasks, while for \texttt{antmaze} tasks we use IQL. We train offline for a total of $1$ million gradient updates. We use the default author-suggested hyperparameters for each of the environments for this training.

\textbf{Offline+FT} and \textbf{ActiveRL}: We define one epoch of Offline+FT or ActiveRL as collecting $X$ transitions from the environment followed by training for $Y$ updates on the newly augmented dataset. We use $X=5000$ and $Y=25000$ when our base algorithm is TD3+BC while we use $X=5000$ and $Y=50000$ when our base algorithm is IQL.

For all of the \texttt{antmaze} tasks we run $8$ epochs, for \texttt{locomotion} tasks we run $10$ epochs. For the pruned versions of \texttt{maze2d-large-v1} we run $16$ epochs and for the smaller \texttt{maze2d-medium-v1} versions we run $8$ epochs.

In cases where our base algorithm is TD3+BC, we exponentially decay $\alpha$ by a factor of $5.0$ across the epochs.

\subsubsection{Online Experiments}

\textbf{TD3}. We run the native TD3 online learning algorithm on an environment where there is no offline dataset to start with. We follow the standard online training procedure of collecting one transition and updating it once.

\textbf{ActiveRL}. Here, we also run in a fully online setting, starting with no offline dataset. For this purpose, we set $\alpha=0$ and ActiveRL for the data-collection as usual. Here, we collect one transition and train for $5$ updates.

We report the results after $300$k environment steps with evaluation every $5000$ steps.

The experiments have been run on a DGX station with $4 \times 32$ GB NVIDIA-V100 GPUs, Intel Xeon 40 core CPU, and 256 GB RAM. We repeat all the experiments for $K=5$ random initial seeds with the seeds being $\{0, 1, ..., K-1\}$.


\begin{figure*}[!t]
    \centering

    \begin{minipage}{\textwidth}
        \centering
        \includegraphics[height=0.35cm]{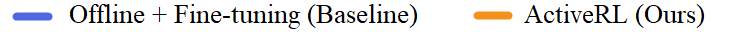}
    \end{minipage}
    
    \begin{minipage}{.31\textwidth}
        \centering
        \includegraphics[width=\textwidth, height=0.8\textwidth]{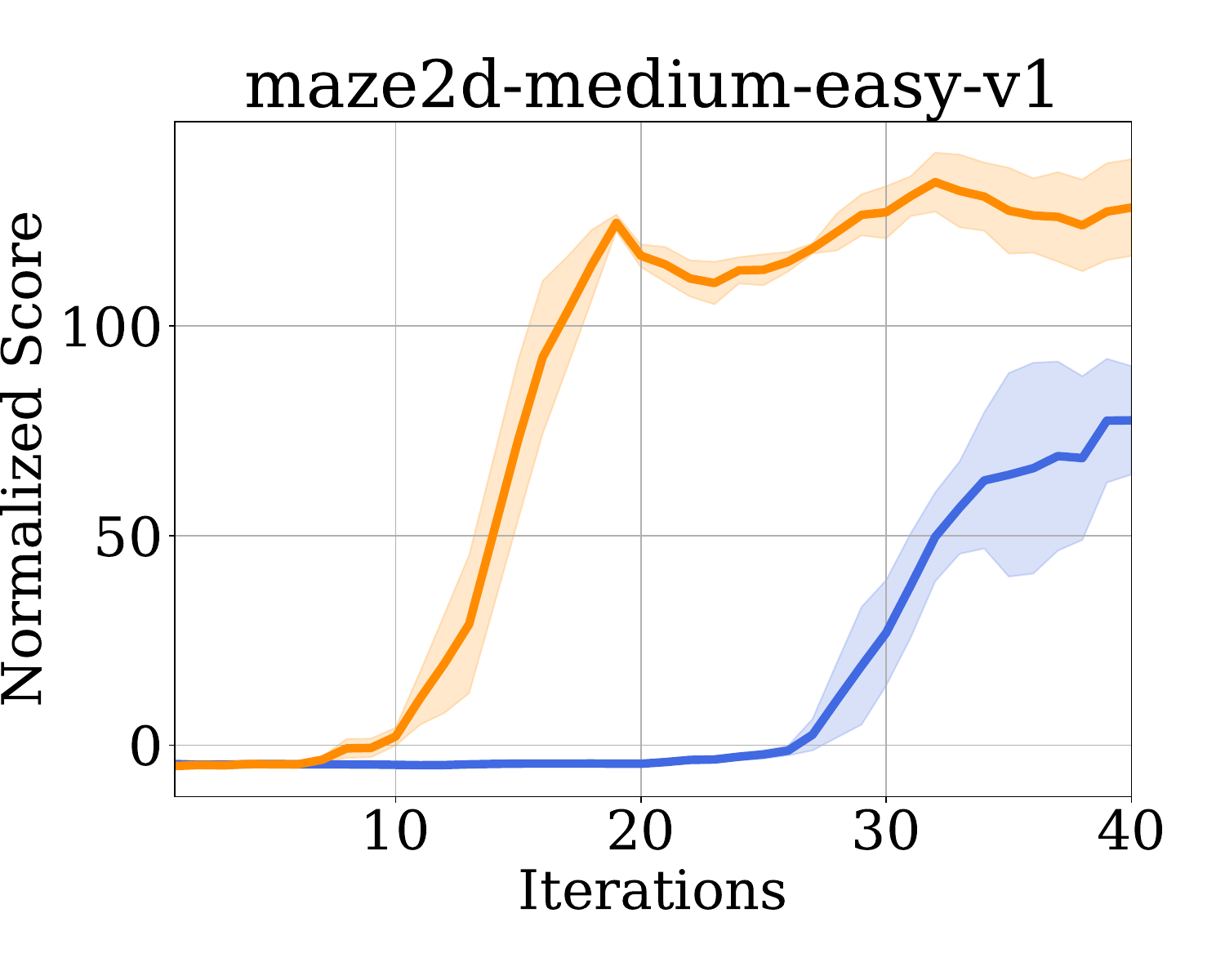}
    \end{minipage}
    \begin{minipage}{.31\textwidth}
        \centering
        \includegraphics[width=\textwidth, height=0.8\textwidth]{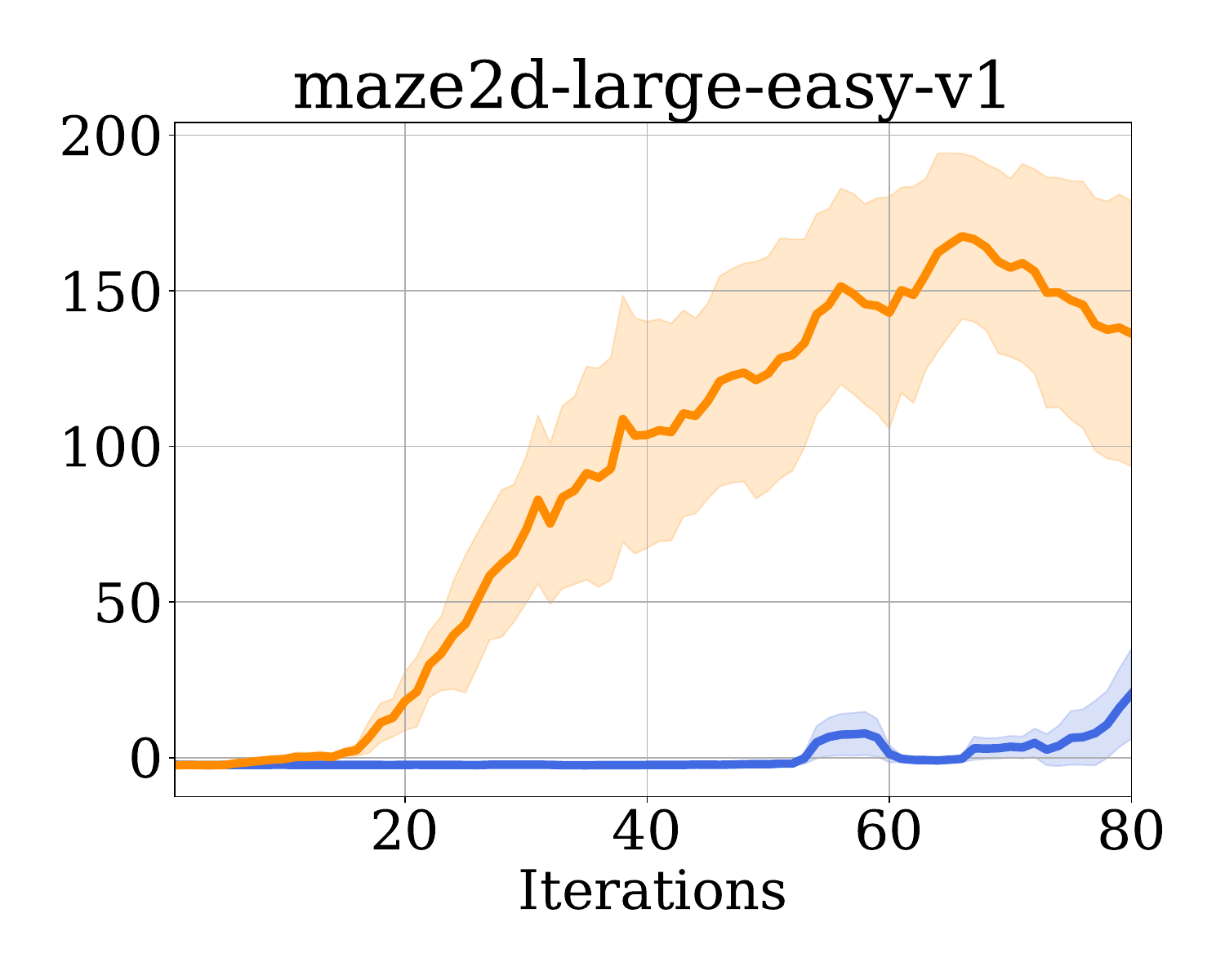}
    \end{minipage}
    \begin{minipage}{.31\textwidth}
        \centering
        \includegraphics[width=\textwidth, height=0.8\textwidth]{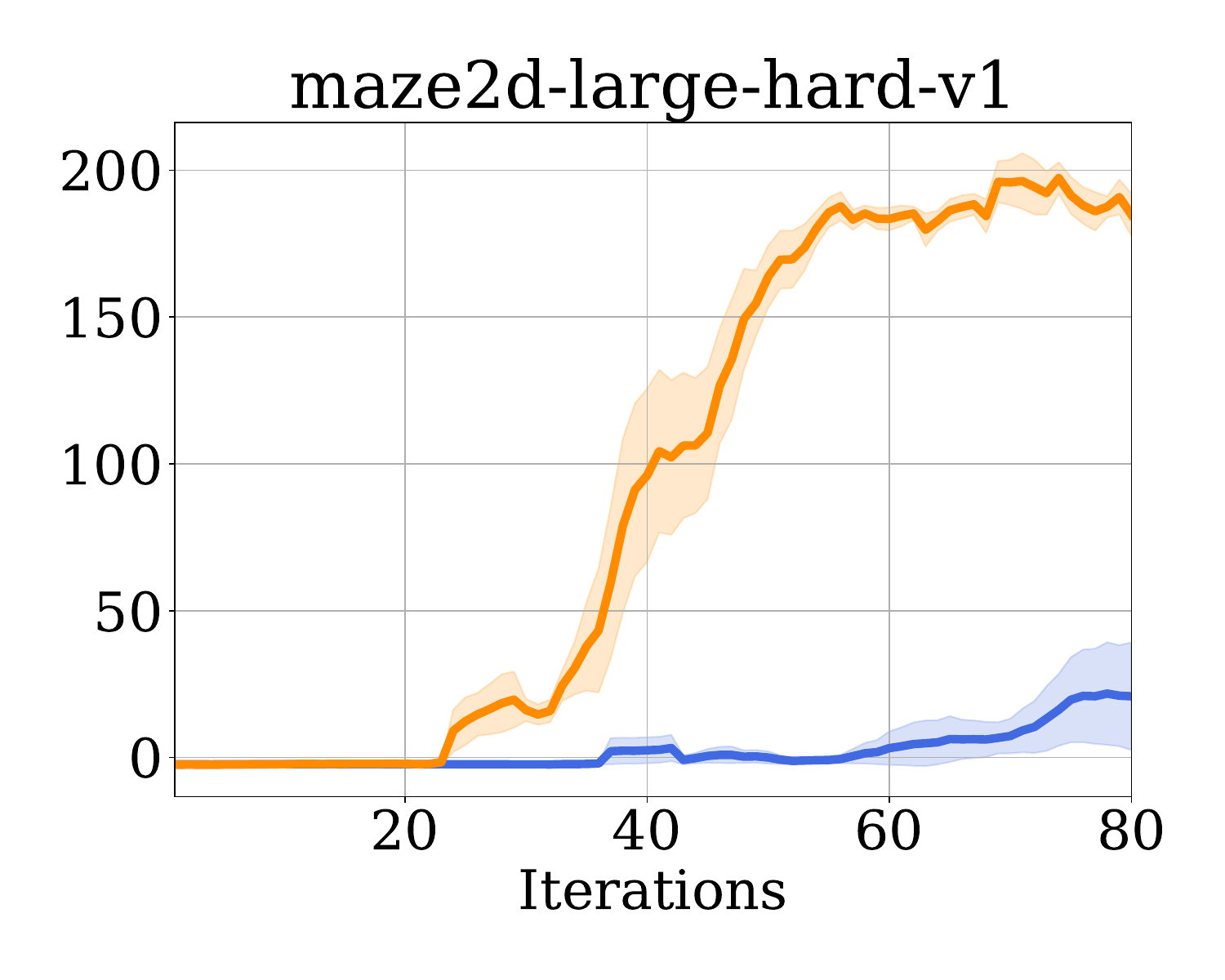}
    \end{minipage}
    
    
    \begin{minipage}{.31\textwidth}
        \centering
        \includegraphics[width=\textwidth, height=0.8\textwidth]{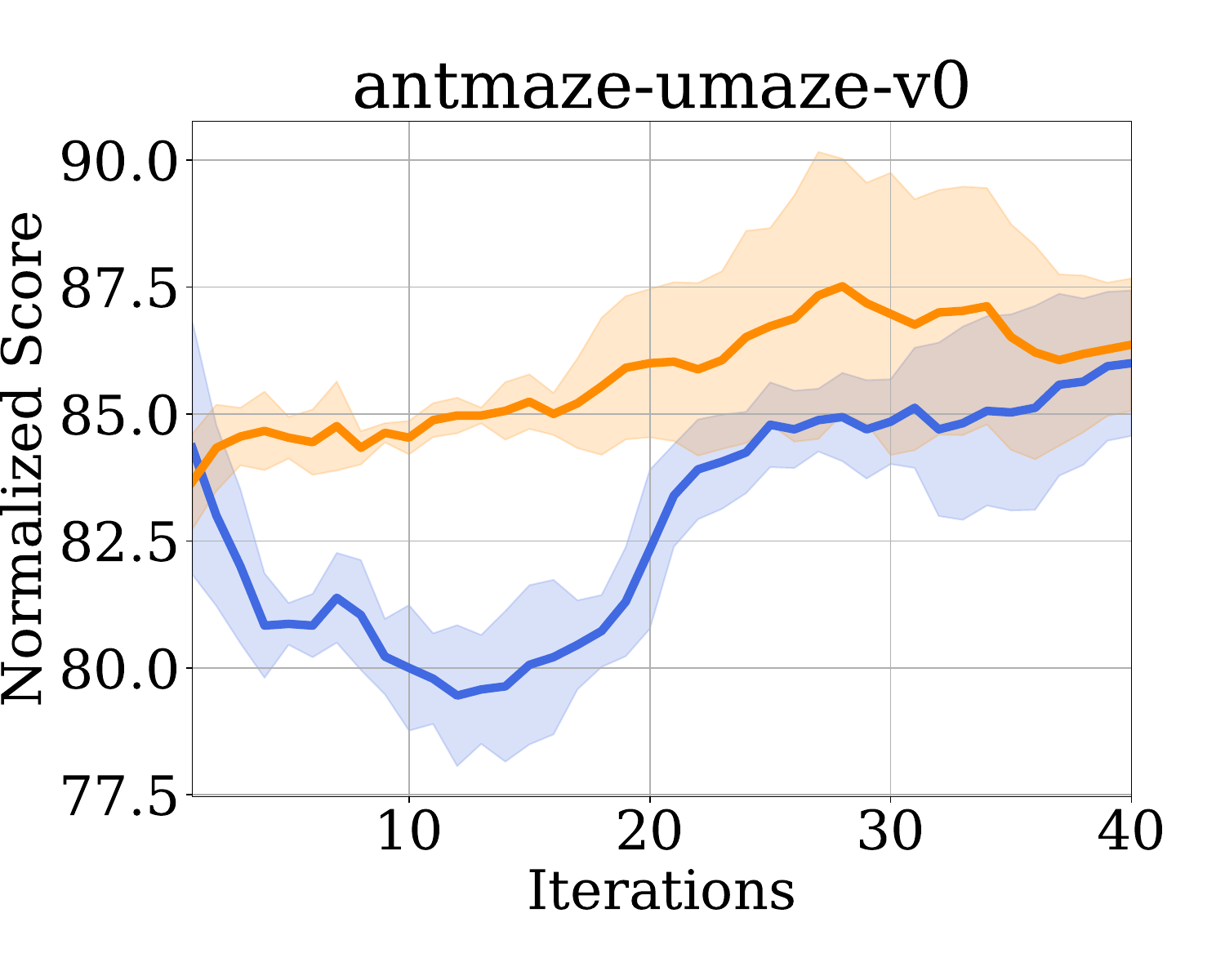}
    \end{minipage}
    \begin{minipage}{.31\textwidth}
        \centering
        \includegraphics[width=\textwidth, height=0.8\textwidth]{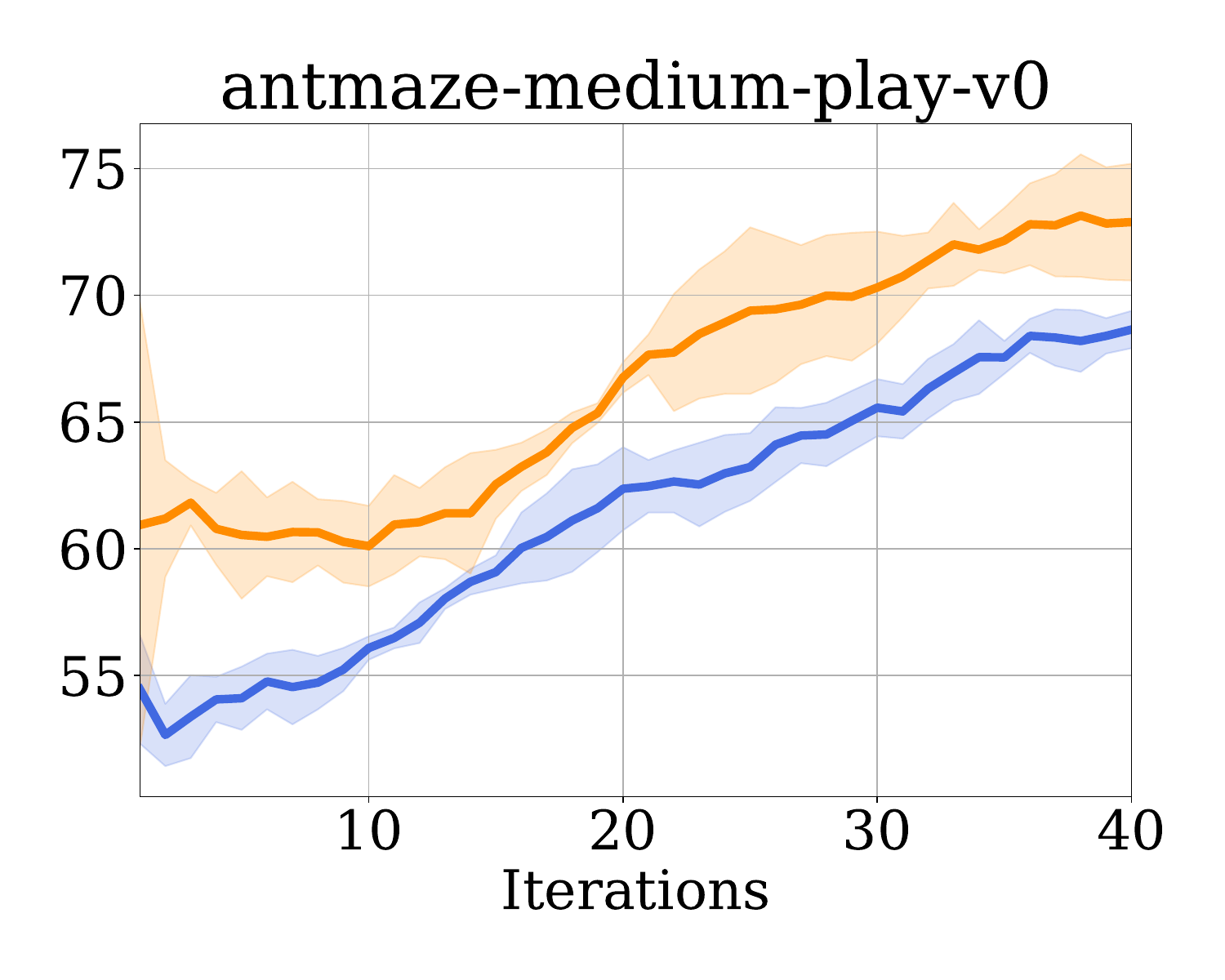}
    \end{minipage}
    \begin{minipage}{.31\textwidth}
        \centering
        \includegraphics[width=\textwidth, height=0.8\textwidth]{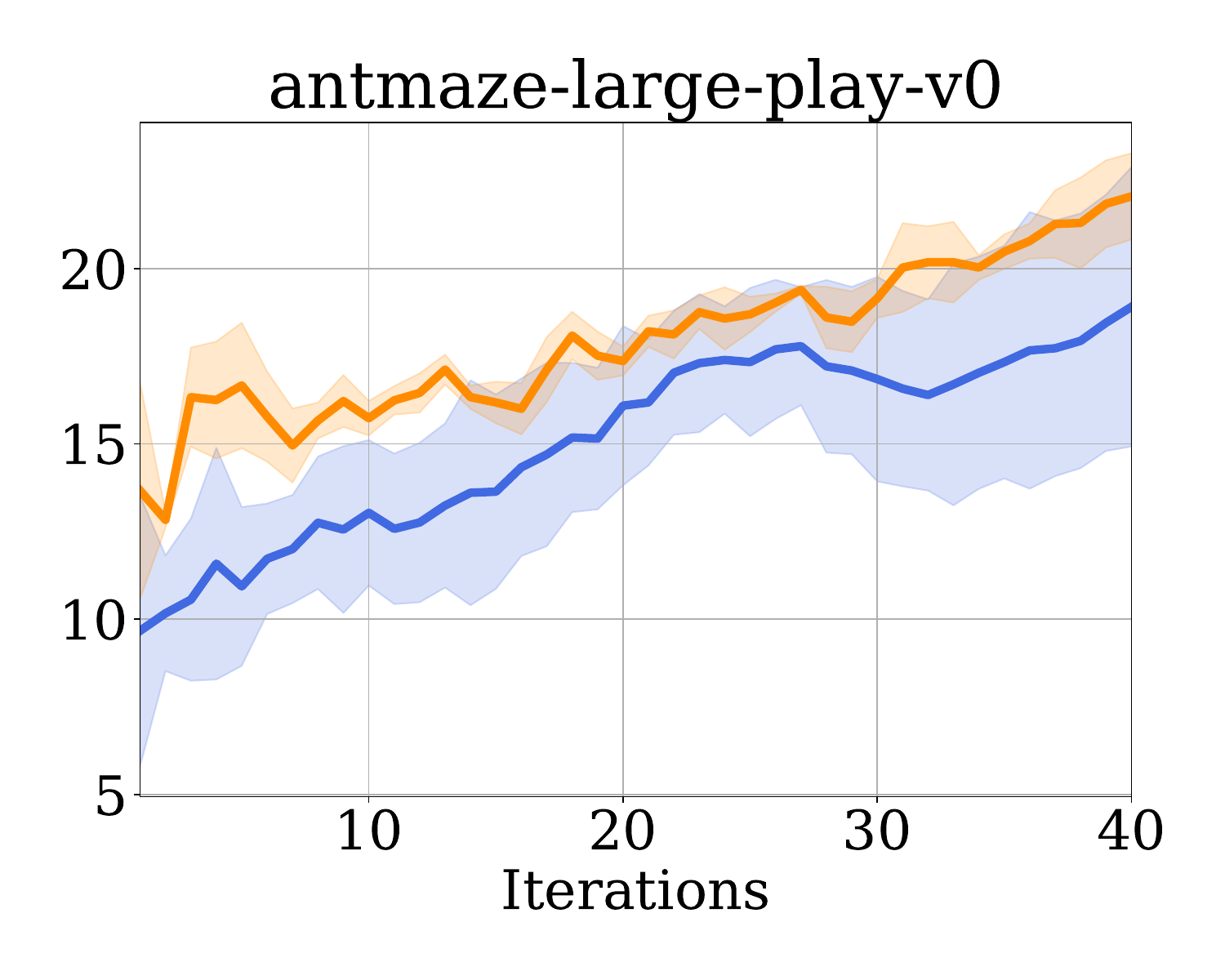}
    \end{minipage}
    
    
    \begin{minipage}{.31\textwidth}
        \centering
        \includegraphics[width=\textwidth, height=0.8\textwidth]{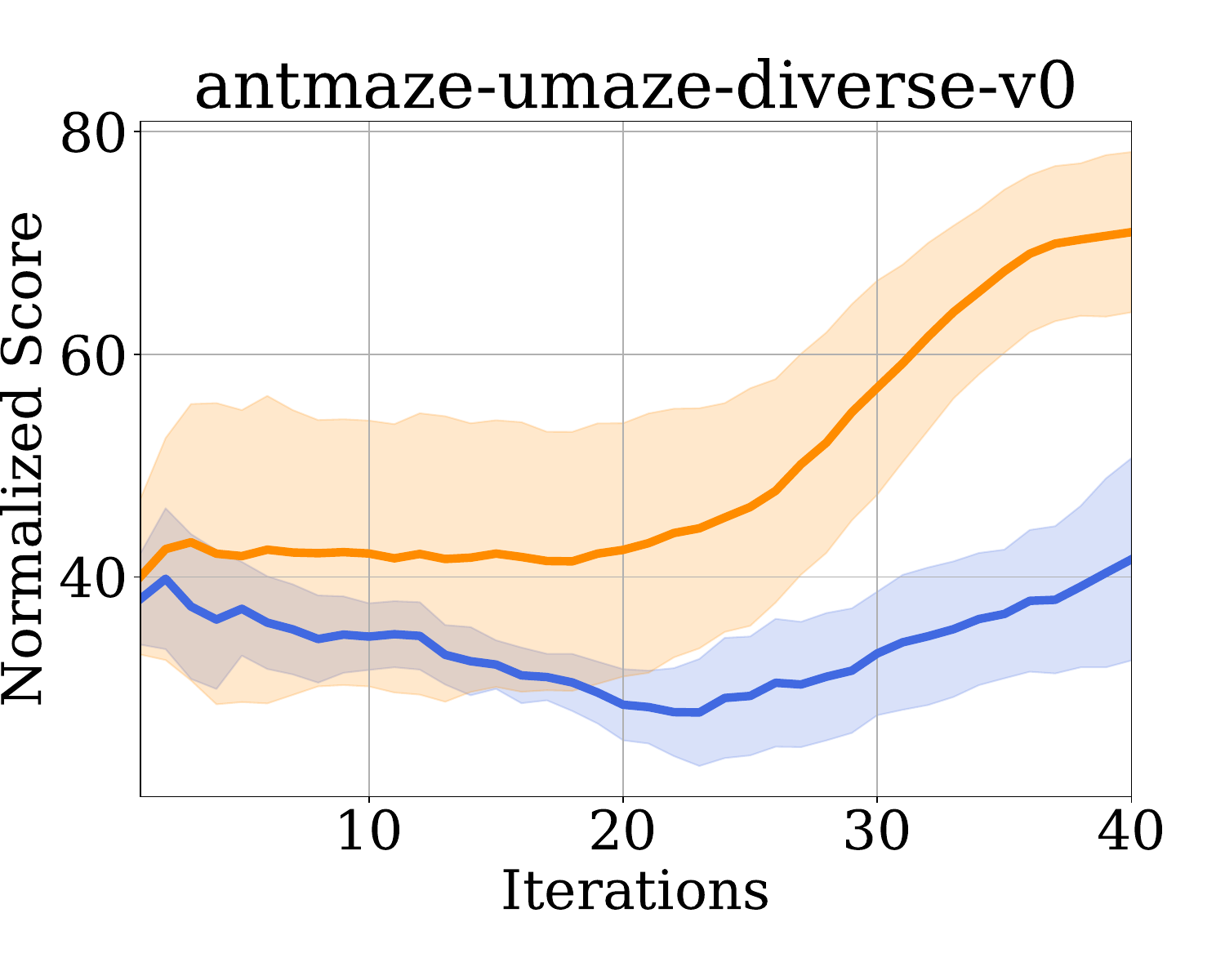}
    \end{minipage}
    \begin{minipage}{.31\textwidth}
        \centering
        \includegraphics[width=\textwidth, height=0.8\textwidth]{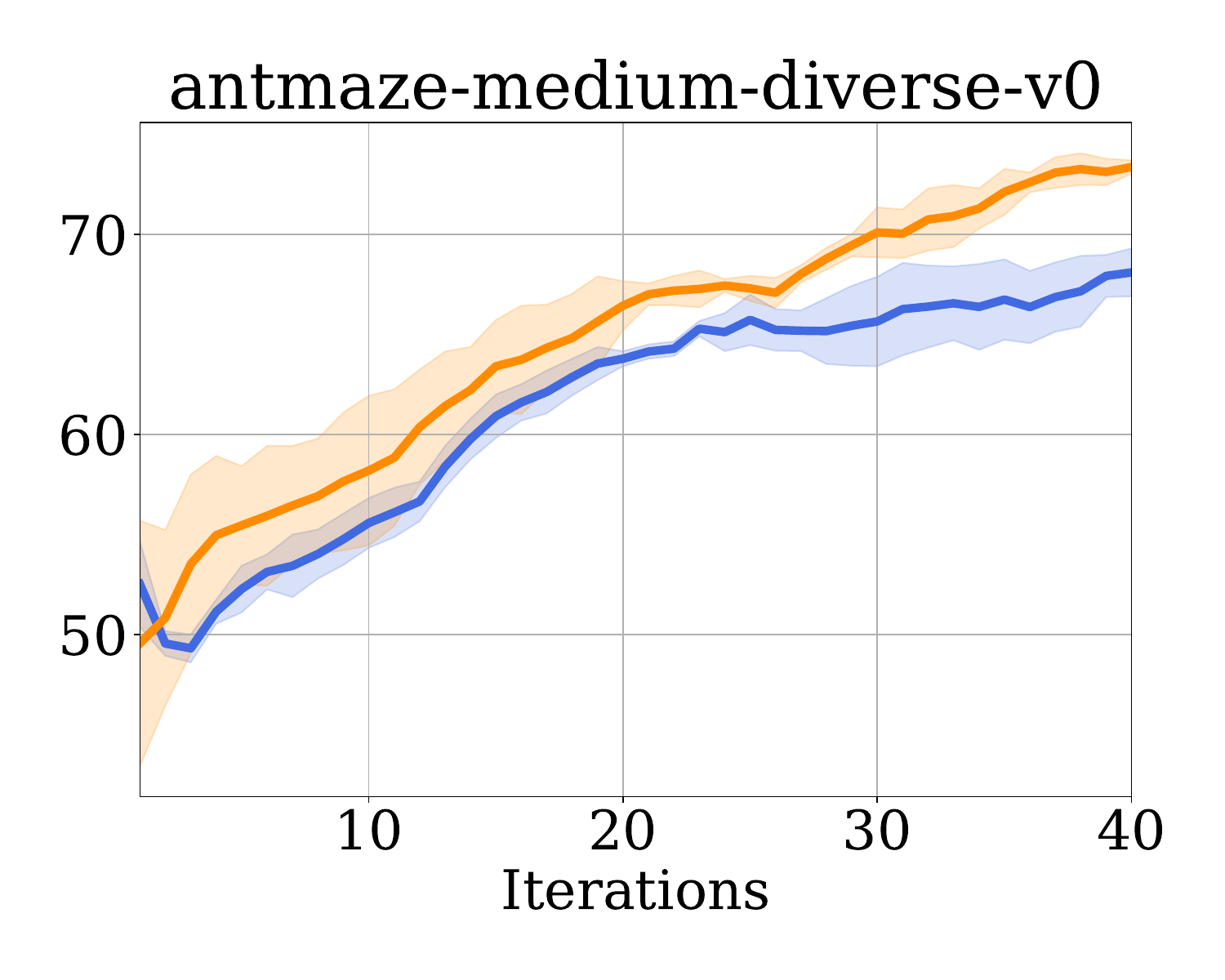}
    \end{minipage}
    \begin{minipage}{.31\textwidth}
        \centering
        \includegraphics[width=\textwidth, height=0.8\textwidth]{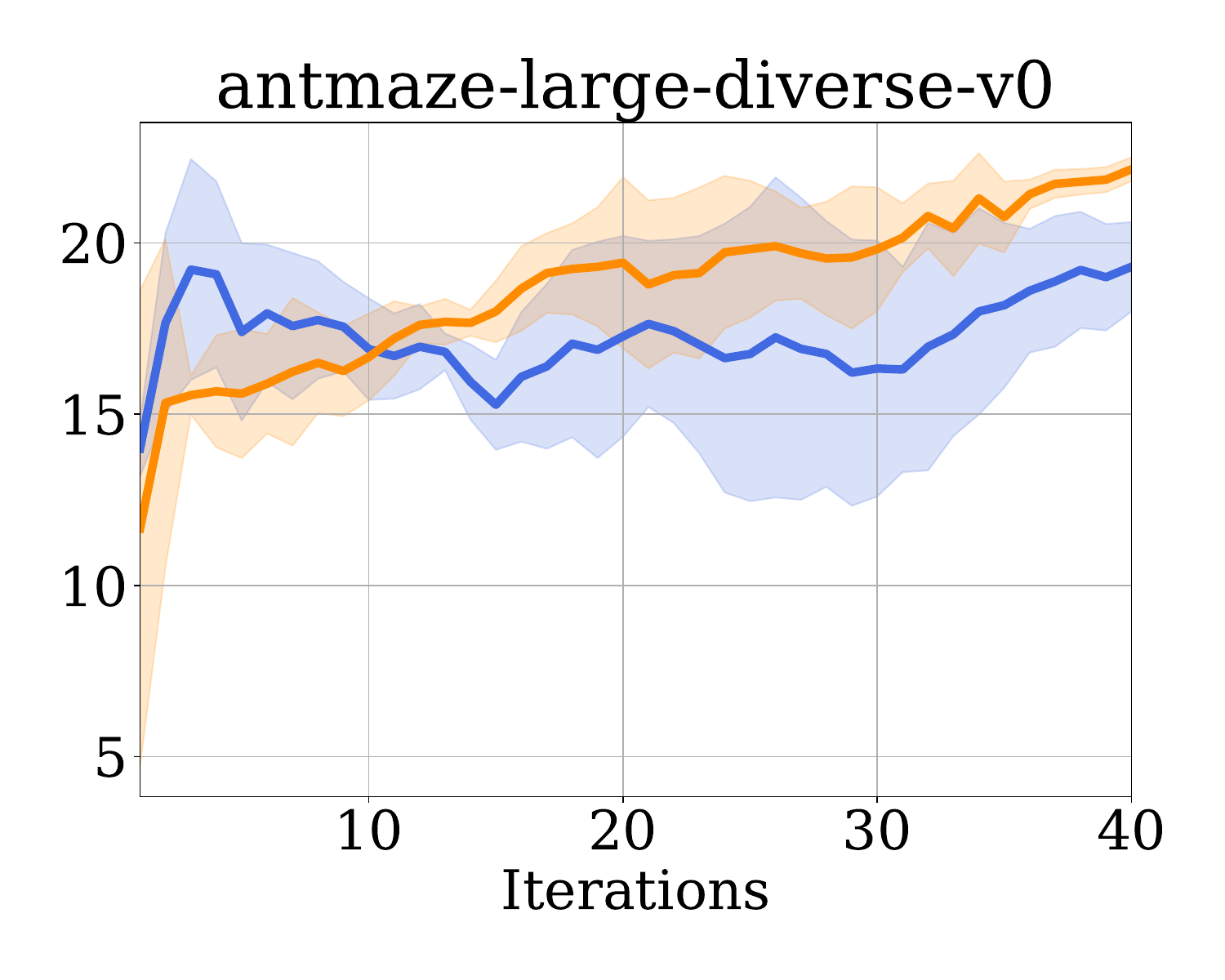}
    \end{minipage}
    
    
    \begin{minipage}{.31\textwidth}
        \centering
        \includegraphics[width=\textwidth, height=0.8\textwidth]{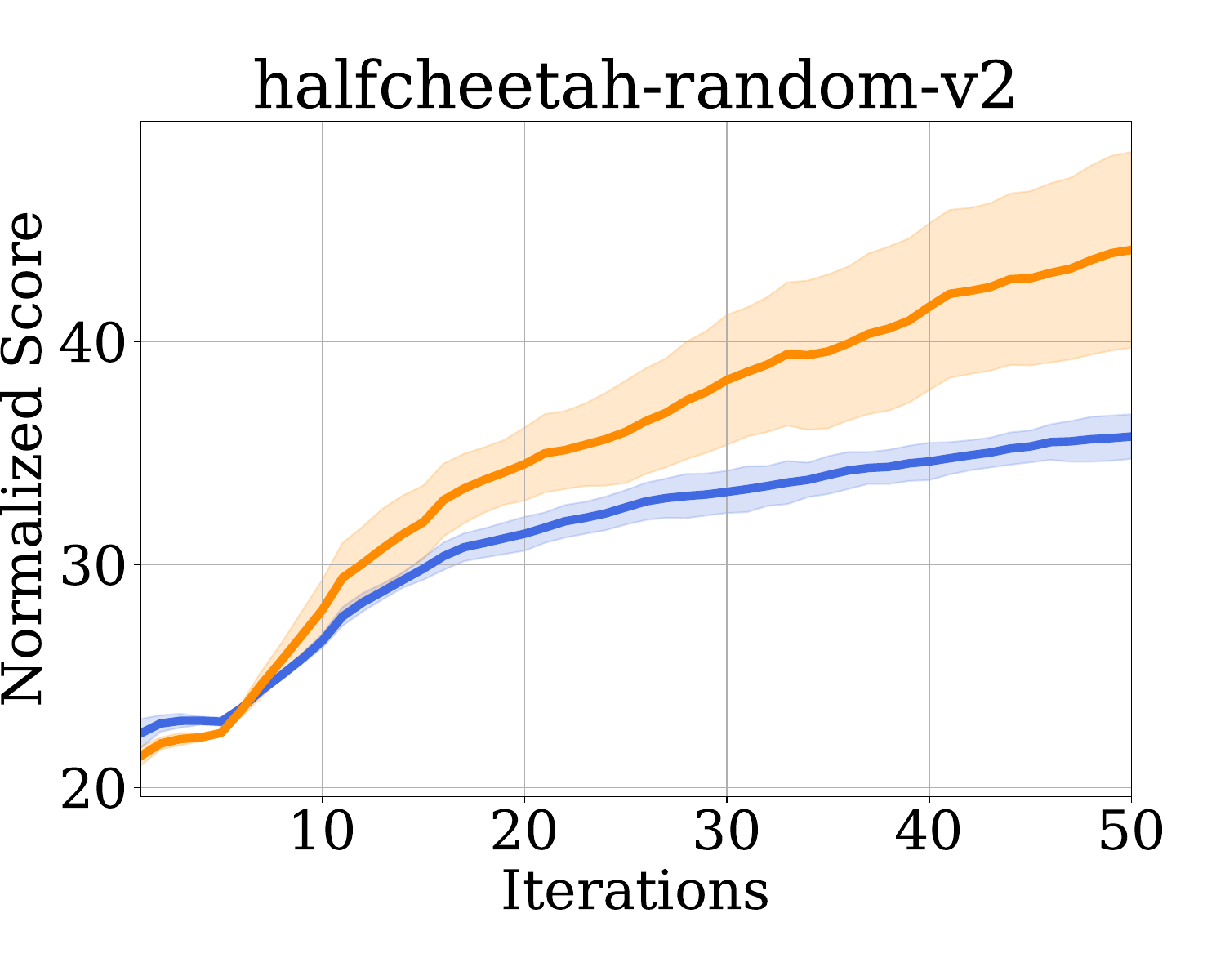}
    \end{minipage}
    \begin{minipage}{.31\textwidth}
        \centering
        \includegraphics[width=\textwidth, height=0.8\textwidth]{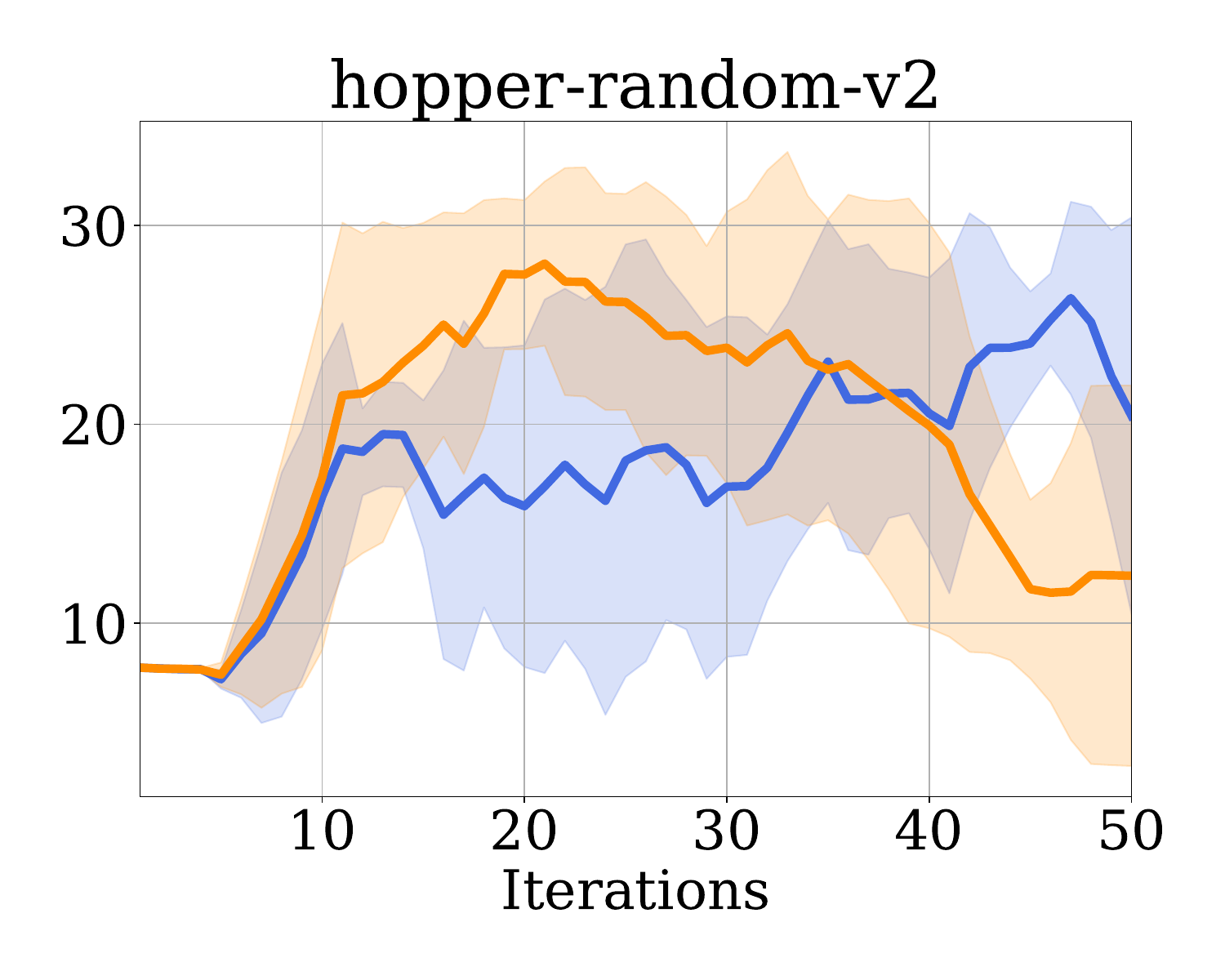}
    \end{minipage}
    \begin{minipage}{.31\textwidth}
        \centering
        \includegraphics[width=\textwidth, height=0.8\textwidth]{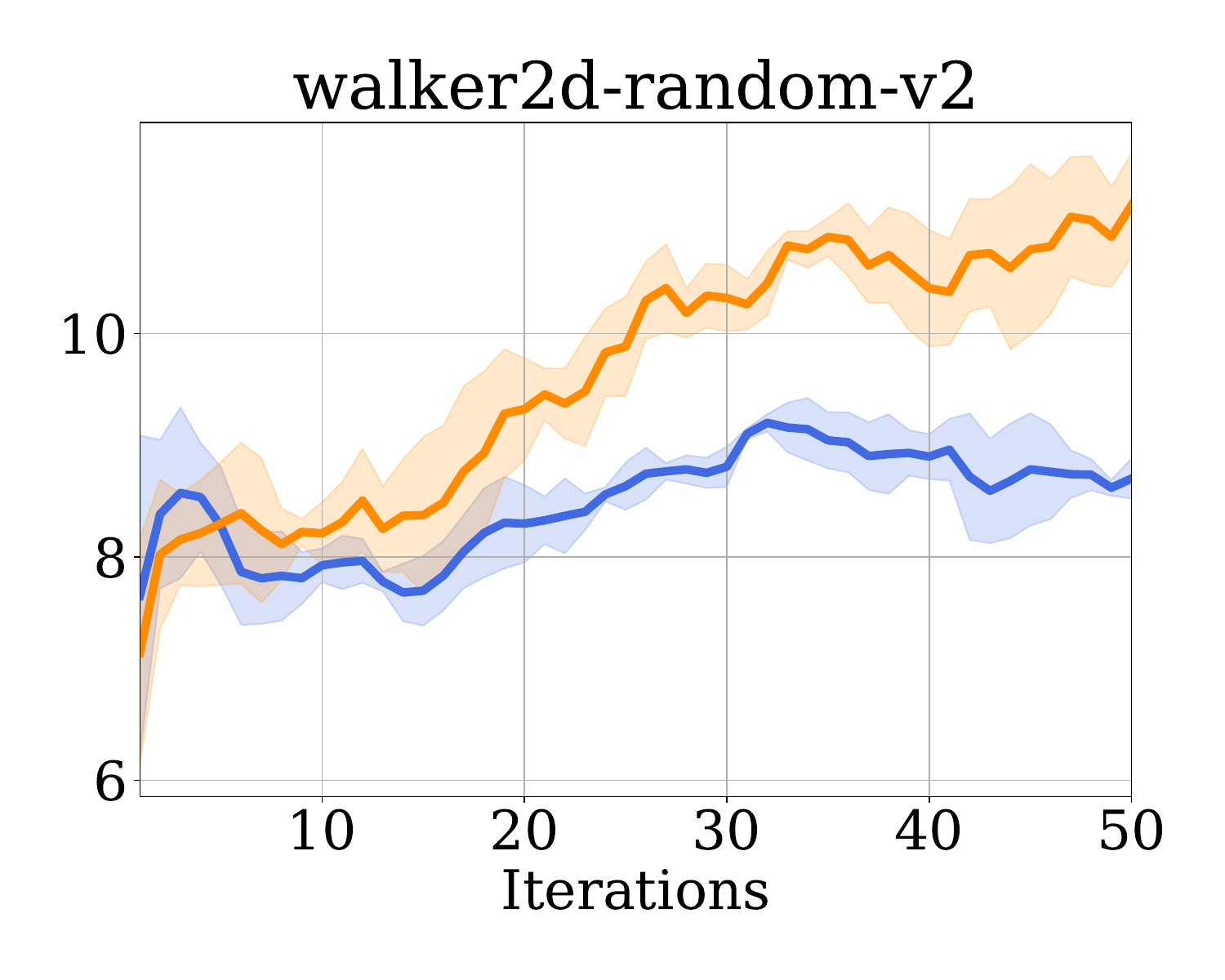}
    \end{minipage}
    
    
    \begin{minipage}{.31\textwidth}
        \centering
        \includegraphics[width=\textwidth, height=0.8\textwidth]{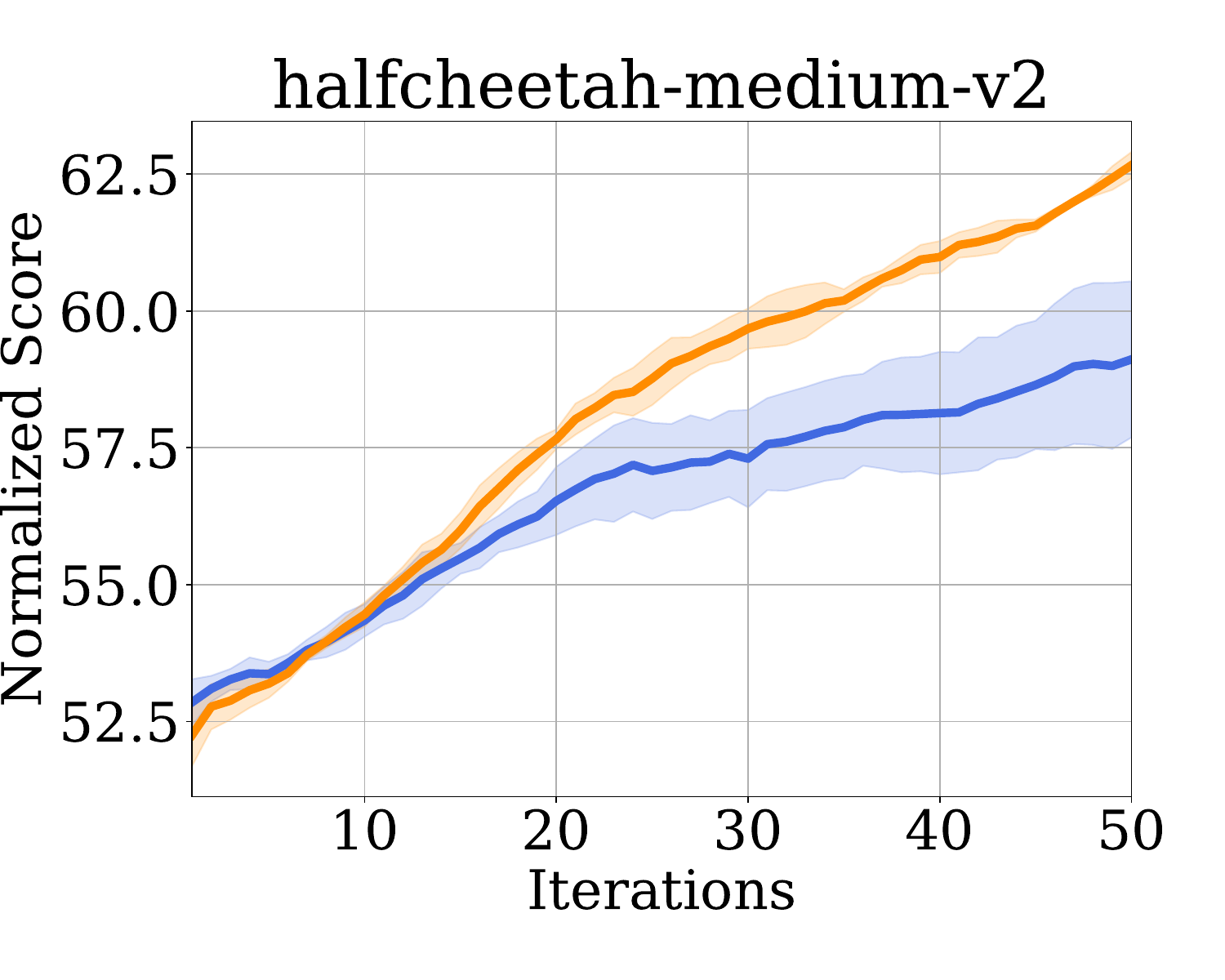}
    \end{minipage}
    \begin{minipage}{.31\textwidth}
        \centering
        \includegraphics[width=\textwidth, height=0.8\textwidth]{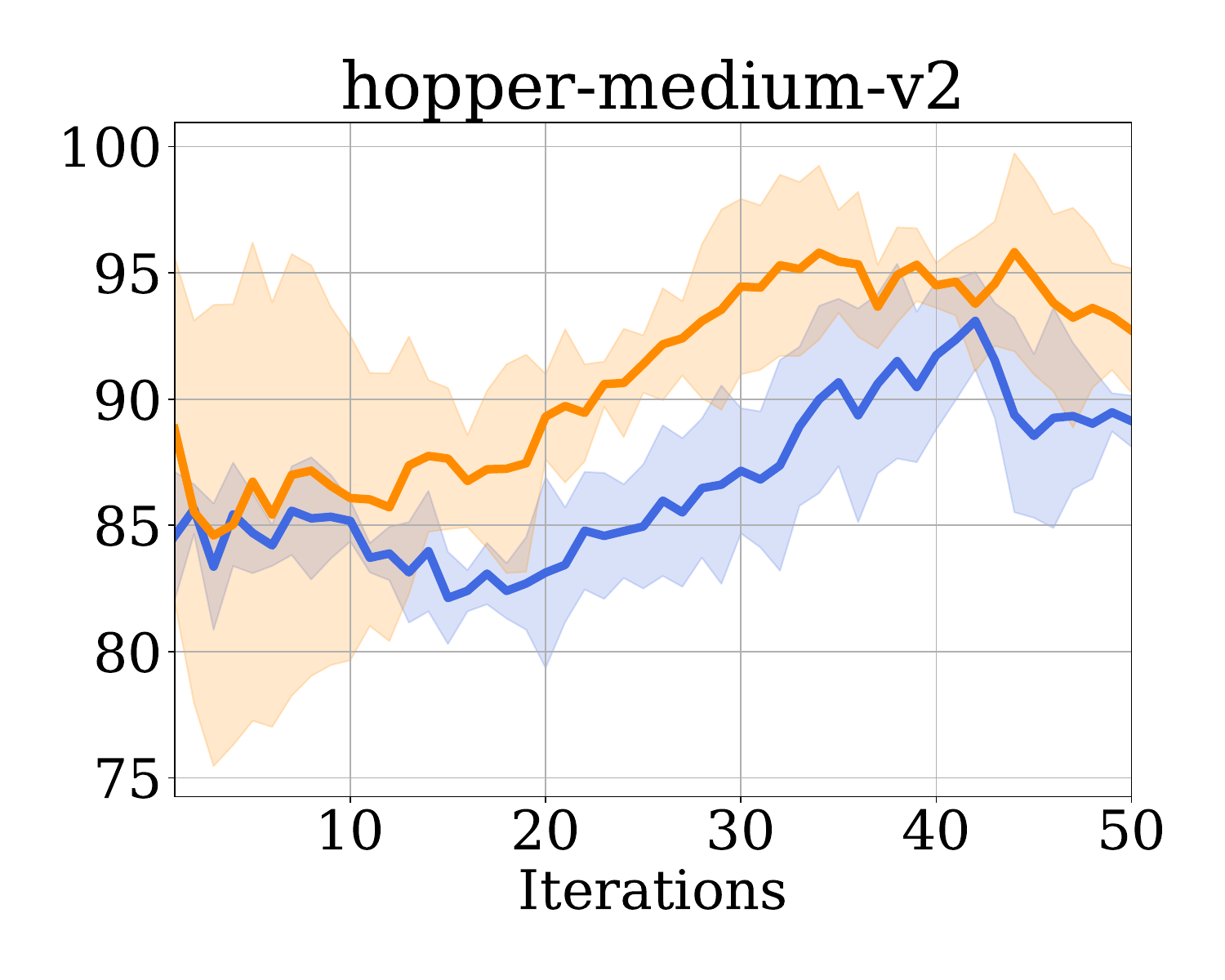}
    \end{minipage}
    \begin{minipage}{.31\textwidth}
        \centering
        \includegraphics[width=\textwidth, height=0.8\textwidth]{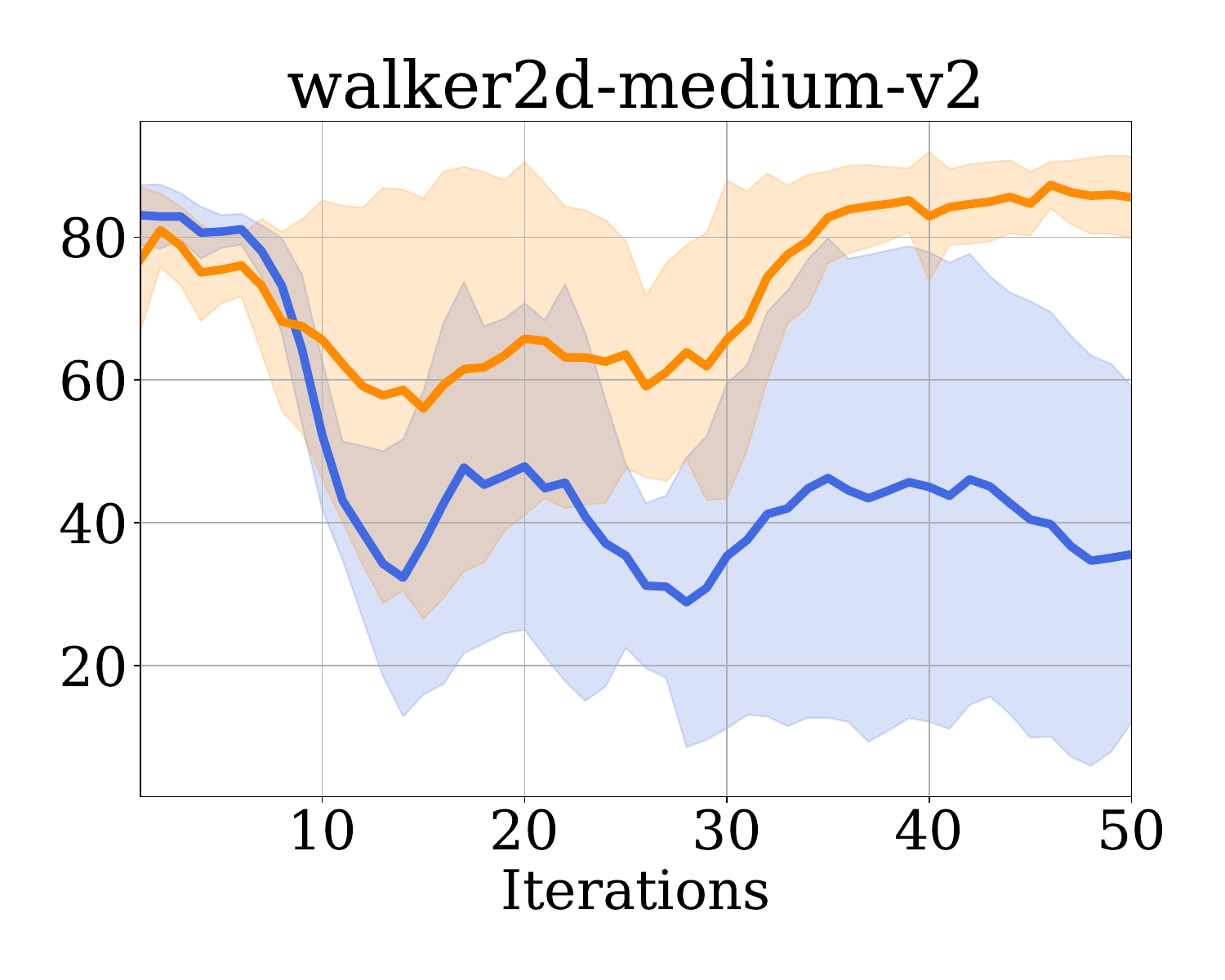}
    \end{minipage}

    \caption{[Best viewed in color] Full Results of our algorithm compared with the corresponding fine-tuning baseline. The plots show the results averaged over multiple random seeds, and the shaded region denotes the standard deviation. It can be observed that our method performs better overall when compared to the corresponding baselines.}

    \label{Figure: MainResultsAppendix}
	
\end{figure*}

\begin{figure*}[t]
    \centering
    \begin{minipage}{\textwidth}
        \centering
        \includegraphics[height=0.35cm]{resources/Legend.png}
    \end{minipage}

    \begin{minipage}{.33\textwidth}
        \centering
        \includegraphics[width=\textwidth, height=0.8\textwidth]{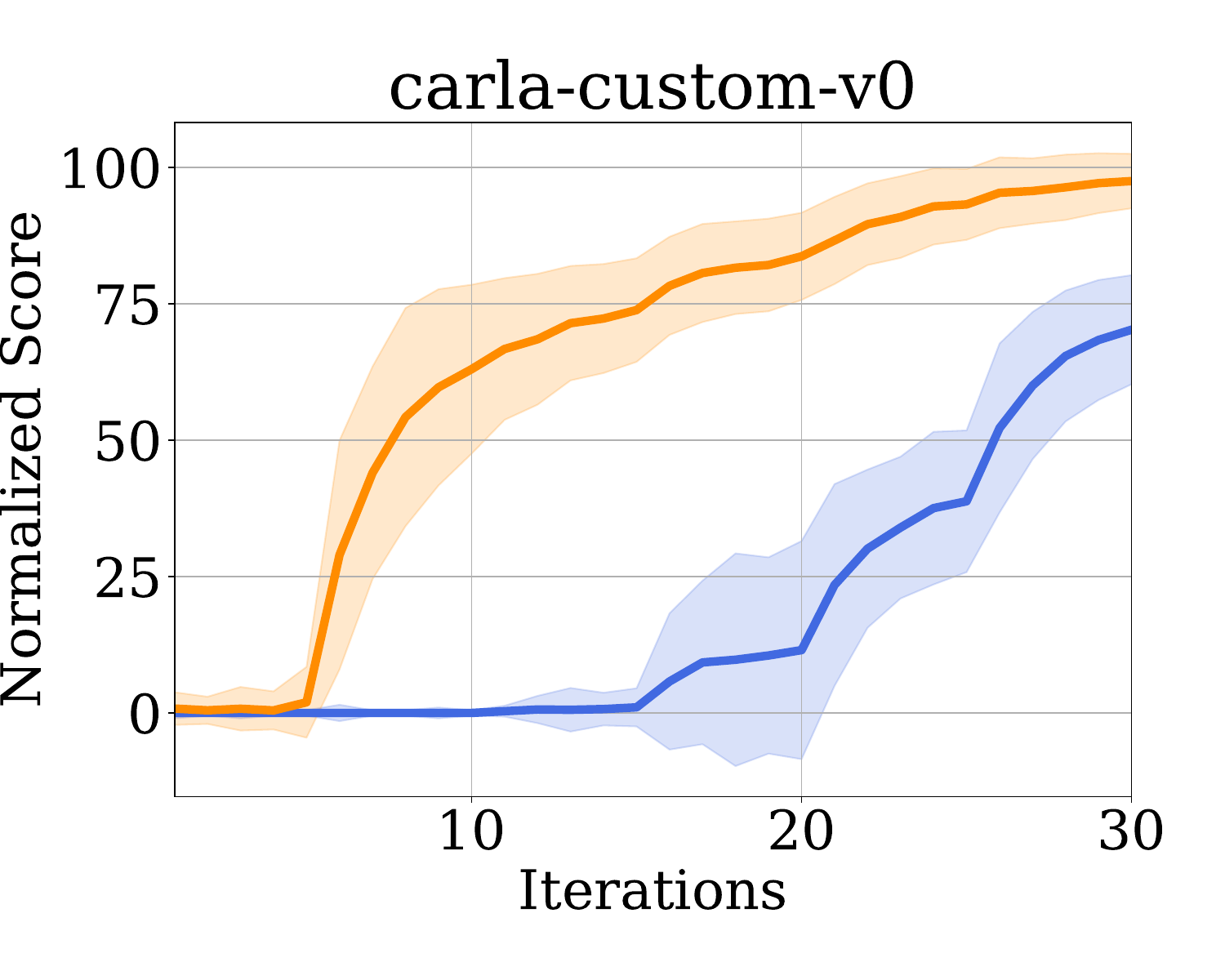}
    \end{minipage}
    \begin{minipage}{.33\textwidth}
        \centering
        \includegraphics[width=\textwidth, height=0.8\textwidth]{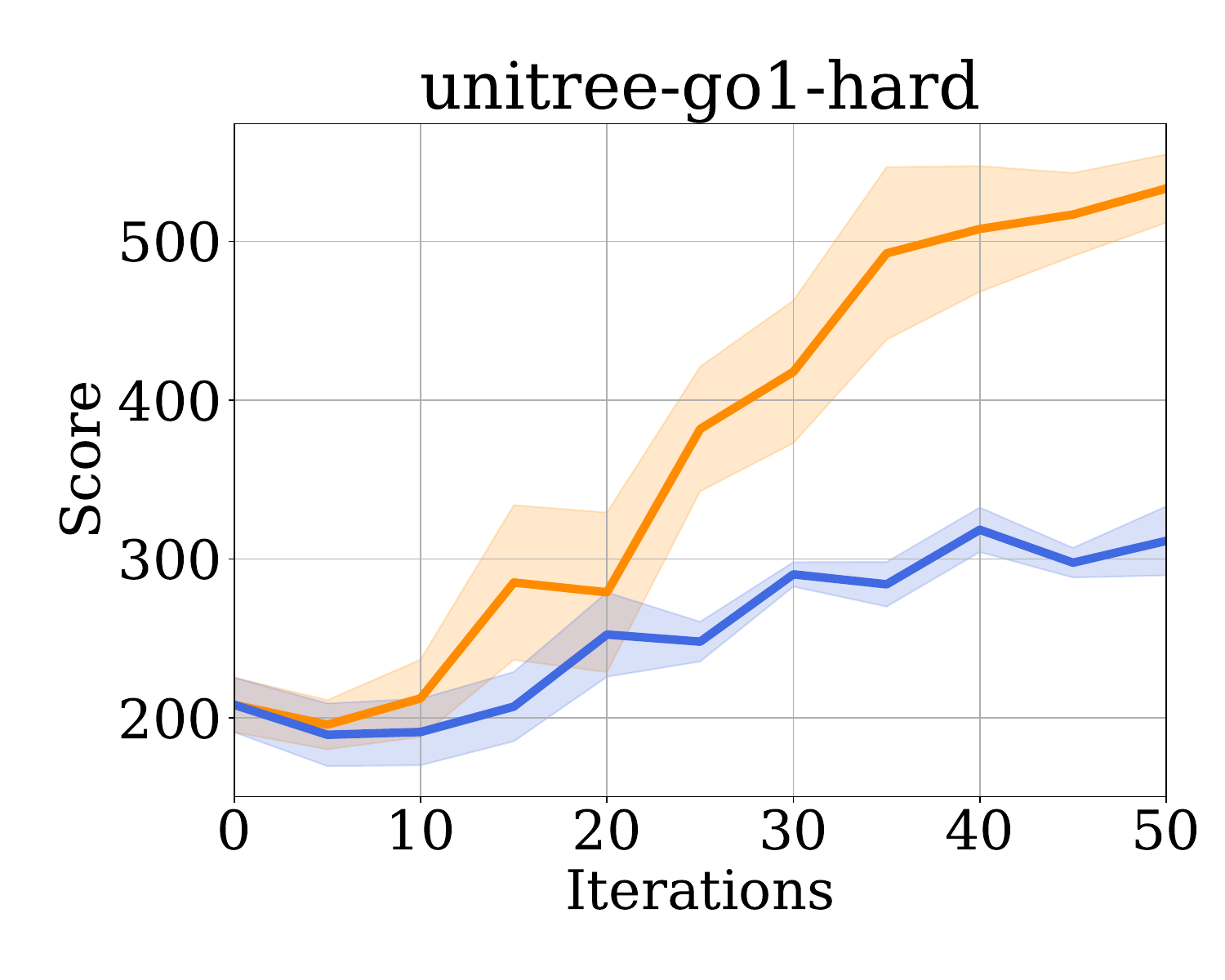}
    \end{minipage}
    \begin{minipage}{.33\textwidth}
        \centering
        \includegraphics[width=\textwidth, height=0.8\textwidth]{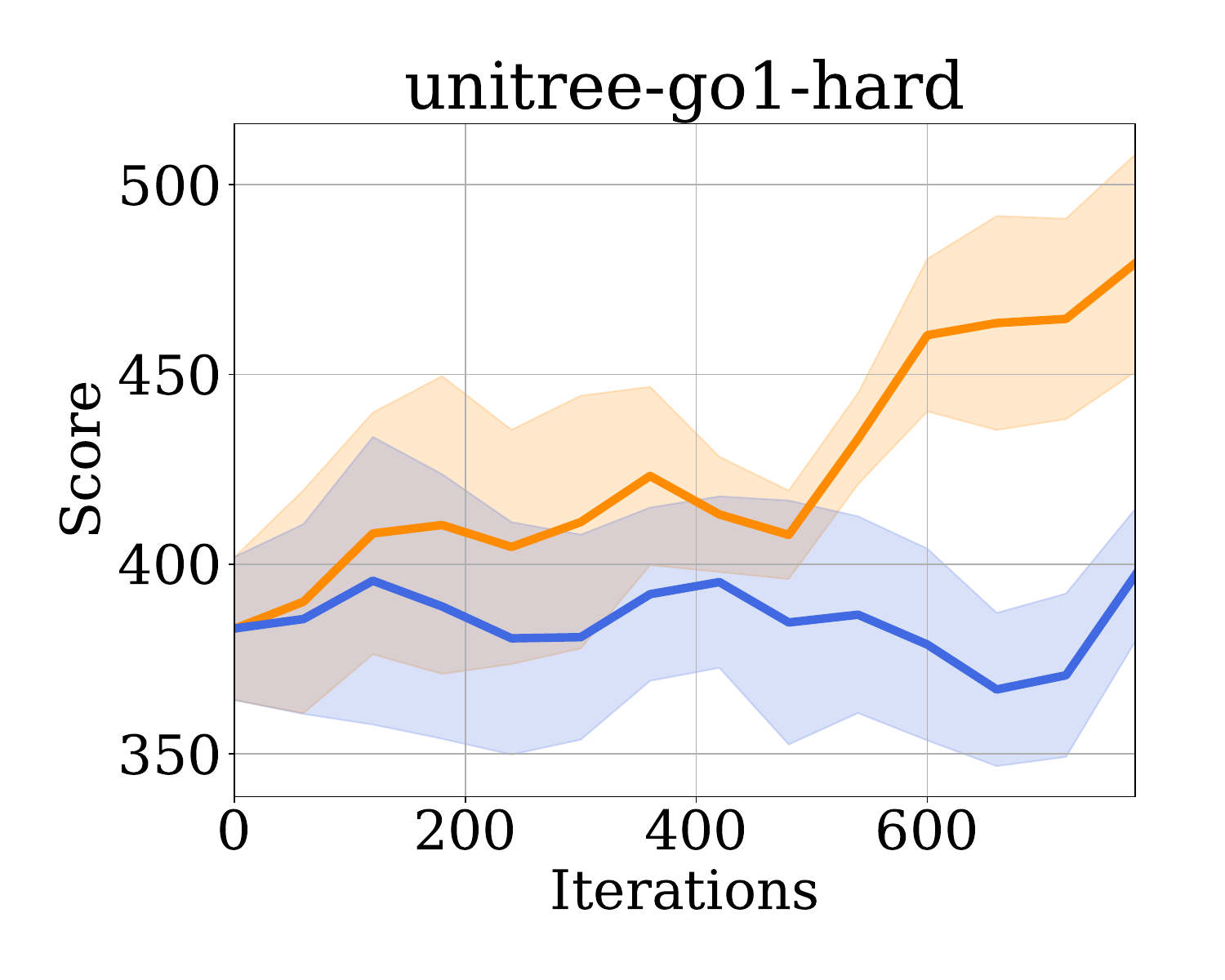}
    \end{minipage}
    
    \begin{minipage}{.33\textwidth}
        \centering
        \includegraphics[width=\textwidth, height=0.8\textwidth]{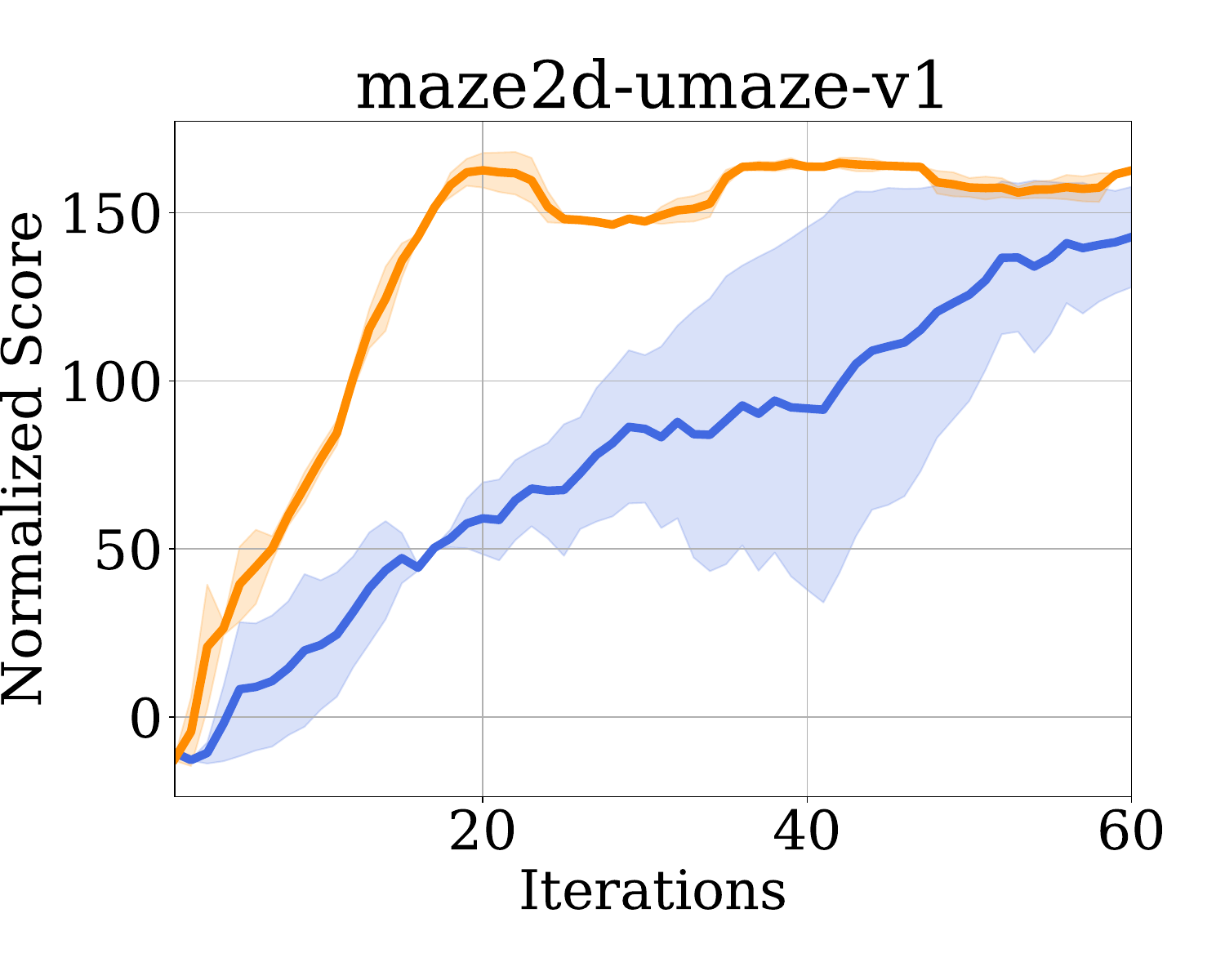}
    \end{minipage}
    \begin{minipage}{.33\textwidth}
        \centering
        \includegraphics[width=\textwidth, height=0.8\textwidth]{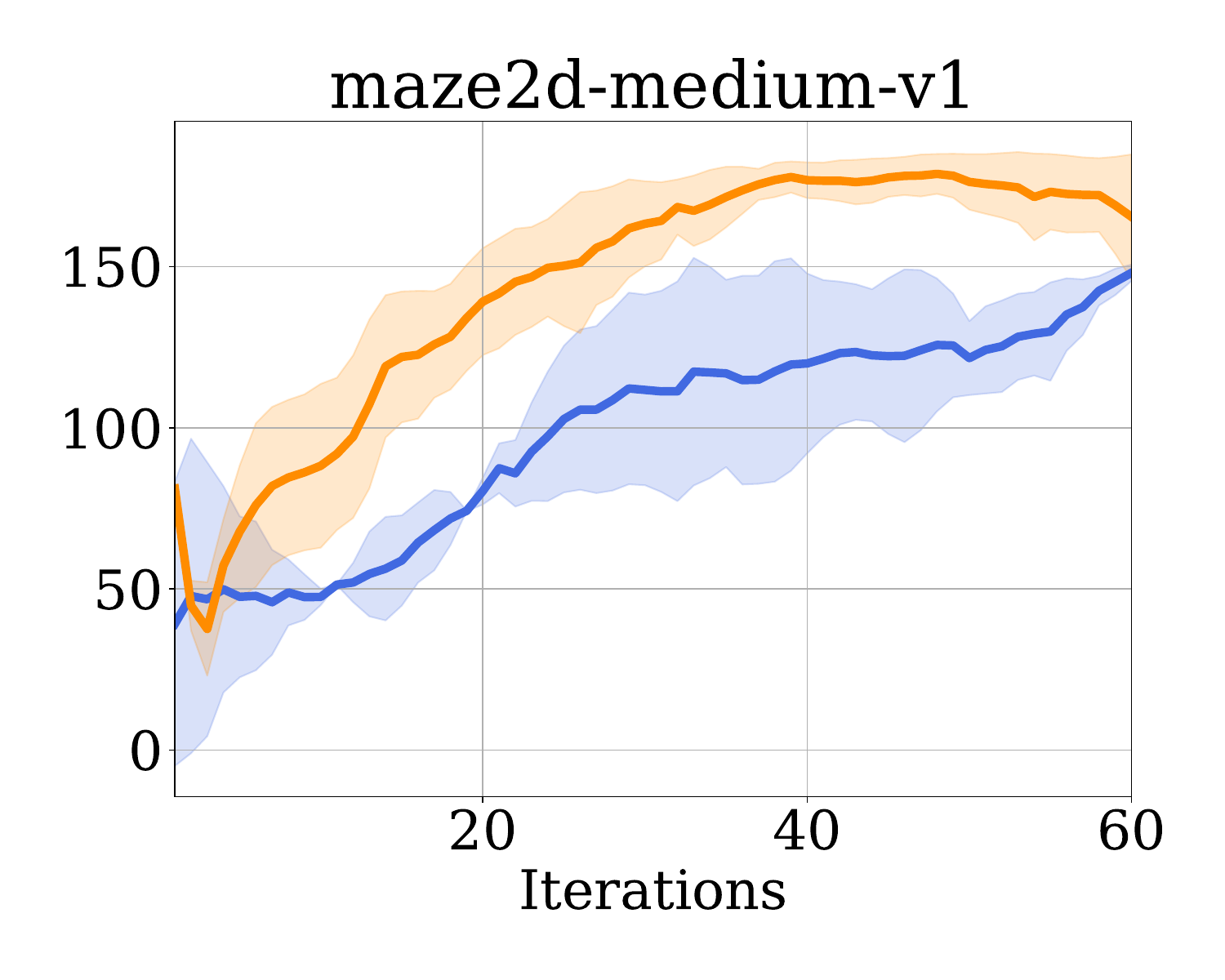}
    \end{minipage}
    \begin{minipage}{.33\textwidth}
        \centering
        \includegraphics[width=\textwidth, height=0.8\textwidth]{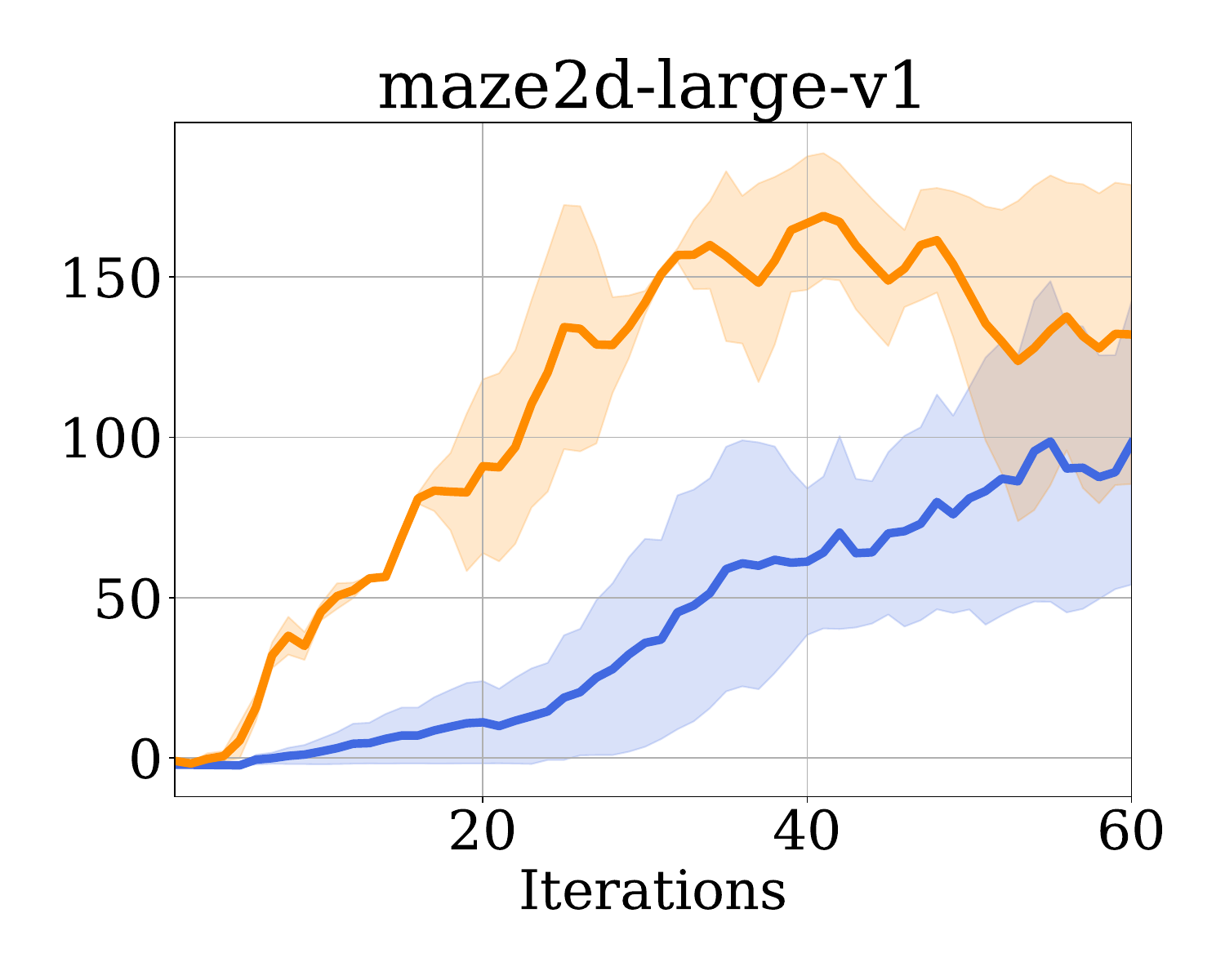}
    \end{minipage}
    
    \caption{[Best viewed in color] The top row left and center plots are additional experiments similar to Figure~\ref{Figure: MainResultsAppendix} and correspond to the CARLA and IsaacSim-Go1 experiments. The top row right plot corresponds to the online experiments using the Go1 Robot. The bottom row corresponds to additional online ablation experiments with zero transitions in the offline dataset in the maze2d environment.}
    \label{Figure: OnlineAblation}
\end{figure*}

\begin{figure*}[t]
    \centering
    \begin{minipage}{0.33\textwidth}
        \includegraphics[width=\textwidth]{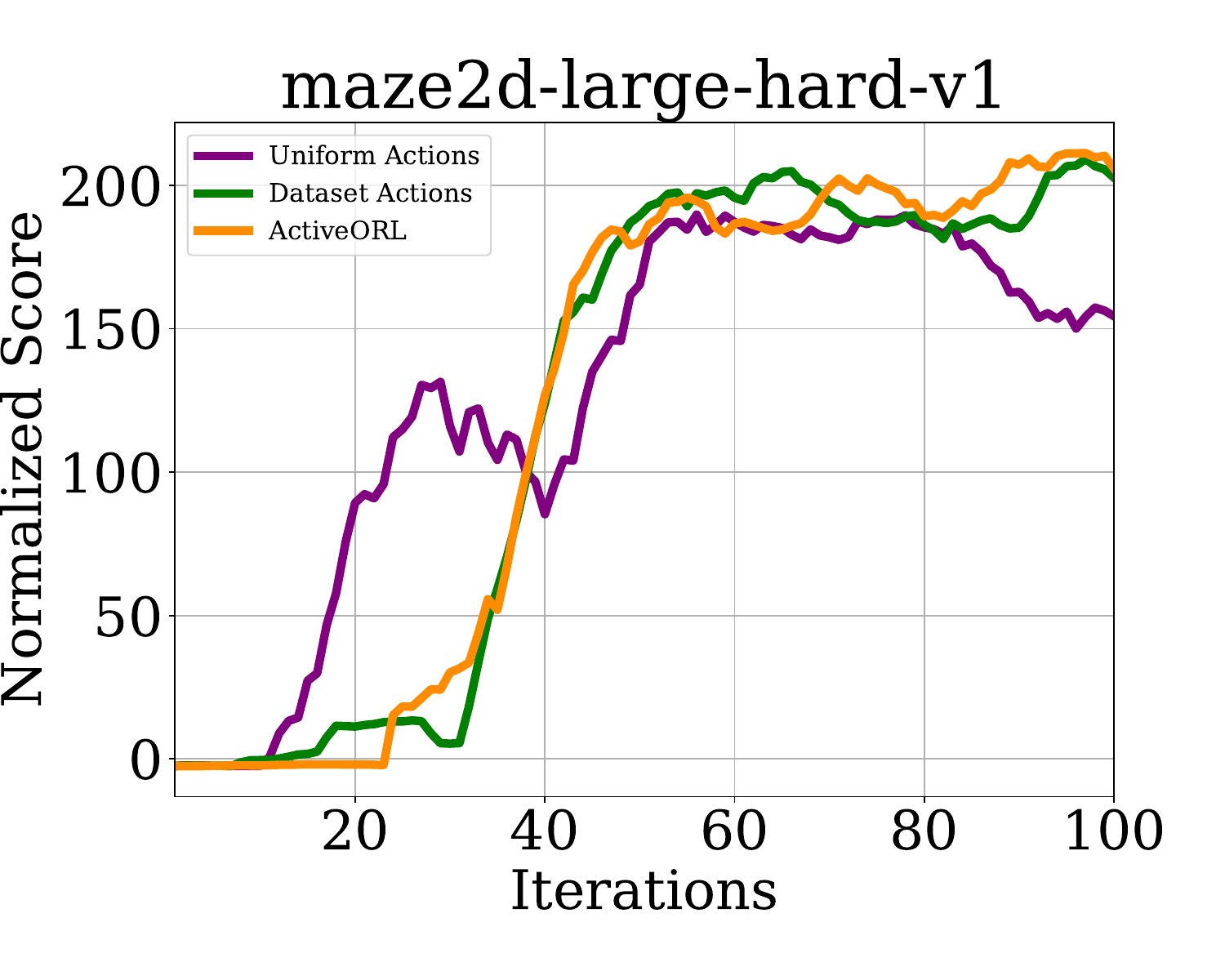}
    \end{minipage}
    \begin{minipage}{0.33\textwidth}
        \includegraphics[width=\textwidth]{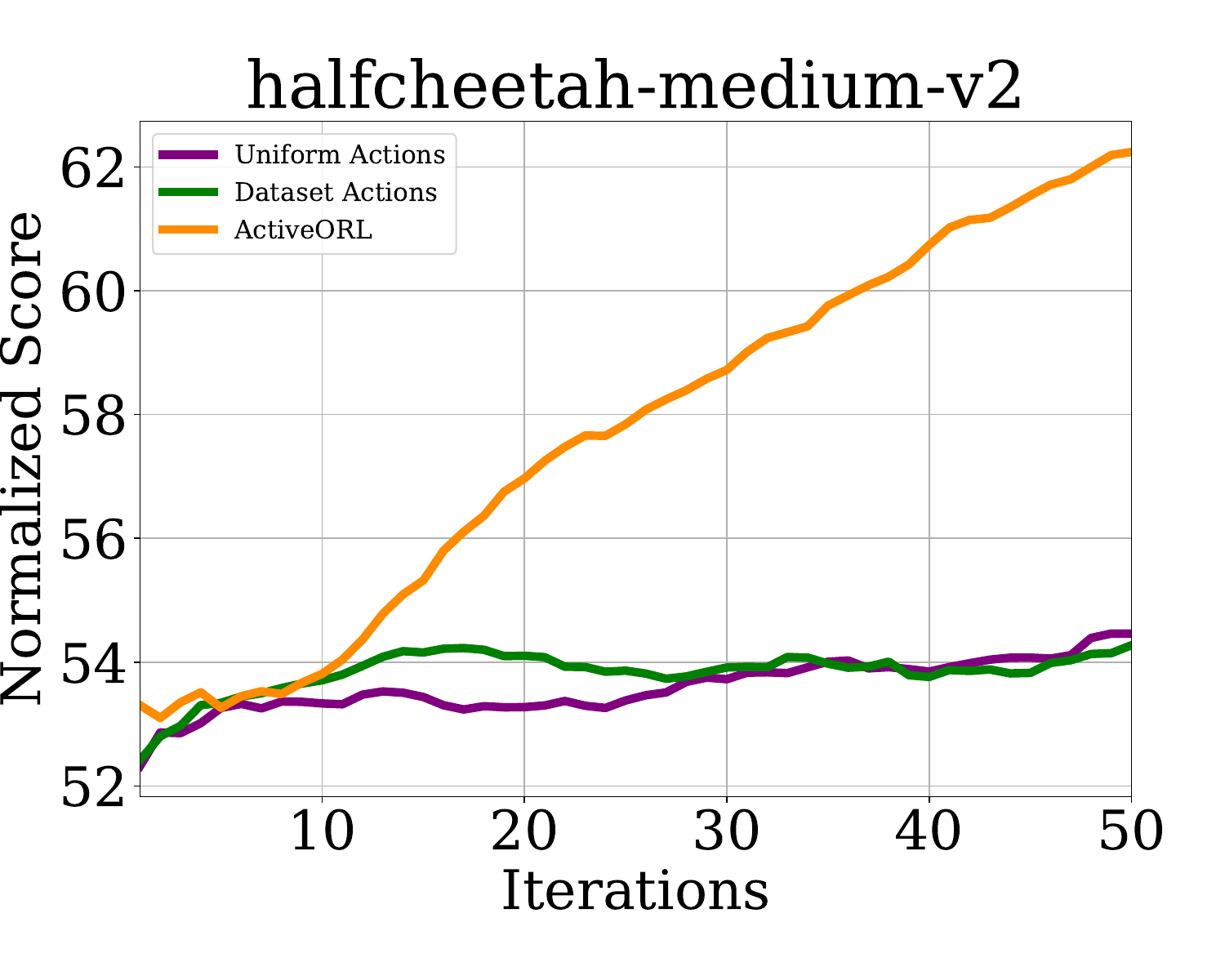}
    \end{minipage}
    \caption{Results of the ablation experiments comparing various choices of action selection during exploration, viz. (i) sampling from the uniform distribution, (ii) using the actions from the offline dataset, and (iii) our $\epsilon-$greedy uncertainty-based exploration policy. As can be seen on the maze environment (left), the uniform policy performs reasonably well for exploration when the state-action space is low dimensional. However, in the high dimensional space of the locomotion task (right), we can see that uniform policy does not work at all. Action selection from the dataset actions is slightly better, however even that does not work in environments such as AntMaze (results not included since no improvement can be observed using action selection strategies other than our $\epsilon-$greedy uncertainty based exploration policy).}
    \label{Figure: ActionAblation}
\end{figure*}

\begin{figure*}[t]
    \centering

    \begin{minipage}{0.20\textwidth}
        \centering
        \includegraphics[trim=4cm 3cm 2.7cm 3.2cm, clip=true, width=\textwidth, height=1.4\textwidth]{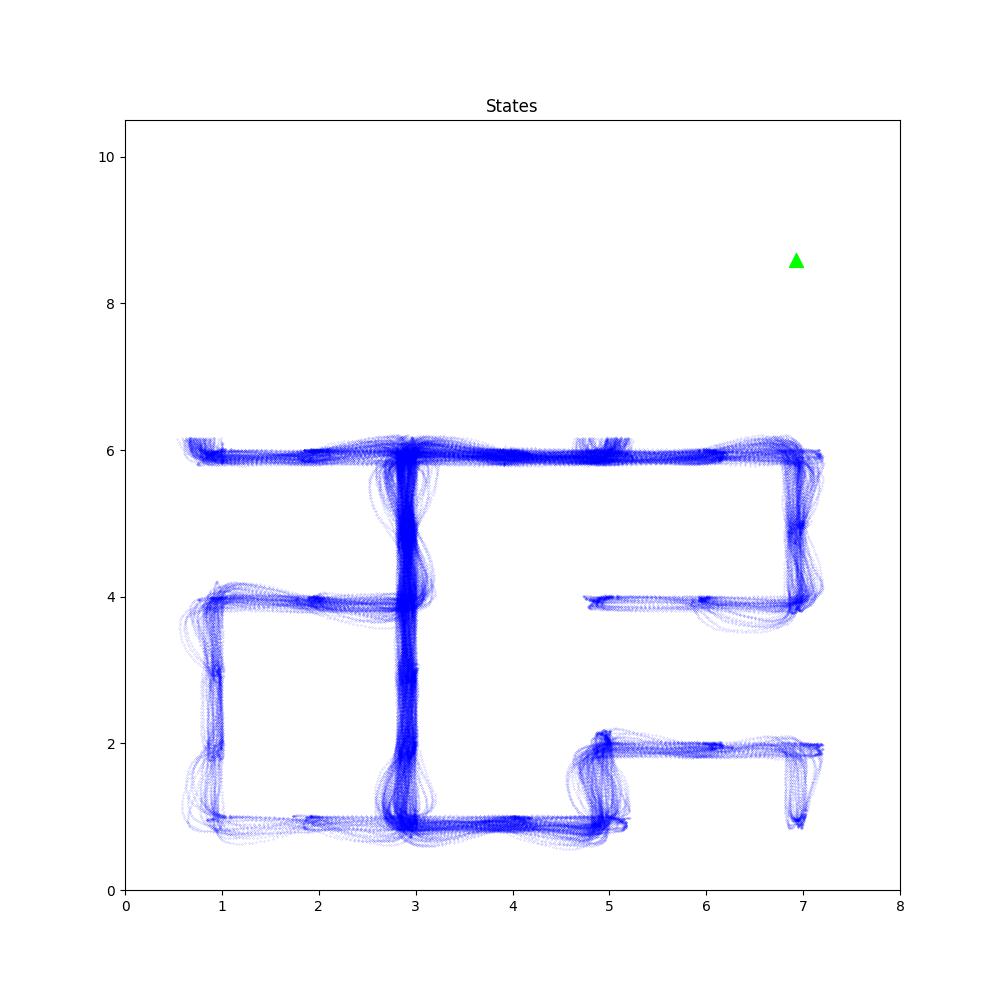}
    \end{minipage}
    \begin{minipage}{0.20\textwidth}
        \centering
        \includegraphics[trim=4cm 3cm 6.8cm 3.2cm, clip=true, width=\textwidth, height=1.4\textwidth]{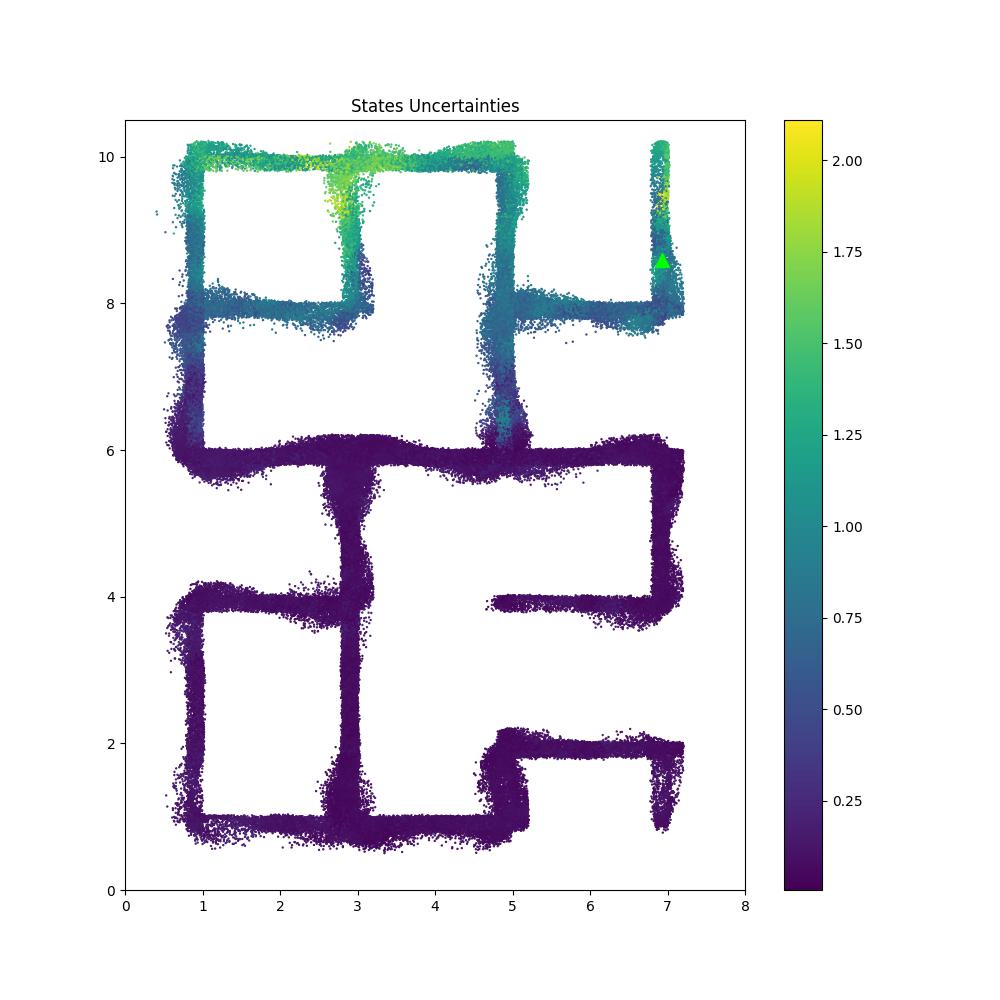}
    \end{minipage}
    \begin{minipage}{0.20\textwidth}
        \centering
        \includegraphics[trim=4cm 3cm 2.7cm 3.2cm, clip=true, width=\textwidth, height=1.4\textwidth]{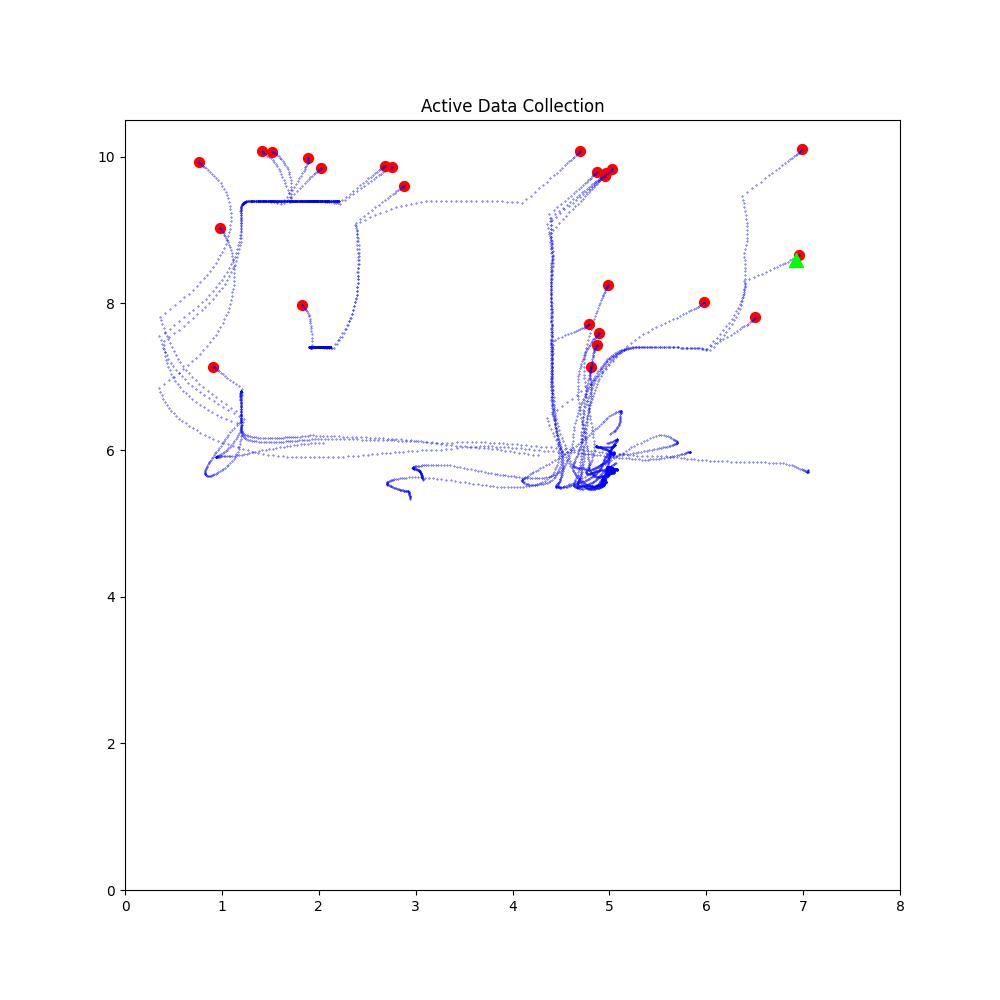}
    \end{minipage}
    \begin{minipage}{0.20\textwidth}
        \centering
        \includegraphics[trim=4cm 2.9cm 2.7cm 3.2cm, clip=true, width=\textwidth, height=1.4\textwidth]{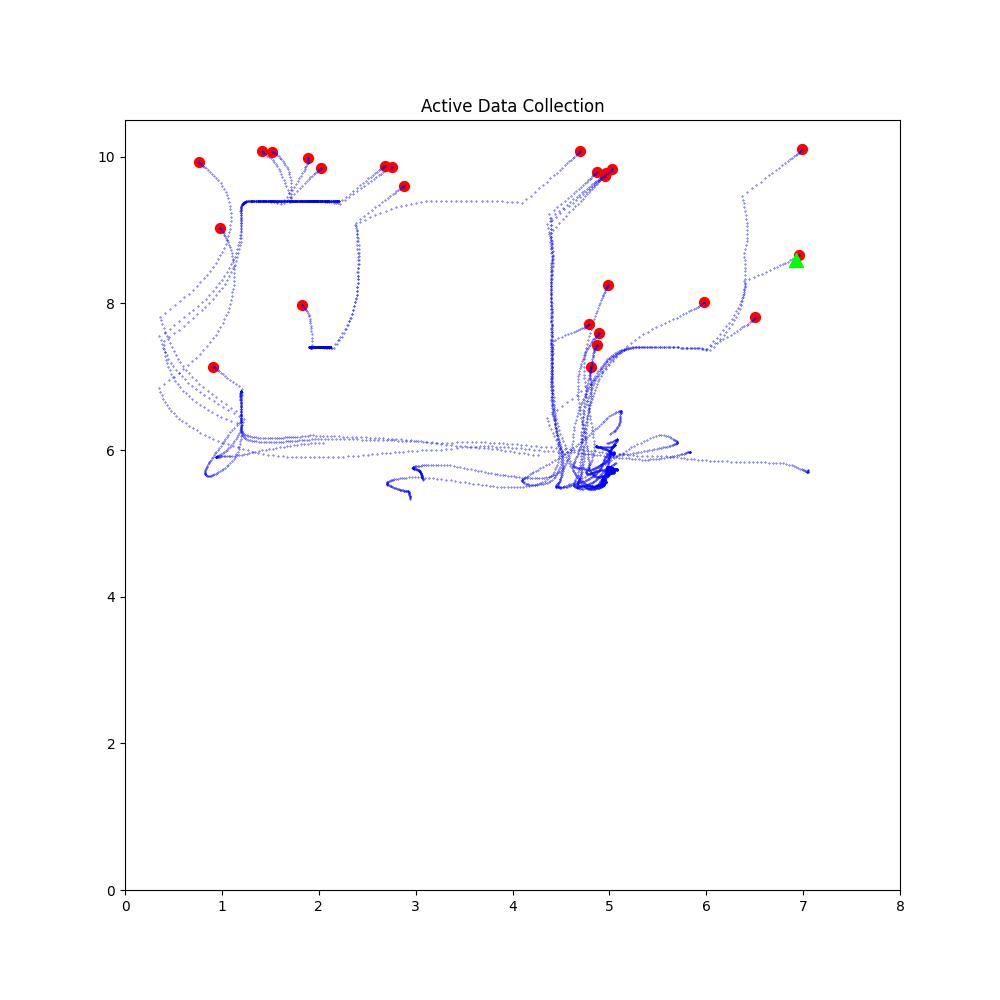}
    \end{minipage}


    \begin{minipage}{0.20\textwidth}
        \centering
        \includegraphics[trim=4cm 3cm 2.7cm 3.2cm, clip=true, width=\textwidth, height=1.4\textwidth]{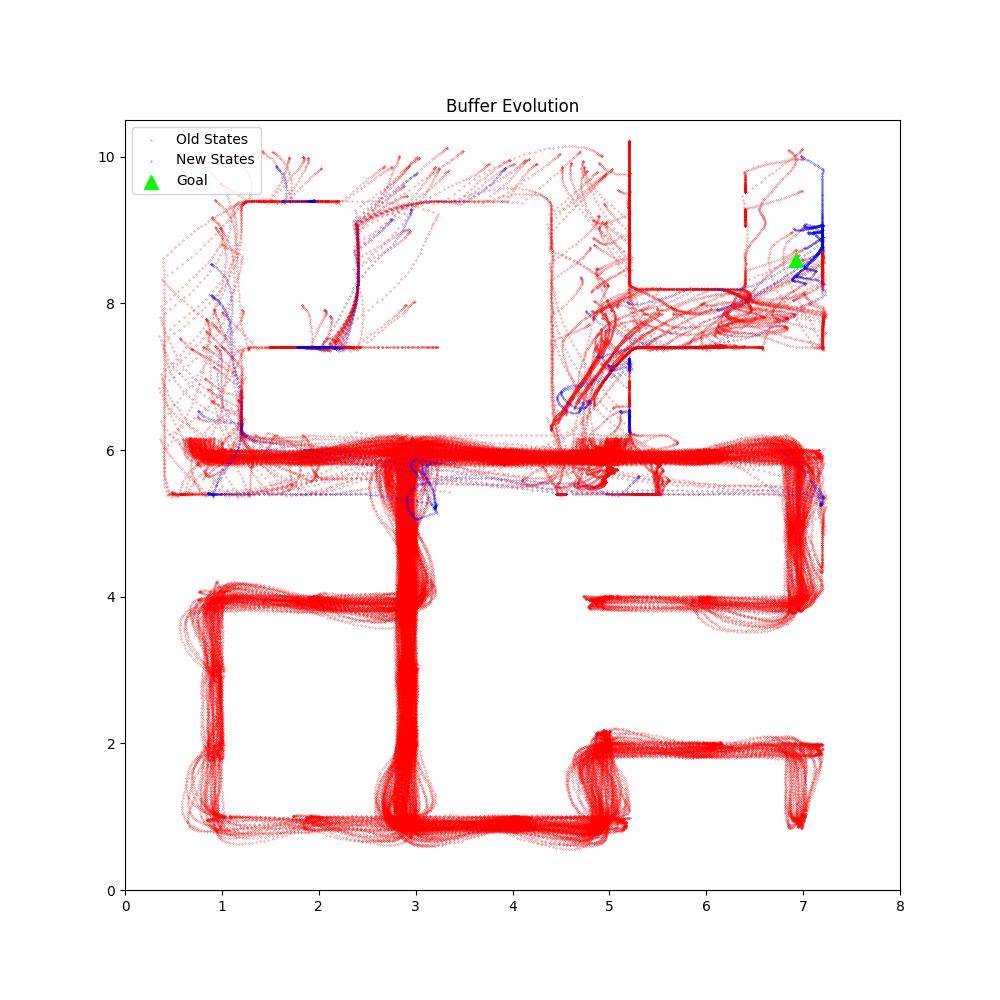}
    \end{minipage}
    \begin{minipage}{0.20\textwidth}
        \centering
        \includegraphics[trim=4cm 3cm 6.8cm 3.2cm, clip=true, width=\textwidth, height=1.4\textwidth]{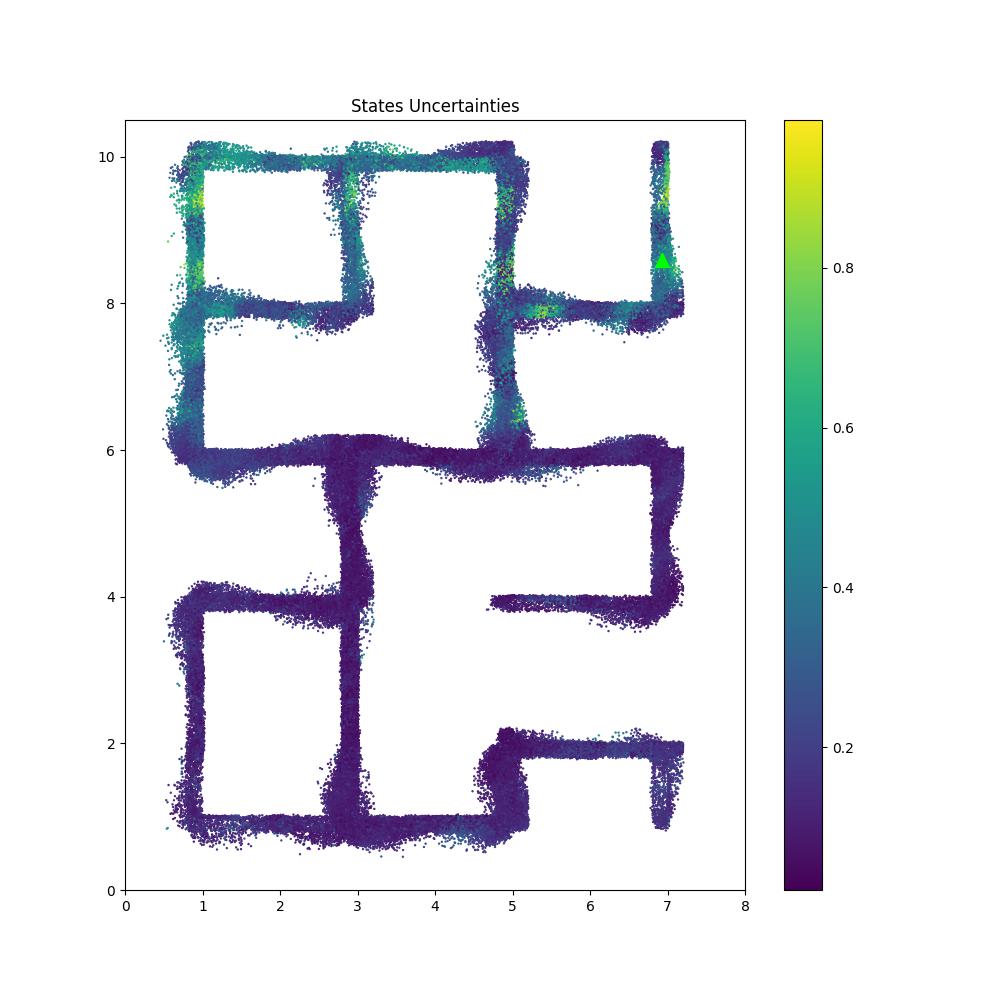}
    \end{minipage}
    \begin{minipage}{0.20\textwidth}
        \centering
        \includegraphics[trim=4cm 3cm 2.7cm 3.2cm, clip=true, width=\textwidth, height=1.4\textwidth]{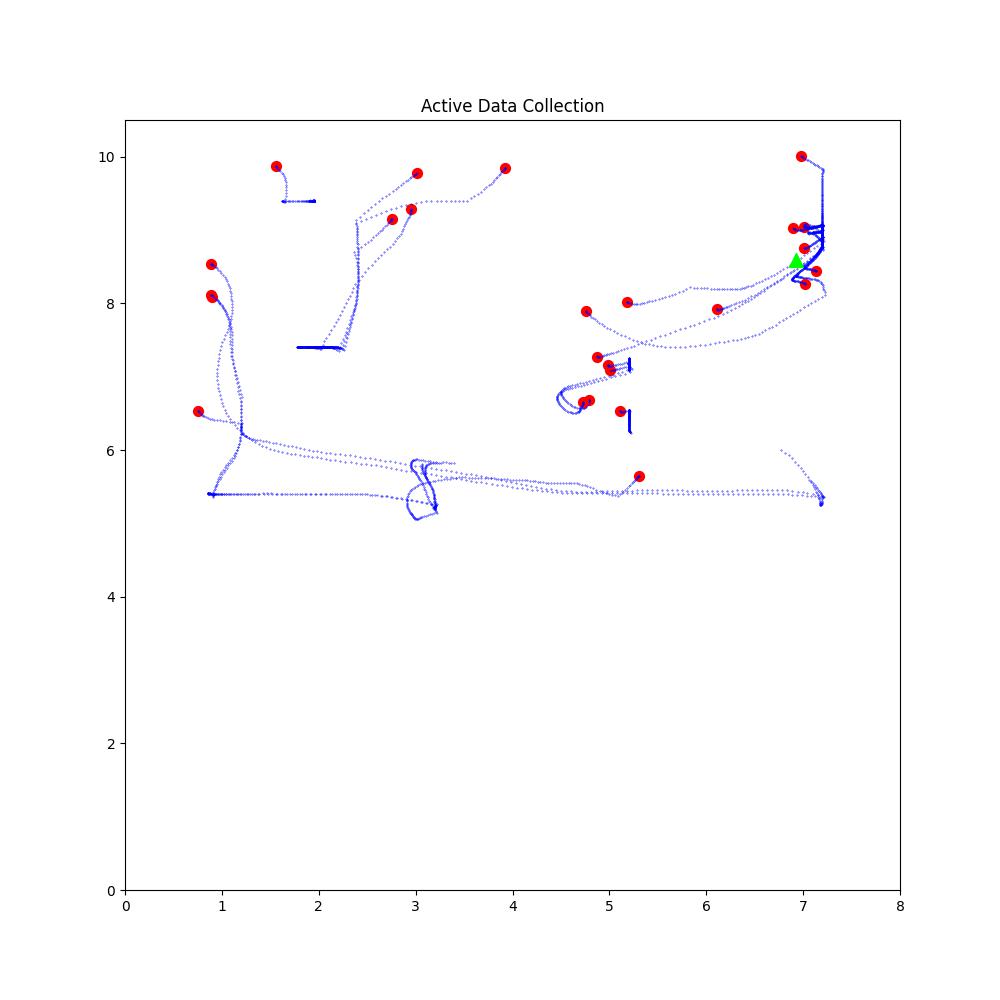}
    \end{minipage}
    \begin{minipage}{0.20\textwidth}
        \centering
        \includegraphics[trim=4cm 2.9cm 2.7cm 3.2cm, clip=true, width=\textwidth, height=1.4\textwidth]{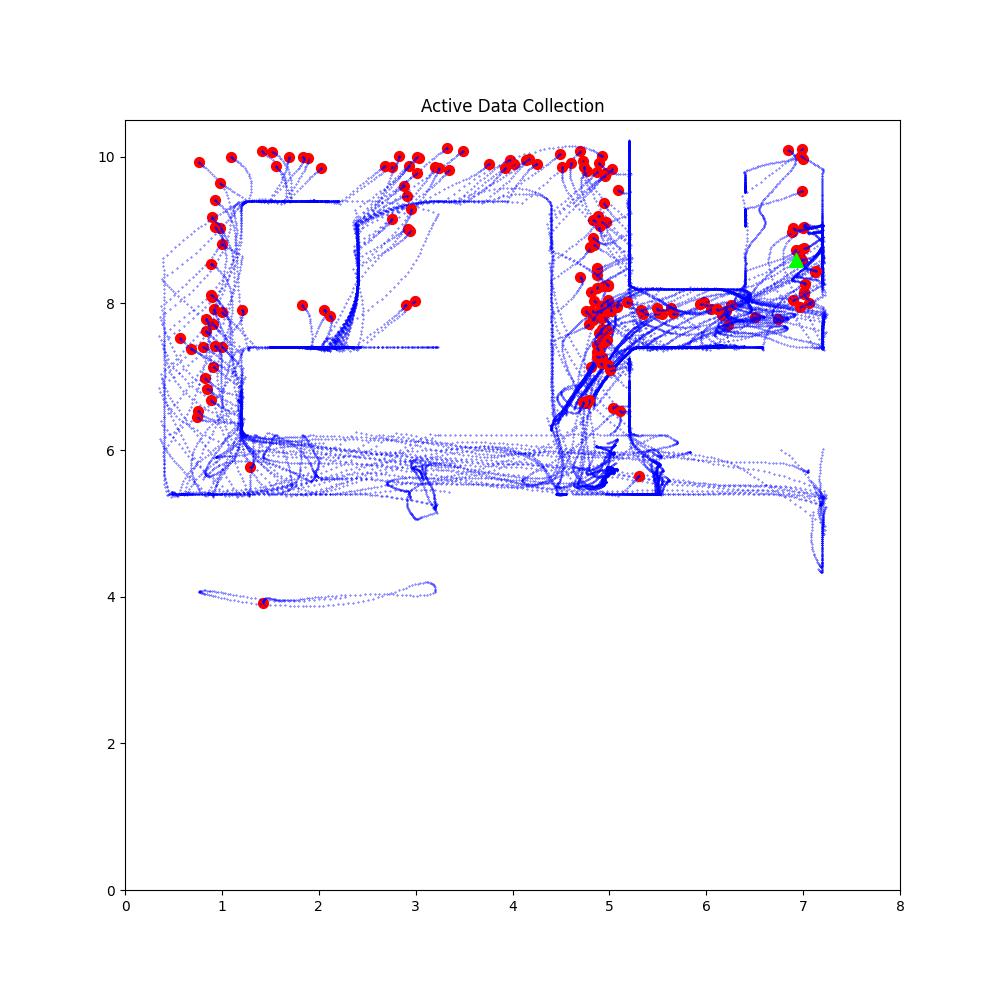}
    \end{minipage}


    \begin{minipage}{0.20\textwidth}
        \centering
        \includegraphics[trim=4cm 3cm 2.7cm 3.2cm, clip=true, width=\textwidth, height=1.4\textwidth]{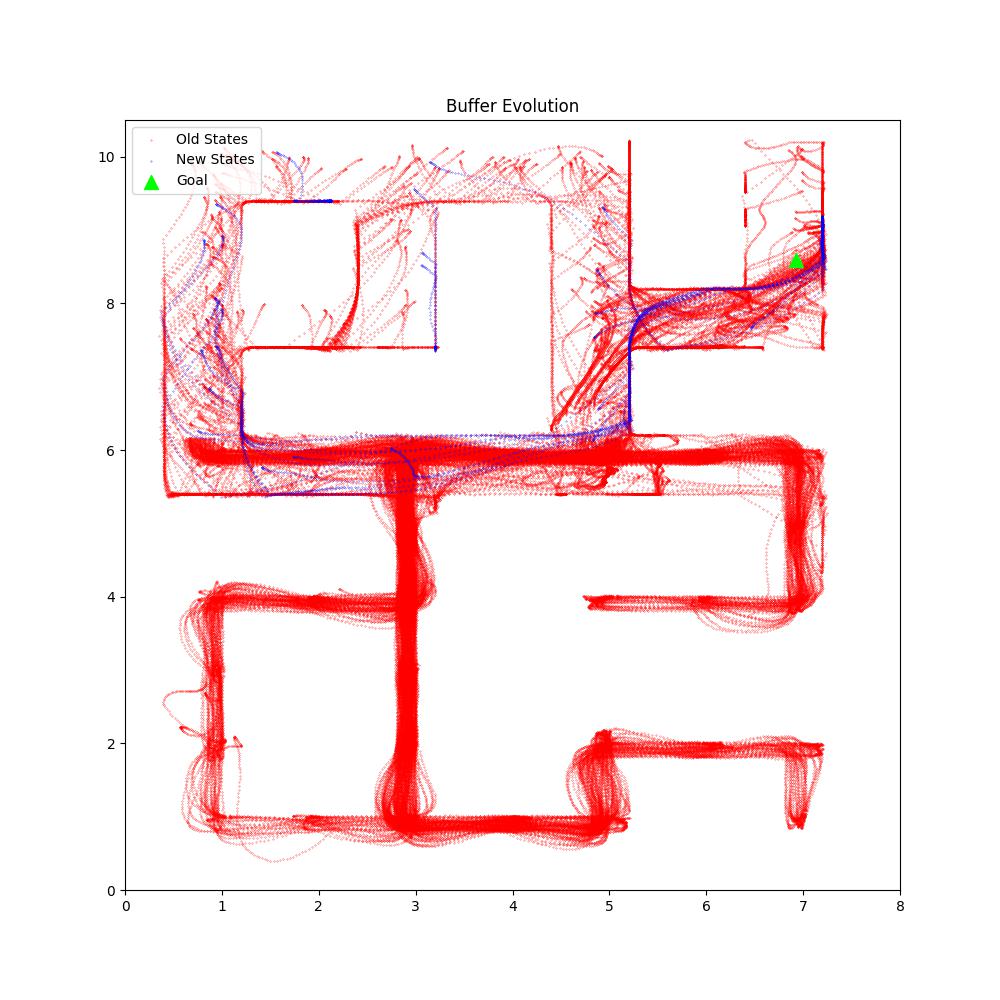}
    \end{minipage}
    \begin{minipage}{0.20\textwidth}
        \centering
        \includegraphics[trim=4cm 3cm 6.8cm 3.2cm, clip=true, width=\textwidth, height=1.4\textwidth]{resources/appendix_resources/our_evolution/uncertainty_6.jpg}
    \end{minipage}
    \begin{minipage}{0.20\textwidth}
        \centering
        \includegraphics[trim=4cm 3cm 2.7cm 3.2cm, clip=true, width=\textwidth, height=1.4\textwidth]{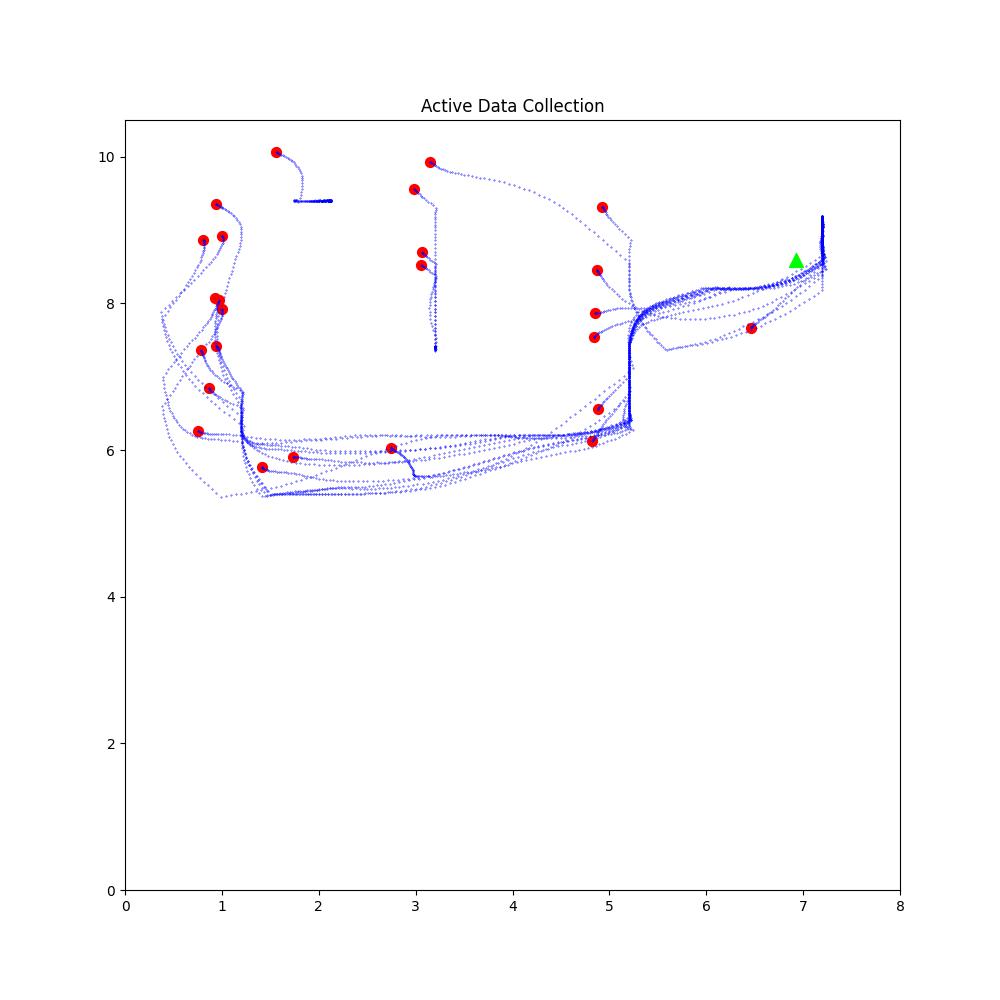}
    \end{minipage}
    \begin{minipage}{0.20\textwidth}
        \centering
        \includegraphics[trim=4cm 2.9cm 2.7cm 3.2cm, clip=true, width=\textwidth, height=1.4\textwidth]{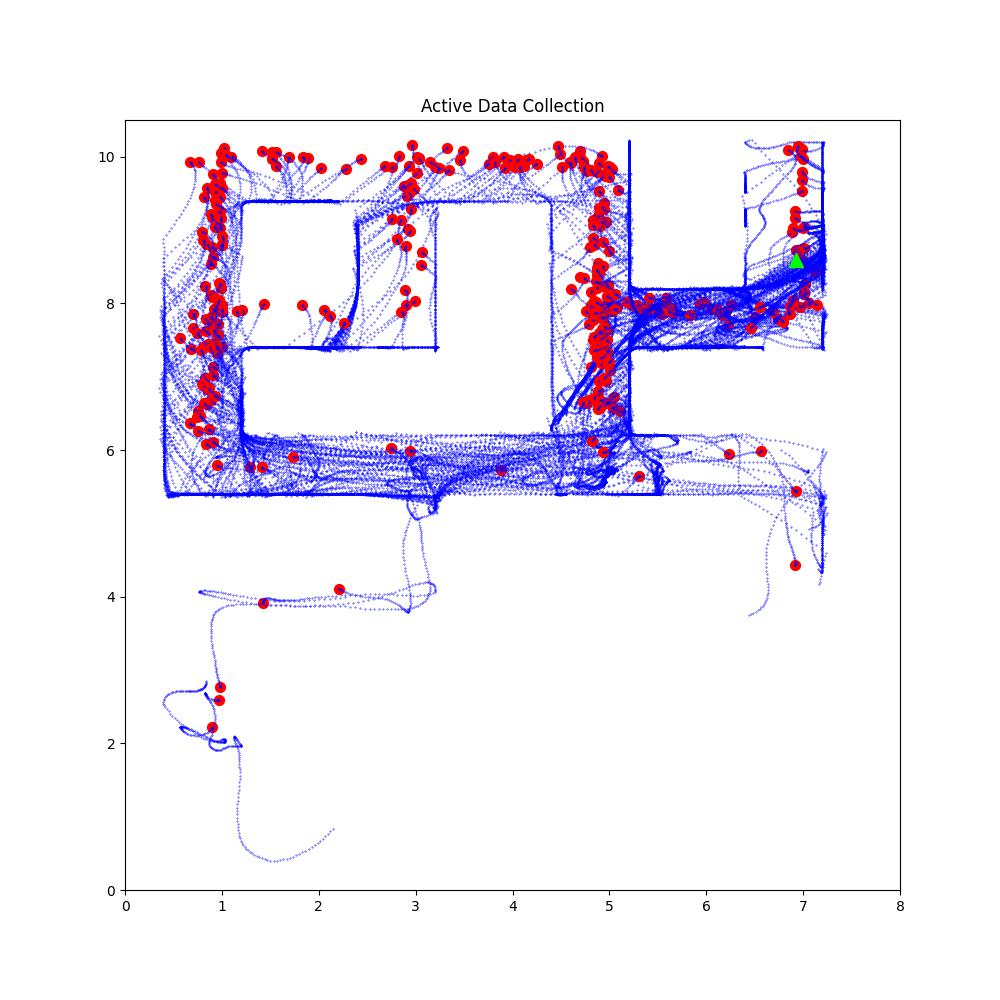}
    \end{minipage}


    \begin{minipage}{0.20\textwidth}
        \centering
        \includegraphics[trim=4cm 3cm 2.7cm 3.2cm, clip=true, width=\textwidth, height=1.4\textwidth]{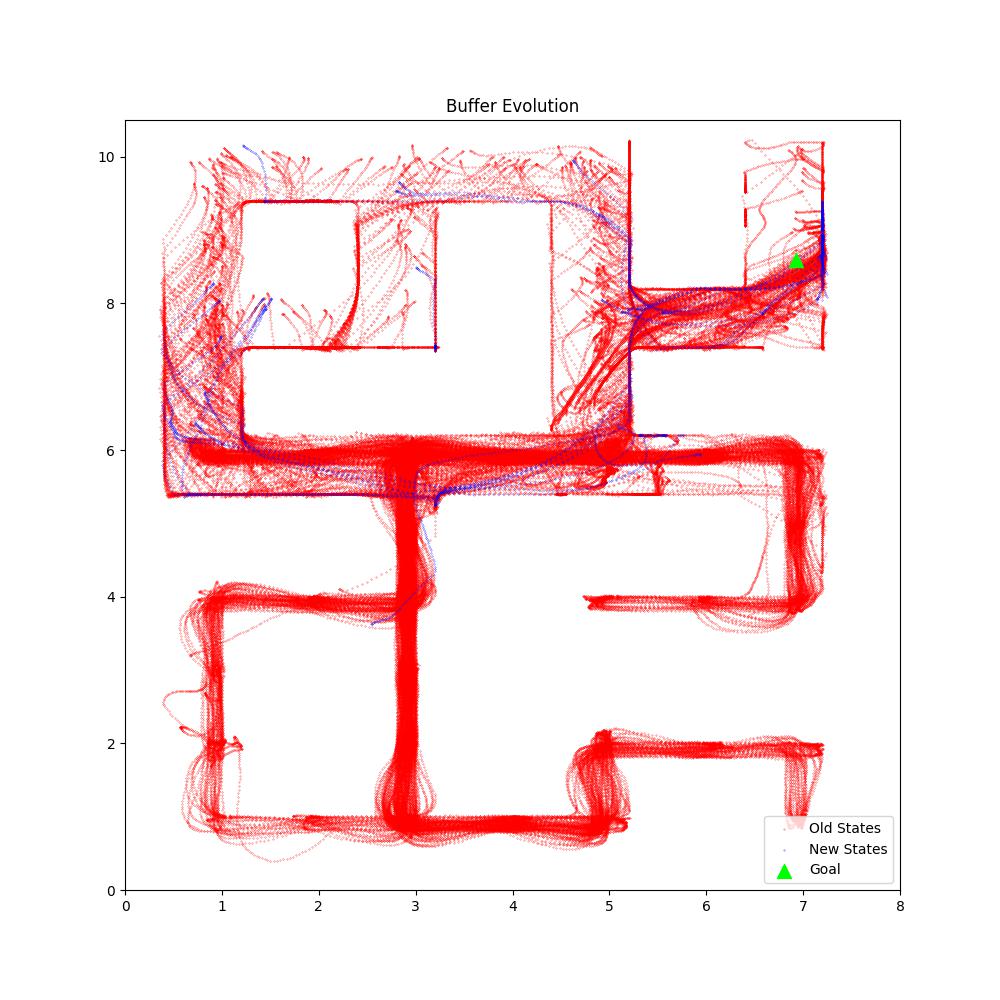}
    \end{minipage}
    \begin{minipage}{0.20\textwidth}
        \centering
        \includegraphics[trim=4cm 3cm 6.8cm 3.2cm, clip=true, width=\textwidth, height=1.4\textwidth]{resources/appendix_resources/our_evolution/uncertainty_6.jpg}
    \end{minipage}
    \begin{minipage}{0.20\textwidth}
        \centering
        \includegraphics[trim=4cm 3cm 2.7cm 3.2cm, clip=true, width=\textwidth, height=1.4\textwidth]{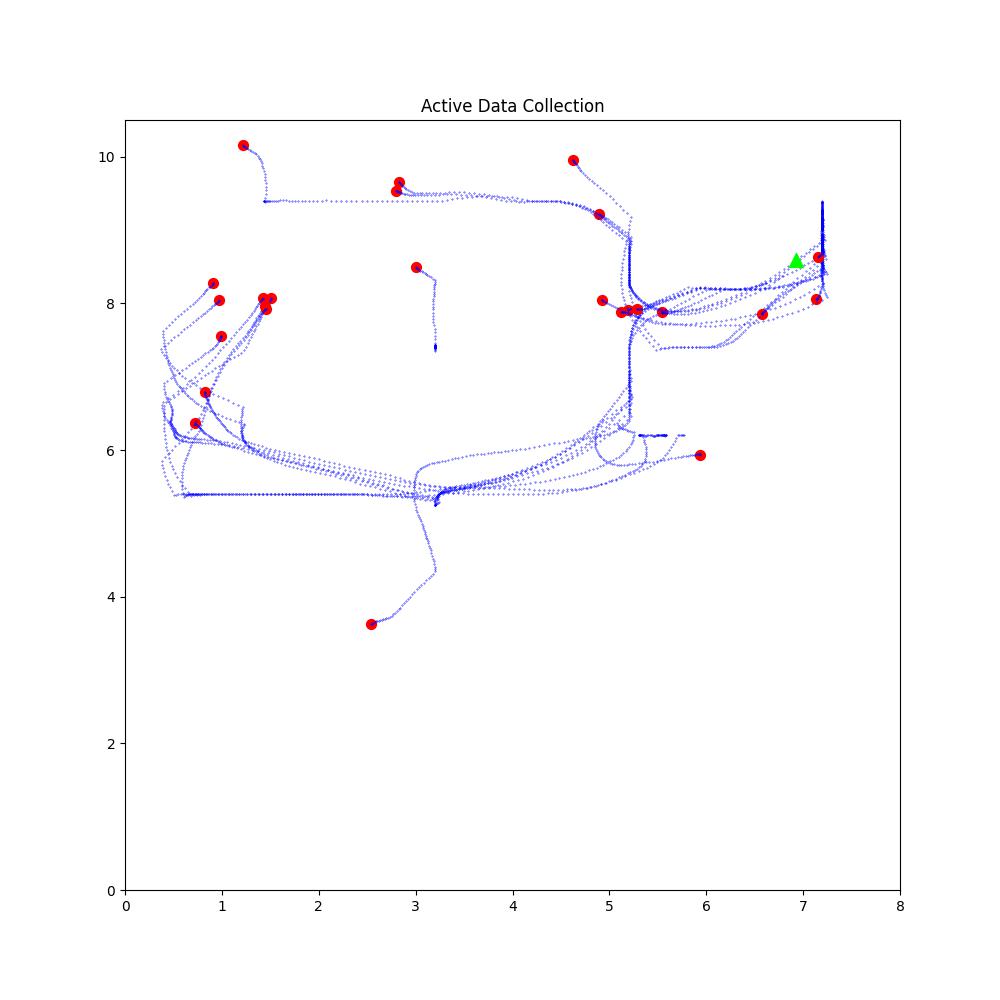}
    \end{minipage}
    \begin{minipage}{0.20\textwidth}
        \centering
        \includegraphics[trim=4cm 2.9cm 2.7cm 3.2cm, clip=true, width=\textwidth, height=1.4\textwidth]{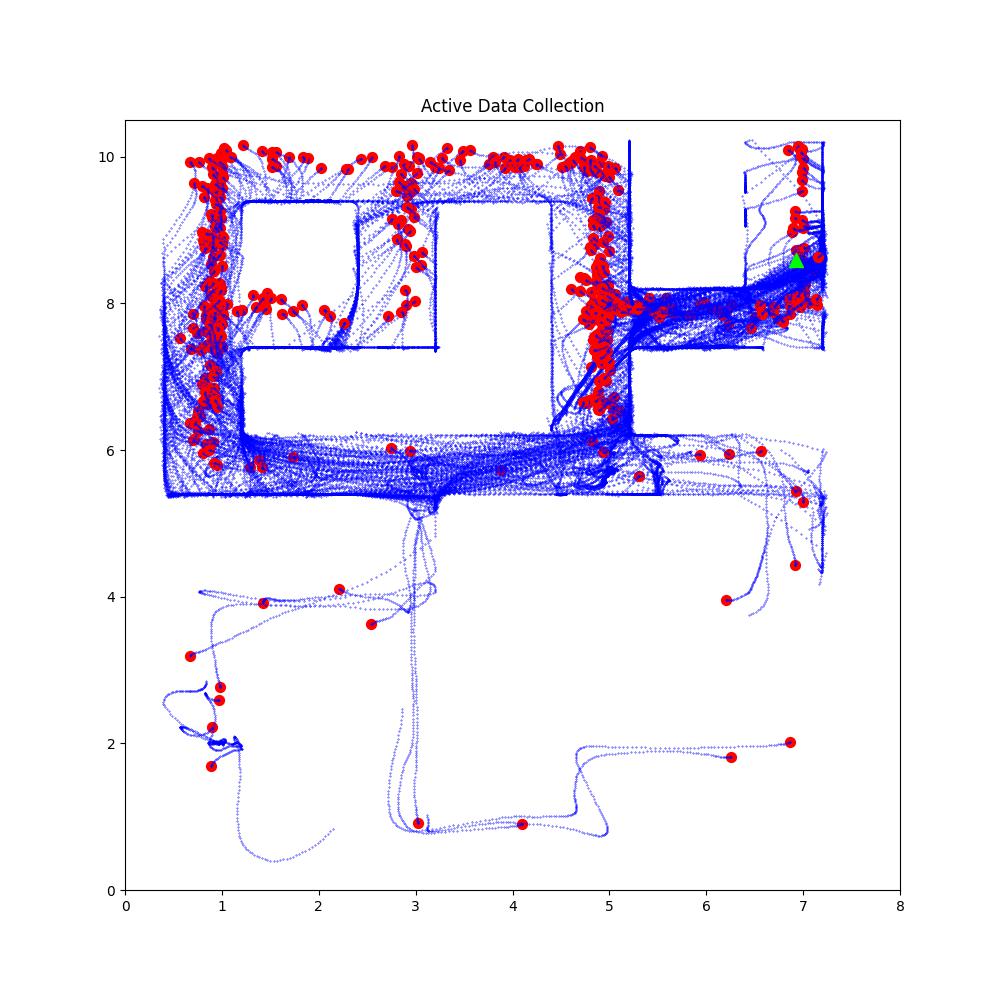}
    \end{minipage}
    
    \caption{[Best viewed in color] Demonstration of the evolution of the dataset during online trajectory collection at regular intervals using ActiveORL. In the first row, the first image on the left is the initial offline dataset. To it's right, we plot the environment-aware uncertainty. It can be observed that uncertainty is high in regions where the data was not available in the offline dataset. To its right, we show the active trajectories collected in the very first epoch of data collection. The last picture (on the right) shows the cumulative trajectories collected. In all the remaining rows, we similarly see the current state of the offline buffer, the corresponding uncertainty, the trajectories collected in that step, and the cumulative trajectories collected, respectively. The second, third, and fourth rows correspond to the sixth, thirteenth, and twentieth epochs of data collection, respectively. Notice that the ActiveORL trajectory collection focuses primarily in the unobserved region of the offline dataset.}
    \label{Figure: OurEvolution}
\end{figure*}

\begin{figure*}[t]
    \centering

    \begin{minipage}{0.20\textwidth}
        \centering
        \includegraphics[trim=4cm 3cm 2.7cm 3.2cm, clip=true, width=\textwidth, height=1.4\textwidth]{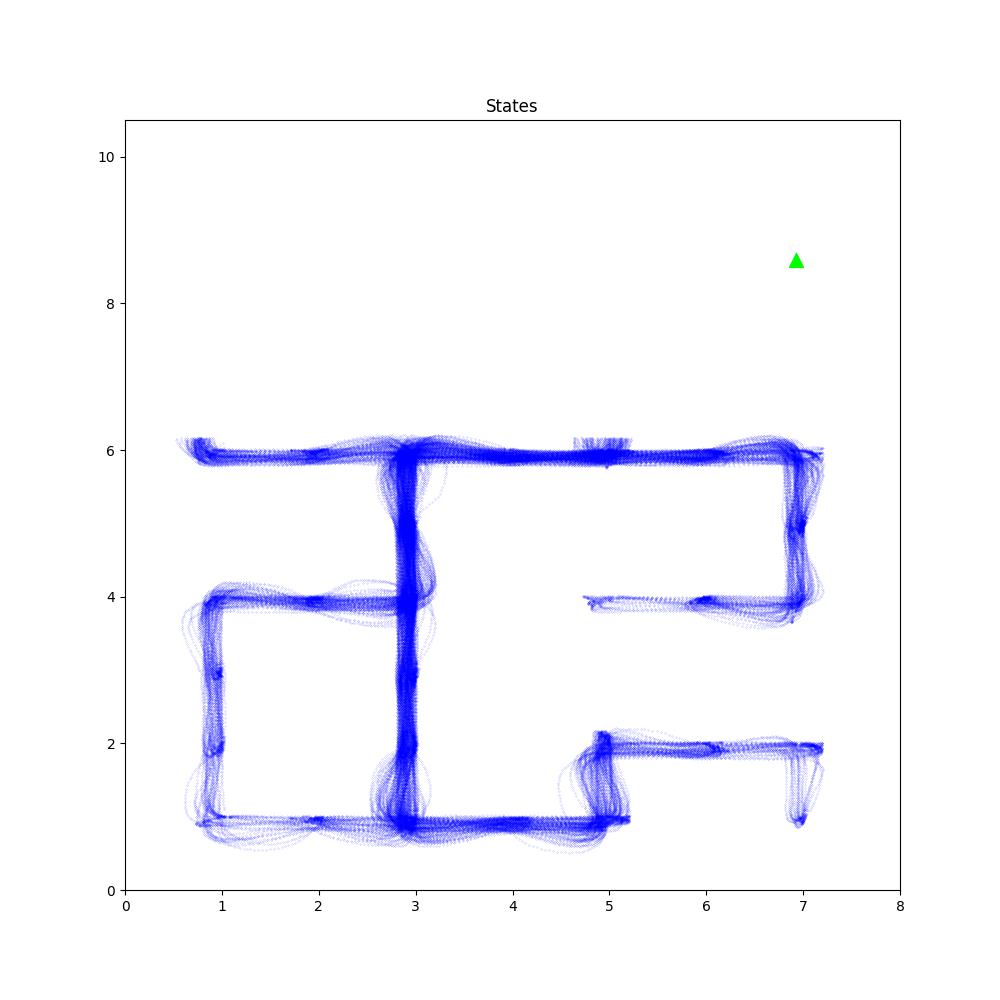}
    \end{minipage}
    \begin{minipage}{0.20\textwidth}
        \centering
        \includegraphics[trim=4cm 3cm 2.7cm 3.2cm, clip=true, width=\textwidth, height=1.4\textwidth]{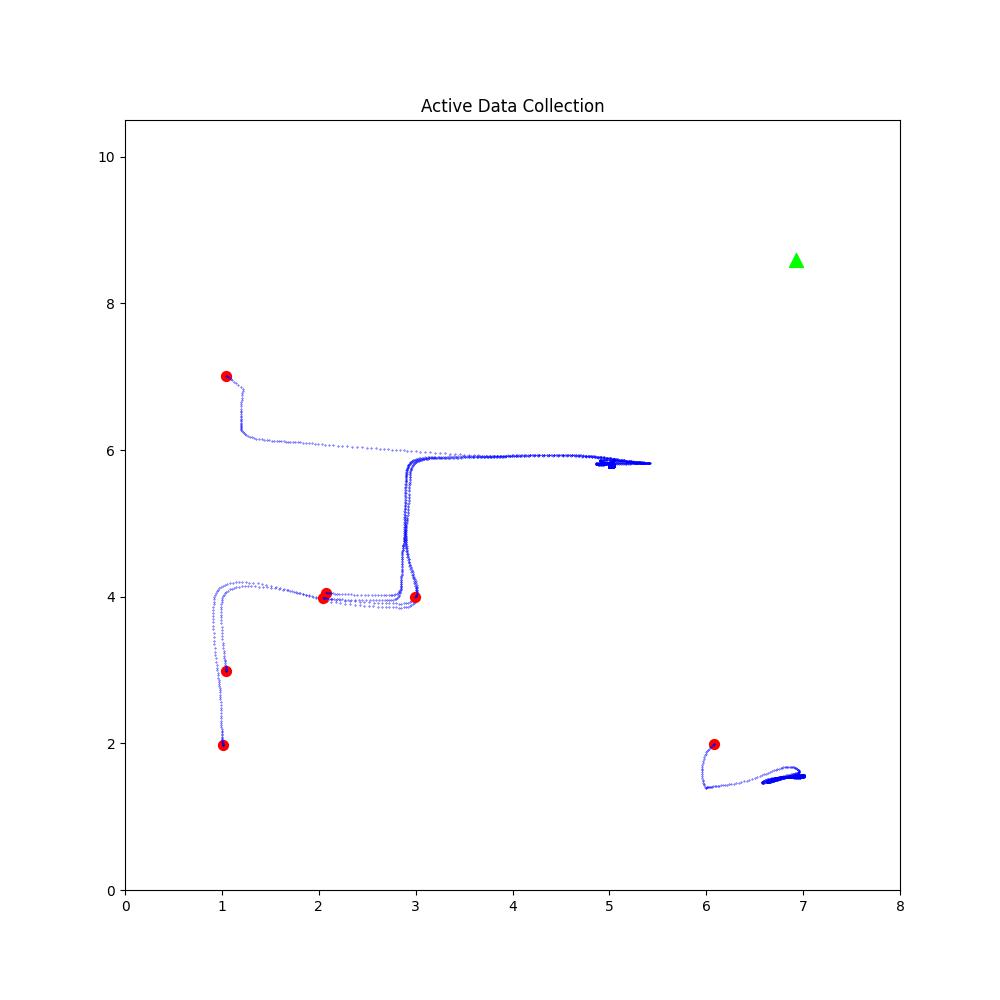}
    \end{minipage}
    \begin{minipage}{0.20\textwidth}
        \centering
        \includegraphics[trim=4cm 2.9cm 2.7cm 3.2cm, clip=true, width=\textwidth, height=1.4\textwidth]{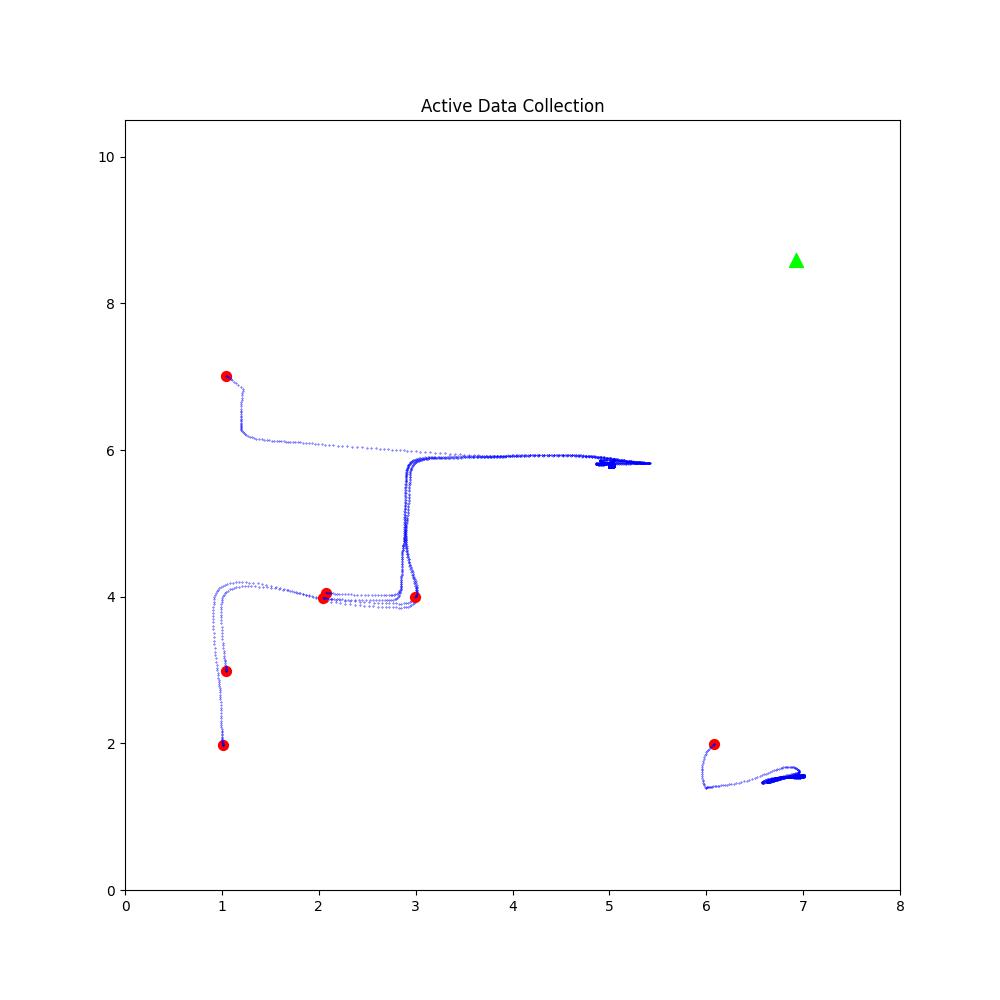}
    \end{minipage}


    \begin{minipage}{0.20\textwidth}
        \centering
        \includegraphics[trim=4cm 3cm 2.7cm 3.2cm, clip=true, width=\textwidth, height=1.4\textwidth]{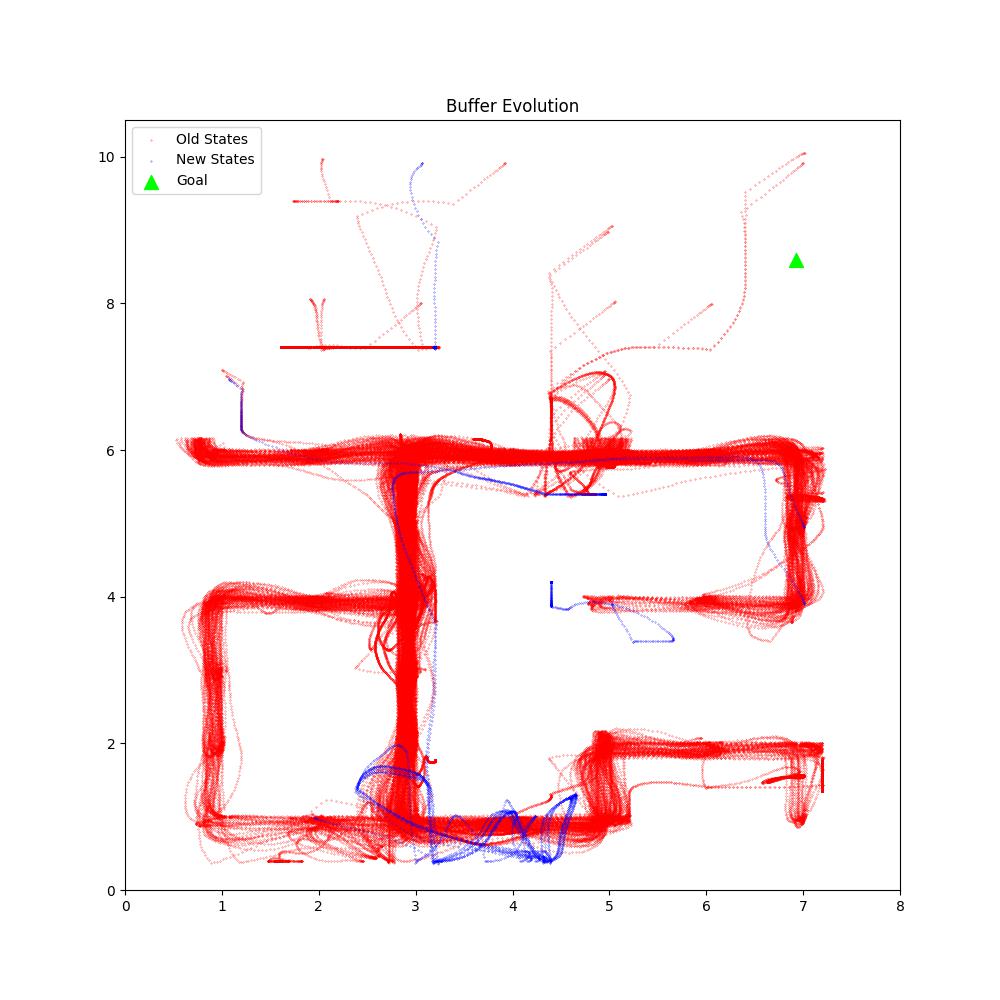}
    \end{minipage}
    \begin{minipage}{0.20\textwidth}
        \centering
        \includegraphics[trim=4cm 3cm 2.7cm 3.2cm, clip=true, width=\textwidth, height=1.4\textwidth]{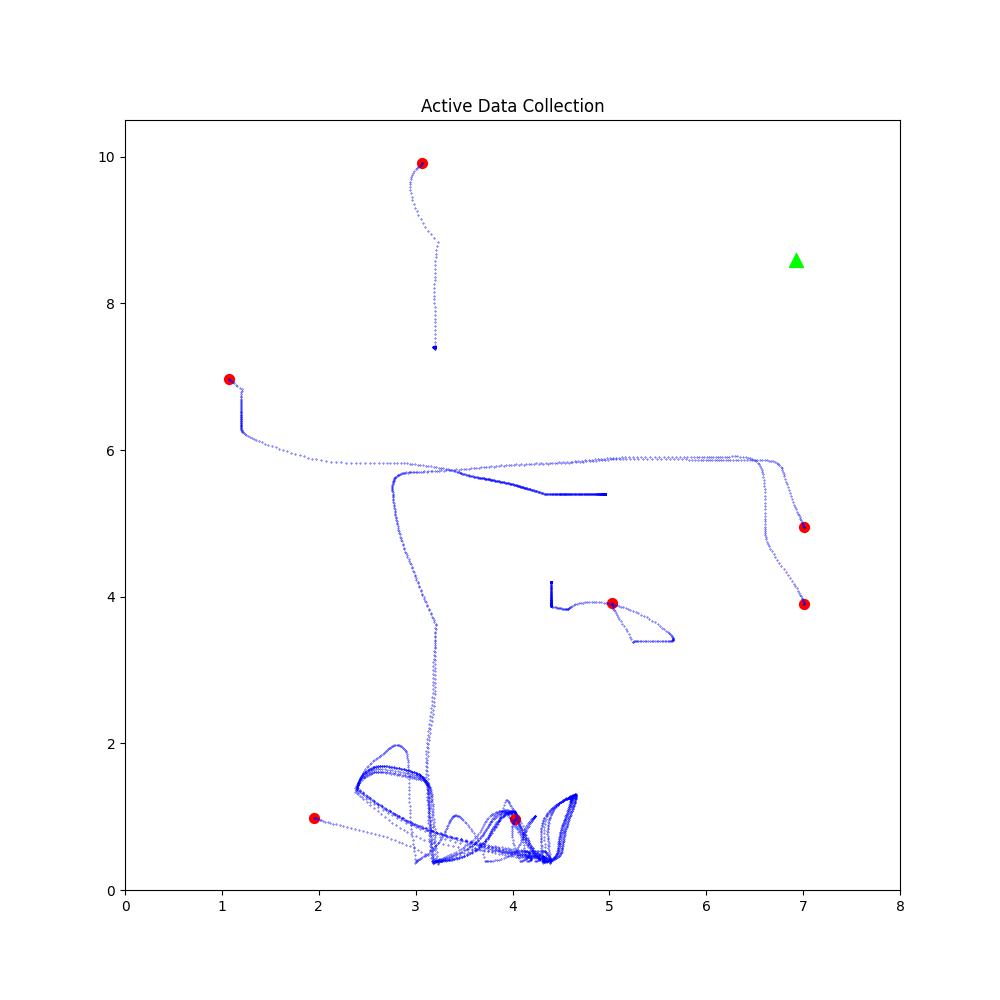}
    \end{minipage}
    \begin{minipage}{0.20\textwidth}
        \centering
        \includegraphics[trim=4cm 2.9cm 2.7cm 3.2cm, clip=true, width=\textwidth, height=1.4\textwidth]{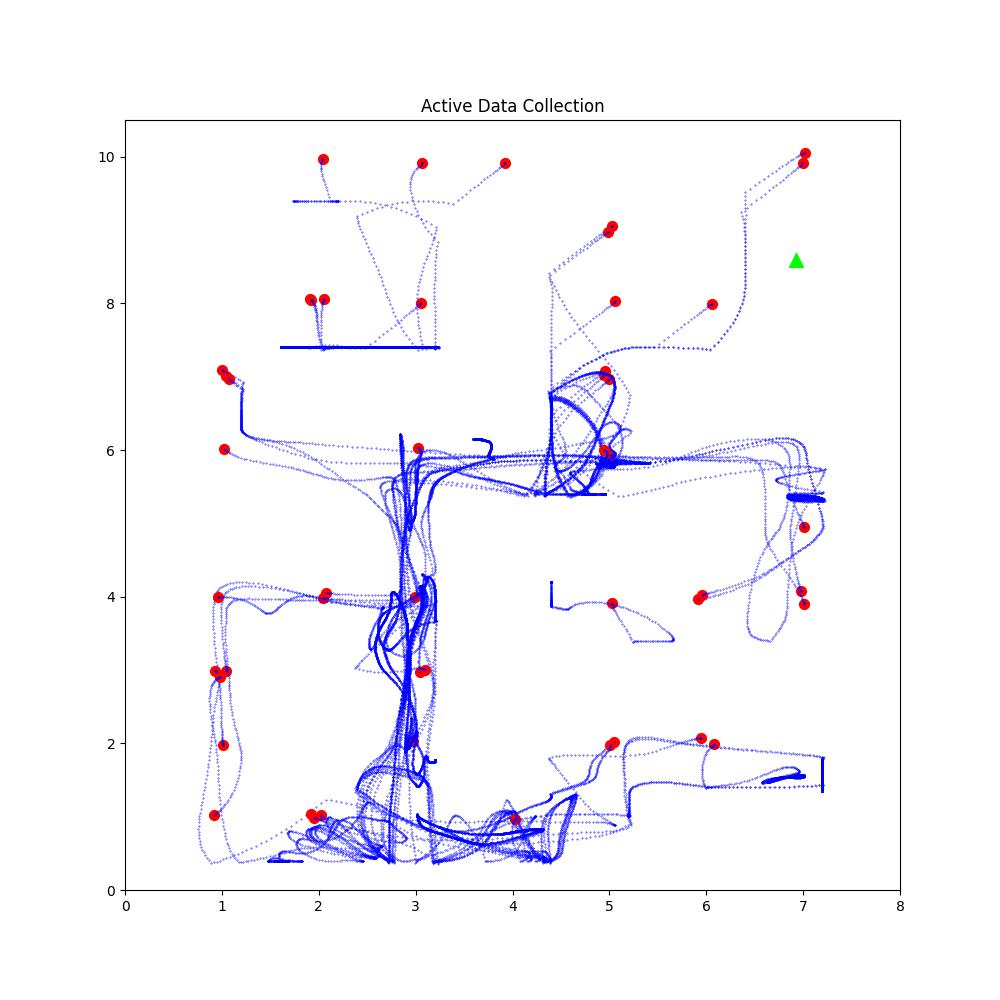}
    \end{minipage}


    \begin{minipage}{0.20\textwidth}
        \centering
        \includegraphics[trim=4cm 3cm 2.7cm 3.2cm, clip=true, width=\textwidth, height=1.4\textwidth]{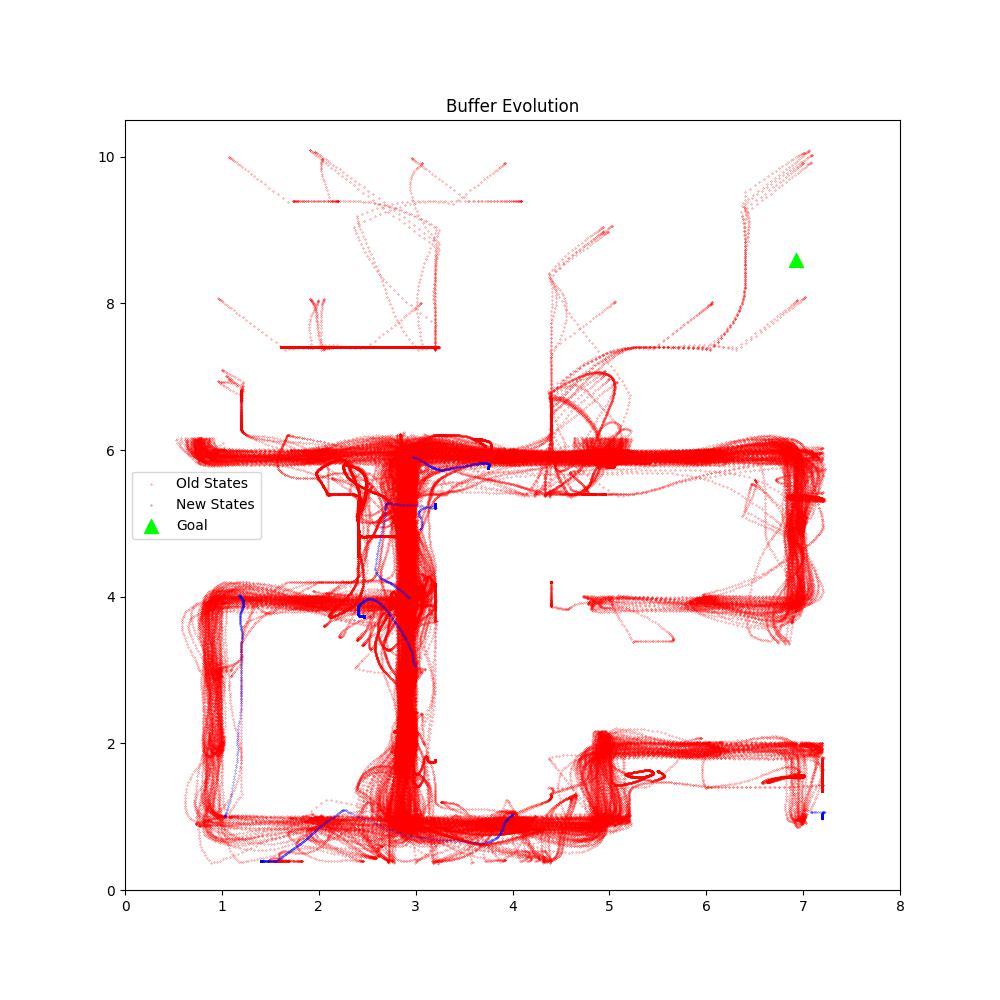}
    \end{minipage}
    \begin{minipage}{0.20\textwidth}
        \centering
        \includegraphics[trim=4cm 3cm 2.7cm 3.2cm, clip=true, width=\textwidth, height=1.4\textwidth]{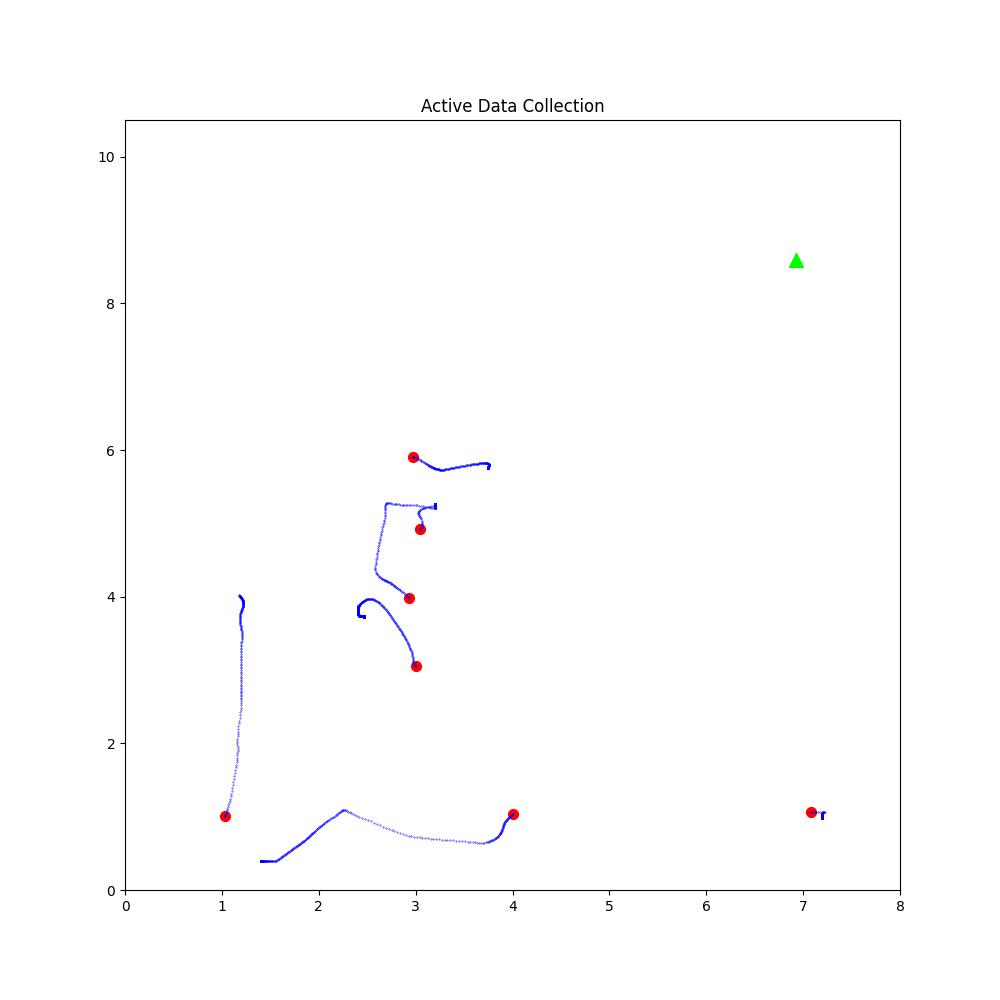}
    \end{minipage}
    \begin{minipage}{0.20\textwidth}
        \centering
        \includegraphics[trim=4cm 2.9cm 2.7cm 3.2cm, clip=true, width=\textwidth, height=1.4\textwidth]{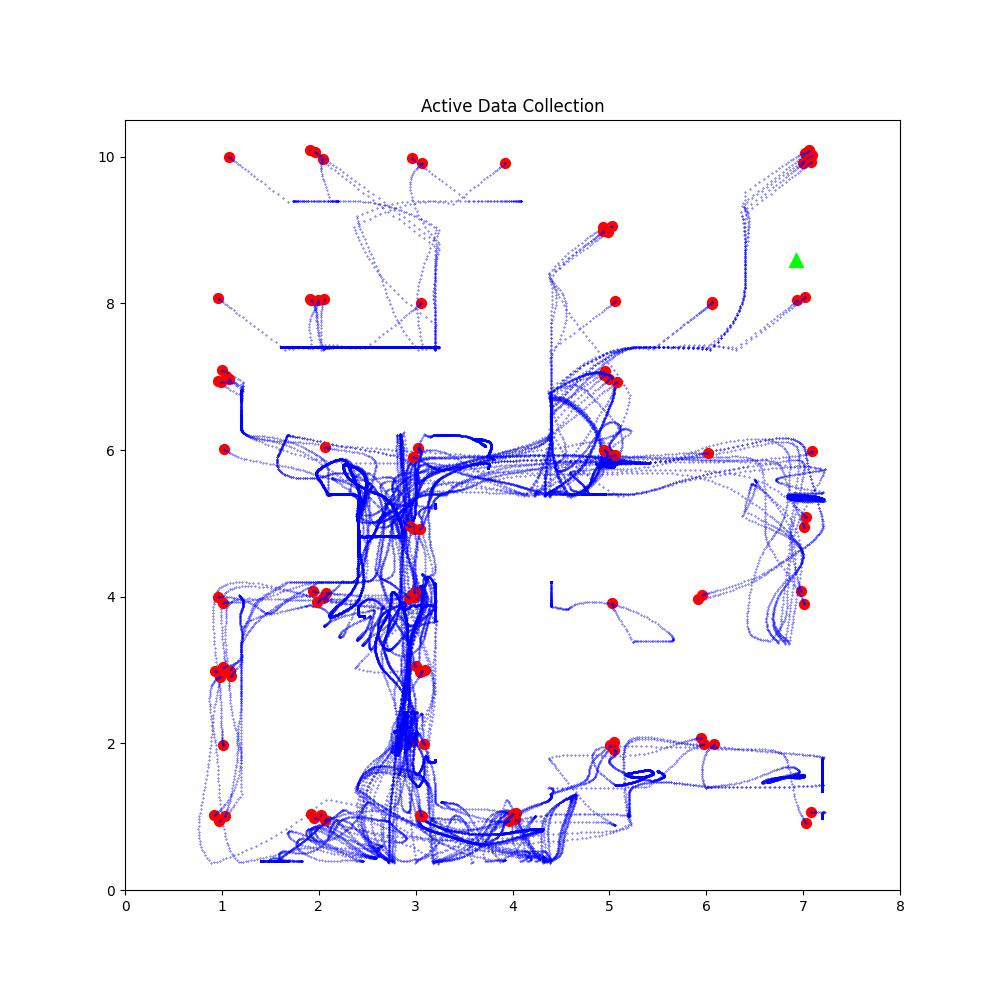}
    \end{minipage}


    \begin{minipage}{0.20\textwidth}
        \centering
        \includegraphics[trim=4cm 3cm 2.7cm 3.2cm, clip=true, width=\textwidth, height=1.4\textwidth]{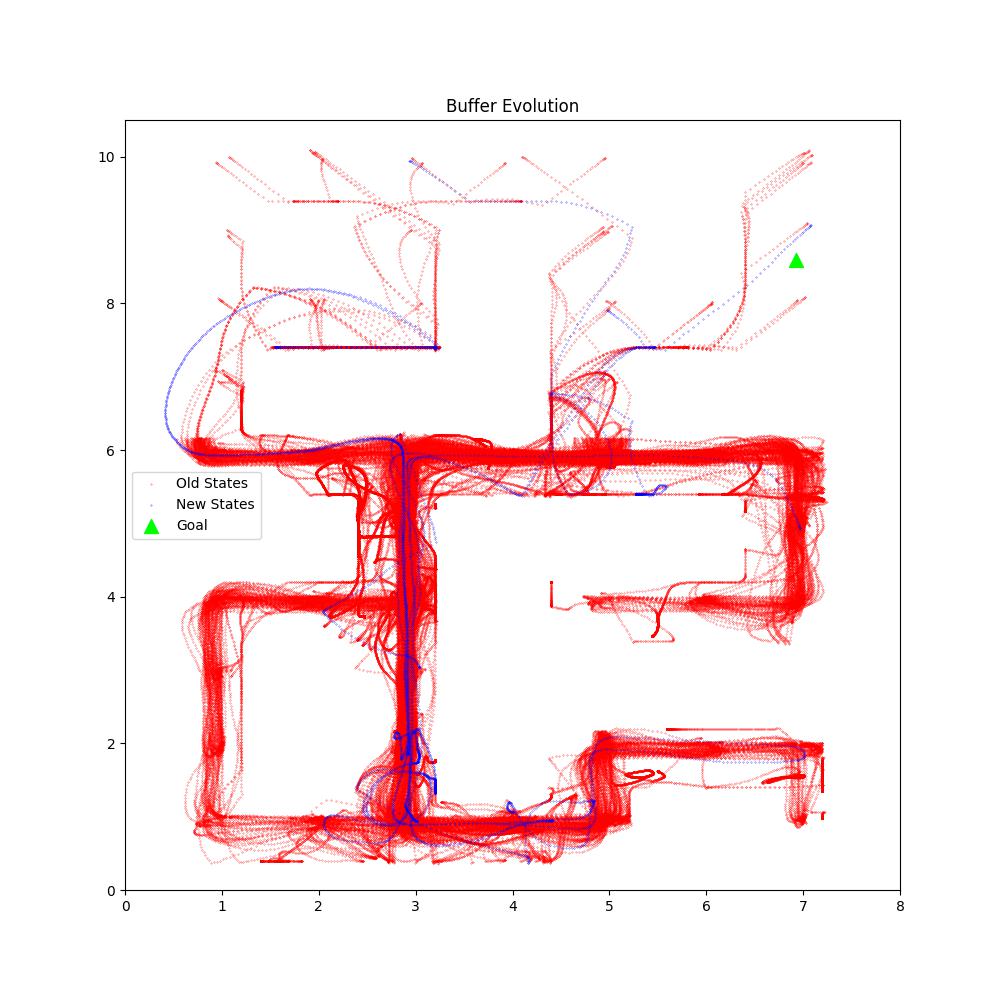}
    \end{minipage}
    \begin{minipage}{0.20\textwidth}
        \centering
        \includegraphics[trim=4cm 3cm 2.7cm 3.2cm, clip=true, width=\textwidth, height=1.4\textwidth]{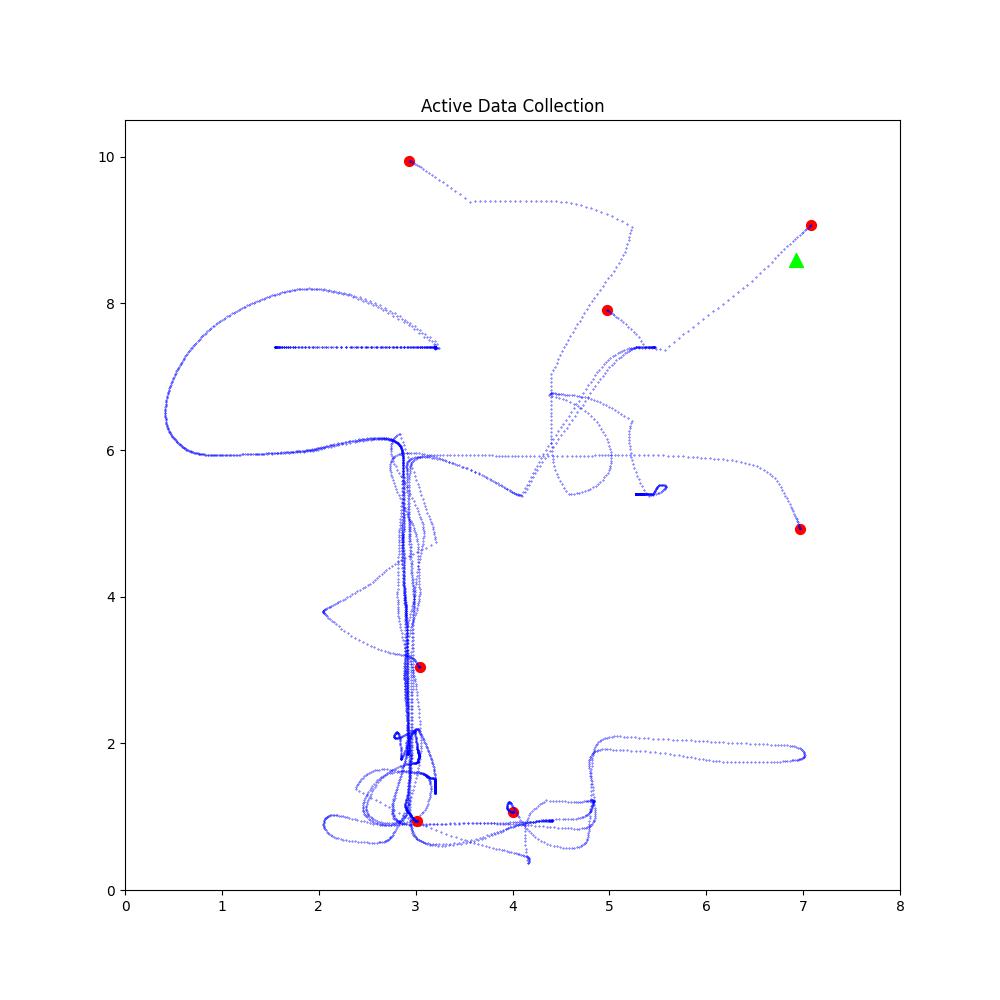}
    \end{minipage}
    \begin{minipage}{0.20\textwidth}
        \centering
        \includegraphics[trim=4cm 2.9cm 2.7cm 3.2cm, clip=true, width=\textwidth, height=1.4\textwidth]{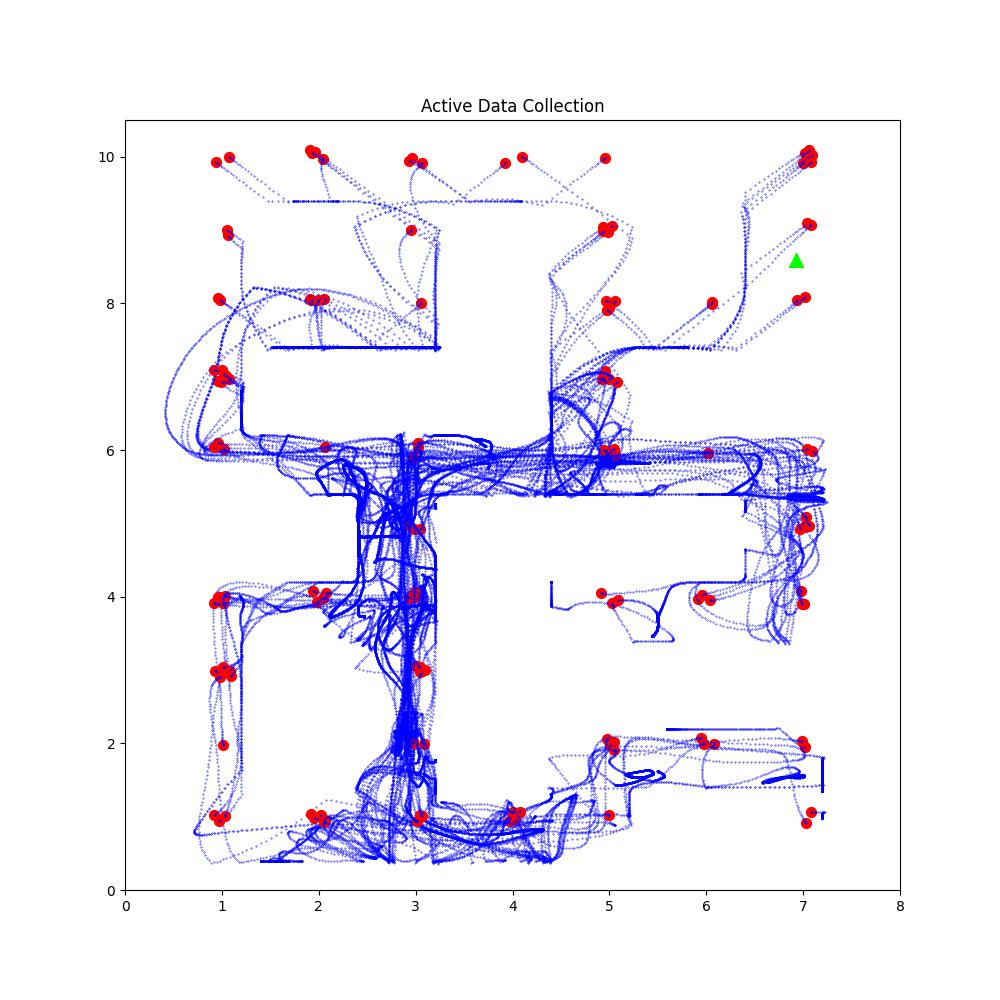}
    \end{minipage}
    
    \caption{[Best viewed in color] Demonstration of the evolution of the dataset during online trajectory collection at regular intervals using baseline fine-tuning methods. The columns have the same meaning as Figure \ref{Figure: OurEvolution}, except of course the uncertainty map (which does not apply in this case). As can be noticed, in contrast to ActiveORL, other methods keep collection redundant trajectories during the online data collection phase and hence do not perform very well, especially if the majority of the reward is in the unexplored regions of the offline dataset. Also, notice that even though starting from near the goal location (by chance), the trajectory collection algorithm just reverts back to the confident region of the dataset, when in fact, exploration should be towards the regions where there is less confidence (as shown in Figure \ref{Figure: OurEvolution}). Furthermore, it should also be noticed that the number of initial states (denoted by red points) is less in this figure when compared to Figure \ref{Figure: OurEvolution} because ActiveORL stops collecting data when confident regions are reached to stop wastage of online interaction; however, no such early stopping mechanism exist in other methods.}
    \label{Figure: BaselineEvolution}
\end{figure*}

\subsection{Fixed initial state distribution}

For the setting in which the initial state is not in the agent's control, we consider a two-stage policy that first travels to the best candidate's initial state and then begins exploring to collect useful samples.

The first stage uses a ``goal-based'' agent trained on the offline dataset that takes coordinates as input along with the current state and reaches the goal using the shortest path. To train this policy, we sample sub-trajectories from the offline dataset of length $32$ and use the last state in each sub-trajectory as the ``goal'' to train a transformer-based model. We provide a unit reward at the goal state and a discounting factor of 0.99 to calculate the values of each state under the given action sampled from the dataset. We train it to maximize over the actions using TD3+BC offline algorithm.

For the distance-weighted graph of states in the offline dataset, we make the edge matrix sparse by removing edges with weight less than $10^{-2}$ and perform hierarchical clustering on the offline dataset using modularity-based heuristic in the Louvain algorithm~\cite{Blondel_2008}. Other methods are very computationally expensive, for instance spectral clustering has $O(n^3)$ time complexity, and hence cannot be used for our dataset which has $\sim100$K transitions.

During the first stage of active collection, we identify the candidate's initial state with the most uncertainty and run the aforementioned ``goal-based'' policy to reach a randomly sampled state in that cluster nearest to this optimal candidate state. Once the agent has reached a point closest to the most uncertain initial state, we switch over to our exploration policy as described in Section~\ref{Section: ActiveORL}.

We run the aforementioned experiments on the maze2d environment and plot the results in Figure \ref{Figure: GoalAblation}. Even in this restricted setting, we still gain a clear advantage over the baseline ``uncertainty unaware'' methods.



\begin{figure*}[t]
    \centering
    \includegraphics[trim=5.5cm 8cm 9cm 4.5cm, clip=true, width=\textwidth]{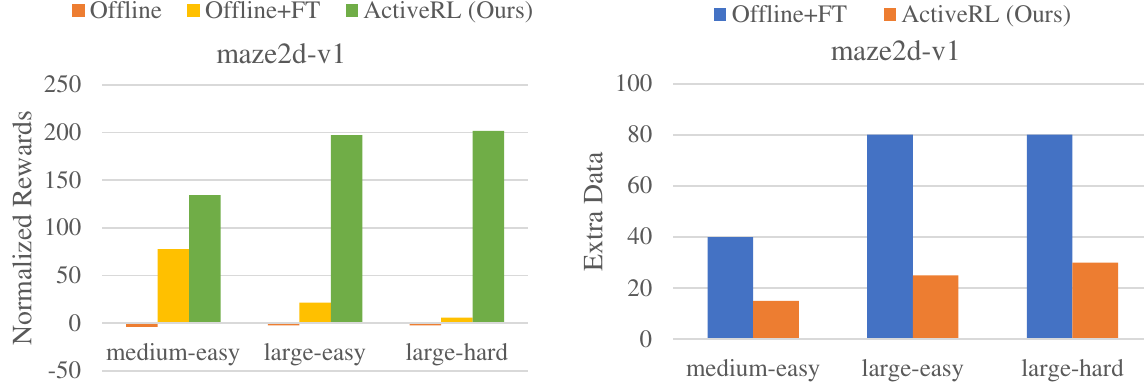}\\
    \vspace{1cm}
    \includegraphics[trim=6cm 8.5cm 9cm 4.5cm, clip=true, width=\textwidth]{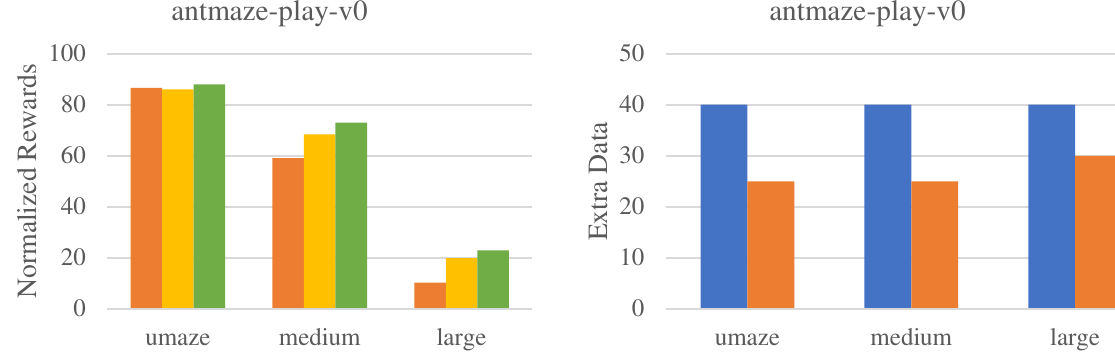}\\
    \vspace{1cm}
    \includegraphics[trim=6cm 8.5cm 9cm 4.5cm, clip=true, width=\textwidth]{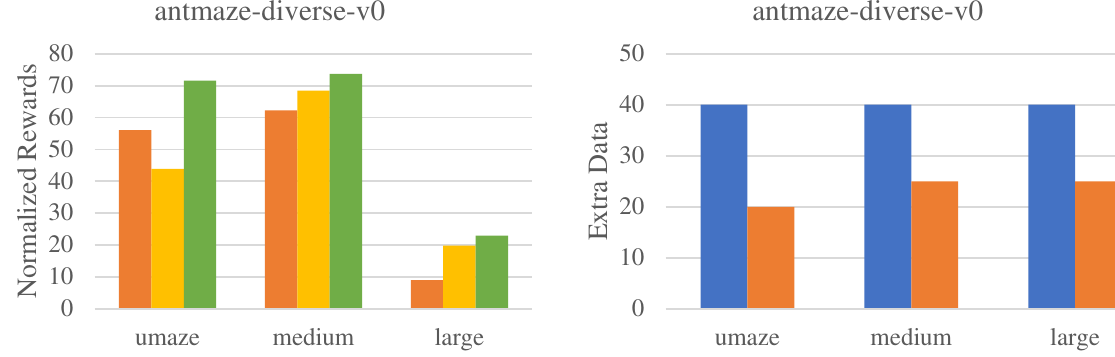}\\
    \caption{Bar plots in the maze-like environments comparing the performance and data reduction. The first column shows the performance on the normalized scale corresponding to the offline algorithm, fine-tuning algorithm, and ActiveORL algorithm. The second column shows the number of epochs of online data collection required to reach the same performance (maximum performance of the worse-performing policy). The first row corresponds to the \texttt{maze2d} environments, while the next two rows correspond to the \texttt{antmaze} environments. It can be seen that overall, we achieved more performance with less data in the maze tasks.}
    \label{Figure: MazeBarPlots}
\end{figure*}

\begin{figure*}[t]
    \centering
    \includegraphics[trim=5.5cm 8cm 9cm 4.5cm, clip=true, width=\textwidth]{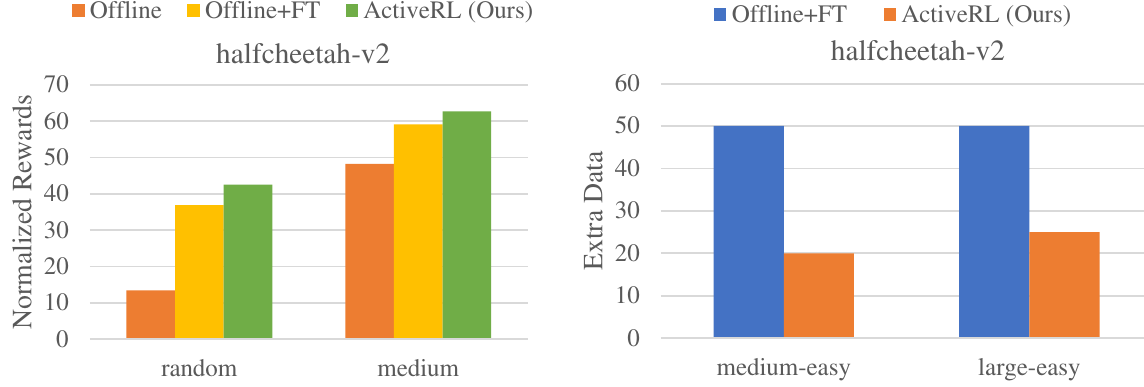}\\
    \vspace{1cm}
    \includegraphics[trim=6cm 8.5cm 9cm 4.5cm, clip=true, width=\textwidth]{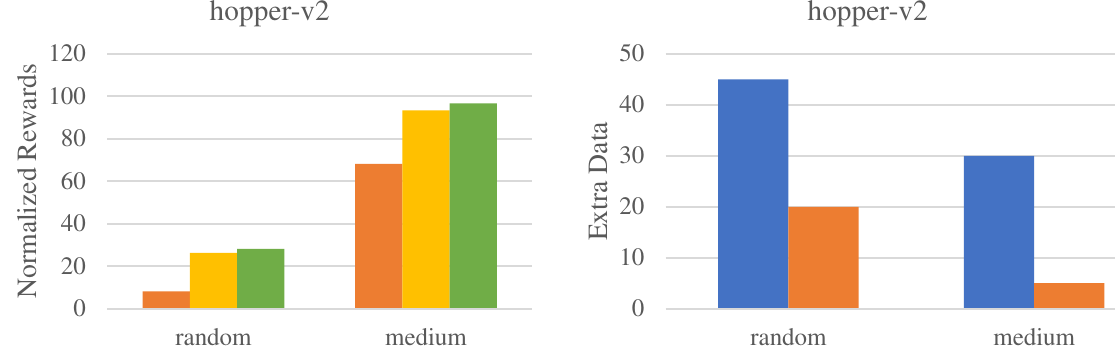}\\
    \vspace{1cm}
    \includegraphics[trim=6cm 8.5cm 9cm 4.5cm, clip=true, width=\textwidth]{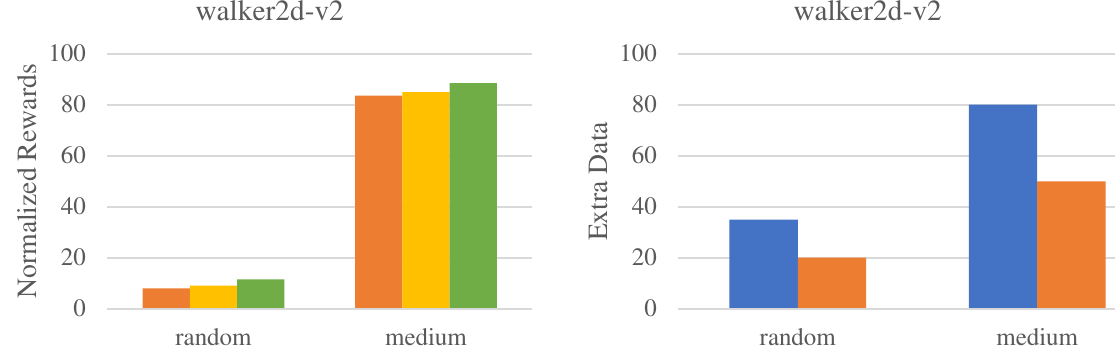}\\
    \caption{Bar plots in the locomotion environments comparing the performance improvement and data reduction. The columns in this figure follow the same convention as Figure \ref{Figure: MazeBarPlots}. The first row corresponds to the \texttt{halfcheetah} environment, the second row corresponds to the \texttt{hopper} environment, and the last row corresponds to the \texttt{walker2d} environment. It can be seen that even in the locomotion tasks, we achieve more performance with less data overall.}
    \label{Figure: LocomotionBarPlots}
\end{figure*}

%% file: anonymous-submission-latex-2025.bbl
\begin{thebibliography}{34}
\providecommand{\natexlab}[1]{#1}

\bibitem[{An et~al.(2021)An, Moon, Kim, and Song}]{an2021uncertainty}
An, G.; Moon, S.; Kim, J.-H.; and Song, H.~O. 2021.
\newblock Uncertainty-based offline reinforcement learning with diversified q-ensemble.
\newblock In \emph{Advances in Neural Information Processing Systems}, volume~34, 7436--7447.

\bibitem[{Bachman, Sordoni, and Trischler(2017)}]{bachman2017learningalgforactivelearning}
Bachman, P.; Sordoni, A.; and Trischler, A. 2017.
\newblock Learning algorithms for active learning.
\newblock In \emph{International Conference on Machine Learning (ICML)}, 301--310.

\bibitem[{Balcan, Beygelzimer, and Langford(2006)}]{balcan2006agnostic}
Balcan, M.-F.; Beygelzimer, A.; and Langford, J. 2006.
\newblock Agnostic active learning.
\newblock In \emph{International Conference on Machine Learning (ICML)}, 65--72.

\bibitem[{Beeson and Montana(2022)}]{beeson2022improving}
Beeson, A.; and Montana, G. 2022.
\newblock Improving {TD}3-{BC}: Relaxed Policy Constraint for Offline Learning and Stable Online Fine-Tuning.
\newblock In \emph{3rd Offline RL Workshop: Offline RL as a ''Launchpad''}.

\bibitem[{Bellemare et~al.(2016)Bellemare, Srinivasan, Ostrovski, Schaul, Saxton, and Munos}]{10.5555/3157096.3157262}
Bellemare, M.~G.; Srinivasan, S.; Ostrovski, G.; Schaul, T.; Saxton, D.; and Munos, R. 2016.
\newblock Unifying Count-Based Exploration and Intrinsic Motivation.
\newblock In \emph{Proceedings of the 30th International Conference on Neural Information Processing Systems}, 1479–1487.

\bibitem[{Blondel et~al.(2008)Blondel, Guillaume, Lambiotte, and Lefebvre}]{Blondel_2008}
Blondel, V.~D.; Guillaume, J.-L.; Lambiotte, R.; and Lefebvre, E. 2008.
\newblock Fast unfolding of communities in large networks.
\newblock \emph{Journal of Statistical Mechanics: Theory and Experiment}, (10): P10008.

\bibitem[{Brockman et~al.(2016)Brockman, Cheung, Pettersson, Schneider, Schulman, Tang, and Zaremba}]{brockman2016openai}
Brockman, G.; Cheung, V.; Pettersson, L.; Schneider, J.; Schulman, J.; Tang, J.; and Zaremba, W. 2016.
\newblock {OpenAI} Gym.
\newblock \emph{arXiv preprint arXiv:1606.01540}.

\bibitem[{Burda et~al.(2019)Burda, Edwards, Storkey, and Klimov}]{1810-12894}
Burda, Y.; Edwards, H.; Storkey, A.; and Klimov, O. 2019.
\newblock Exploration by random network distillation.
\newblock In \emph{International Conference on Learning Representations (ICLR)}, 1--17.

\bibitem[{Cohn, Ghahramani, and Jordan(1996)}]{cohn1996activelearning}
Cohn, D.~A.; Ghahramani, Z.; and Jordan, M.~I. 1996.
\newblock Active learning with statistical models.
\newblock \emph{Journal of Artificial Intelligence Research}, 4: 129--145.

\bibitem[{Ecoffet et~al.(2021)Ecoffet, Huizinga, Lehman, Stanley, and Clune}]{ecoffet2021goExplore}
Ecoffet, A.; Huizinga, J.; Lehman, J.; Stanley, K.~O.; and Clune, J. 2021.
\newblock First return, then explore.
\newblock \emph{Nature}, 590(7847): 580--586.

\bibitem[{Fujimoto and Gu(2021)}]{NEURIPS2021_a8166da0}
Fujimoto, S.; and Gu, S.~S. 2021.
\newblock A Minimalist Approach to Offline Reinforcement Learning.
\newblock In \emph{Advances in Neural Information Processing Systems}, volume~34, 20132--20145.

\bibitem[{Fujimoto, Meger, and Precup(2019)}]{fujimoto2019offpolicy}
Fujimoto, S.; Meger, D.; and Precup, D. 2019.
\newblock Off-policy deep reinforcement learning without exploration.
\newblock In \emph{International Conference on Machine Learning (ICML)}, 2052--2062.

\bibitem[{Gal, Islam, and Ghahramani(2017)}]{gal2017deep}
Gal, Y.; Islam, R.; and Ghahramani, Z. 2017.
\newblock Deep {B}ayesian Active Learning with Image Data.
\newblock In \emph{International Conference on Machine Learning (ICML)}, 1183--1192.

\bibitem[{Kaelbling, Littman, and Moore(1996)}]{kaelbling1996reinforcement}
Kaelbling, L.~P.; Littman, M.~L.; and Moore, A.~W. 1996.
\newblock Reinforcement Learning: A Survey.
\newblock \emph{Journal of Artificial Intelligence Research}, 4: 237--285.

\bibitem[{Kidambi et~al.(2020)Kidambi, Rajeswaran, Netrapalli, and Joachims}]{kidambi2020morel}
Kidambi, R.; Rajeswaran, A.; Netrapalli, P.; and Joachims, T. 2020.
\newblock {MOREL}: Model-based offline reinforcement learning.
\newblock In \emph{Advances in Neural Information Processing Systems}, volume~33, 21810--21823.

\bibitem[{Konyushova et~al.(2021)Konyushova, Chen, Paine, Gulcehre, Paduraru, Mankowitz, Denil, and de~Freitas}]{konyushova2021active}
Konyushova, K.; Chen, Y.; Paine, T.; Gulcehre, C.; Paduraru, C.; Mankowitz, D.~J.; Denil, M.; and de~Freitas, N. 2021.
\newblock Active offline policy selection.
\newblock In \emph{Advances in Neural Information Processing Systems}, volume~34, 24631--24644.

\bibitem[{Kostrikov, Nair, and Levine(2022)}]{IQL}
Kostrikov, I.; Nair, A.; and Levine, S. 2022.
\newblock Offline Reinforcement Learning with Implicit Q-Learning.
\newblock In \emph{International Conference on Learning Representations (ICML)}.

\bibitem[{Kumar et~al.(2020)Kumar, Zhou, Tucker, and Levine}]{CQL}
Kumar, A.; Zhou, A.; Tucker, G.; and Levine, S. 2020.
\newblock Conservative Q-Learning for Offline Reinforcement Learning.
\newblock In \emph{Advances in Neural Information Processing Systems}, volume~33, 1179--1191.

\bibitem[{Levine et~al.(2020)Levine, Kumar, Tucker, and Fu}]{levine2020offline}
Levine, S.; Kumar, A.; Tucker, G.; and Fu, J. 2020.
\newblock Offline reinforcement learning: Tutorial, review, and perspectives on open problems.
\newblock \emph{arXiv preprint arXiv:2005.01643}.

\bibitem[{Mai, Mani, and Paull(2022)}]{mai2022sample}
Mai, V.; Mani, K.; and Paull, L. 2022.
\newblock Sample Efficient Deep Reinforcement Learning via Uncertainty Estimation.
\newblock In \emph{International Conference on Learning Representations (ICLR)}.

\bibitem[{Nair et~al.(2020)Nair, Dalal, Gupta, and Levine}]{d4rl}
Nair, A.; Dalal, M.; Gupta, A.; and Levine, S. 2020.
\newblock Accelerating Online Reinforcement Learning with Offline Datasets.
\newblock \emph{CoRR}, abs/2006.09359.

\bibitem[{Osband et~al.(2016)Osband, Blundell, Pritzel, and Van~Roy}]{NIPS2016_8d8818c8}
Osband, I.; Blundell, C.; Pritzel, A.; and Van~Roy, B. 2016.
\newblock Deep Exploration via Bootstrapped DQN.
\newblock In \emph{Advances in Neural Information Processing Systems}, volume~29.

\bibitem[{Paszke et~al.(2019)Paszke, Gross, Massa, Lerer, Bradbury, Chanan, Killeen, Lin, Gimelshein, Antiga, Desmaison, Kopf, Yang, DeVito, Raison, Tejani, Chilamkurthy, Steiner, Fang, Bai, and Chintala}]{NEURIPS2019_9015}
Paszke, A.; Gross, S.; Massa, F.; Lerer, A.; Bradbury, J.; Chanan, G.; Killeen, T.; Lin, Z.; Gimelshein, N.; Antiga, L.; Desmaison, A.; Kopf, A.; Yang, E.; DeVito, Z.; Raison, M.; Tejani, A.; Chilamkurthy, S.; Steiner, B.; Fang, L.; Bai, J.; and Chintala, S. 2019.
\newblock {PyTorch}: An Imperative Style, High-Performance Deep Learning Library.
\newblock In \emph{Advances in Neural Information Processing Systems}, 8024--8035.

\bibitem[{Pathak et~al.(2017)Pathak, Agrawal, Efros, and Darrell}]{8014804}
Pathak, D.; Agrawal, P.; Efros, A.~A.; and Darrell, T. 2017.
\newblock Curiosity-Driven Exploration by Self-Supervised Prediction.
\newblock In \emph{IEEE Conference on Computer Vision and Pattern Recognition Workshops (CVPRW)}, 488--489.

\bibitem[{Rudin et~al.(2022)Rudin, Hoeller, Reist, and Hutter}]{rudin2022learningwalkminutesusing}
Rudin, N.; Hoeller, D.; Reist, P.; and Hutter, M. 2022.
\newblock Learning to walk in minutes using massively parallel deep reinforcement learning.
\newblock In \emph{Conference on Robot Learning (CoRL)}, 91--100.

\bibitem[{Seno and Imai(2022)}]{d3rlpy}
Seno, T.; and Imai, M. 2022.
\newblock d3rlpy: An Offline Deep Reinforcement Learning Library.
\newblock \emph{Journal of Machine Learning Research}, 23(315): 1--20.

\bibitem[{Settles(2011)}]{settlesactivelearninginpractice}
Settles, B. 2011.
\newblock From Theories to Queries: Active Learning in Practice.
\newblock In Guyon, I.; Cawley, G.; Dror, G.; Lemaire, V.; and Statnikov, A., eds., \emph{Active Learning and Experimental Design workshop In conjunction with AISTATS 2010}, volume~16 of \emph{Proceedings of Machine Learning Research}, 1--18. Sardinia, Italy: PMLR.

\bibitem[{Sutton and Barto(2018)}]{sutton2018reinforcement}
Sutton, R.~S.; and Barto, A.~G. 2018.
\newblock \emph{Reinforcement Learning: An Introduction}.
\newblock Cambridge: MIT Press, 2nd edition.

\bibitem[{Todorov, Erez, and Tassa(2012)}]{mujoco}
Todorov, E.; Erez, T.; and Tassa, Y. 2012.
\newblock MuJoCo: A physics engine for model-based control.
\newblock In \emph{IEEE/RSJ International Conference on Intelligent Robots and Systems}, 5026--5033. IEEE.

\bibitem[{Wu et~al.(2021)Wu, Zhai, Srivastava, Susskind, Zhang, Salakhutdinov, and Goh}]{wu2021uncertainty}
Wu, Y.; Zhai, S.; Srivastava, N.; Susskind, J.~M.; Zhang, J.; Salakhutdinov, R.; and Goh, H. 2021.
\newblock Uncertainty Weighted Actor-Critic for Offline Reinforcement Learning.
\newblock In \emph{International Conference on Machine Learning (ICML)}, 11319--11328.

\bibitem[{Yin et~al.(2023)Yin, Thiagarajan, Lazic, Rajaraman, Hao, and Szepesvari}]{yin2023sample}
Yin, D.; Thiagarajan, S.; Lazic, N.; Rajaraman, N.; Hao, B.; and Szepesvari, C. 2023.
\newblock Sample Efficient Deep Reinforcement Learning via Local Planning.
\newblock arXiv:2301.12579.

\bibitem[{Yu et~al.(2021)Yu, Kumar, Rafailov, Rajeswaran, Levine, and Finn}]{yu2021combo}
Yu, T.; Kumar, A.; Rafailov, R.; Rajeswaran, A.; Levine, S.; and Finn, C. 2021.
\newblock {COMBO}: Conservative offline model-based policy optimization.
\newblock In \emph{Advances in Neural Information Processing Systems}, volume~34, 28954--28967.

\bibitem[{Yu et~al.(2020)Yu, Thomas, Yu, Ermon, Zou, Levine, Finn, and Ma}]{yu2020mopo}
Yu, T.; Thomas, G.; Yu, L.; Ermon, S.; Zou, J.~Y.; Levine, S.; Finn, C.; and Ma, T. 2020.
\newblock {MOPO}: Model-based offline policy optimization.
\newblock In \emph{Advances in Neural Information Processing Systems}, volume~33, 14129--14142.

\bibitem[{Zhuang et~al.(2023)Zhuang, Kun, Liu, Wang, and Guo}]{zhuang2023behaviorproximalpolicyoptimization}
Zhuang, Z.; Kun, L.; Liu, J.; Wang, D.; and Guo, Y. 2023.
\newblock Behavior Proximal Policy Optimization.
\newblock In \emph{International Conference on Learning Representations (ICLR)}.

\end{thebibliography}
